\RequirePackage{fix-cm}
\documentclass{svjour3}       % onecolumn (second format)
\smartqed  % flush right qed marks, e.g. at end of proof
\usepackage{graphicx}
\usepackage{amsmath}
\usepackage[lmargin=1in,rmargin=1in,bmargin=1in]{geometry}
\usepackage{comment}
\newcommand{\twocases}[4]{\left\{ \begin{array}{ll} #1 &
                \mbox{#2} \\ #3 & \mbox{#4} \\
           \end{array} \right.}

\newcommand{\equ}[1]{(\ref{EQU:#1})}

\newcommand{\eps}       {\varepsilon}

\newcommand{\singval}{s}
\newcommand{\eigval}{\lambda}
\newcommand{\sumSQsing}{\Lambda}

\newcommand{\Std}{\sigma}

\newcommand{\LayerIndex}{k}   % \ell
\newcommand{\TrainData}{x} % training data for the entire DN

\newcommand{\inlayer}{z}
\newcommand{\inlayerone}{z_1}
\newcommand{\inlayeri}{z_{\LayerIndex}}
\newcommand{\inlayert}{z}     % trained layer
\newcommand{\outlayer}{z}  %
\newcommand{\outlayeri}{z_{\LayerIndex+1}}
\newcommand{\outlayerp}{z_{p+1}}
\newcommand{\outDN}{\widehat{y}}    % {y_p}
\newcommand{\outlayert}{z_1}    % trained layer

\newcommand{\reals}{{\rm I \mkern-2.5mu
              \nonscript\mkern-.5mu R}}

\newcommand{\dcounter}{g}

\newtheorem{thm}{Theorem}
\newtheorem{prop}[thm]{Proposition}
\newtheorem{cor}[thm]{Corollary}
\newtheorem{lem}[thm]{Lemma}
\newtheorem{defi}[thm]{Definition}
\usepackage{wasysym}

\usepackage{color}

\newcommand {\rolf}[1]{{\color{red}\sf{[rolf: #1]}}}

\newcommand{\PertVar}{\rho}
\newcommand{\CondNb}{\kappa}
\newcommand{\AnyM}{A}  % any matrix
\newcommand{\AnyMt}{{\tilde{\AnyM}}}  % any matrix tilde
\newcommand{\rr}{r'}

\newcommand{\MoreDetails}[1]{}

\newcommand{\Mult}{{\mathcal{E}}}
\newcommand{\discr}{{\chi}}

\newcommand{\batch}{{\rm batch}}
\newcommand{\Lbatch}{{L_\batch}}
\newcommand{\Pbatch}{{P_\batch}}
\newcommand{\Qbatch}{{Q_\batch}}

\begin{document}

\title{Singular Value Perturbation and Deep Network Optimization}

\author{Rudolf~H.~Riedi\and Randall~Balestriero\and Richard~G.~Baraniuk% <-this % stops a space
}

\institute{Rudolf~H.~Riedi \at
              Mathematics Department, HES-SO
The University of Applied Sciences of Western Switzerland\\ Fribourg, Switzerland
           \and
           Randall Balestriero \at
           Meta AI, FAIR\\
           NYC, USA
           \and
           Richard G.\ Baraniuk \at
           Department
of Electrical and Computer Engineering\\
Rice University, Houston, TX, 77005 USA
}

\date{Received: 21 November 2021 / Accepted: date}
\date{}
\maketitle

\begin{abstract}
\noindent
We develop new theoretical results on matrix perturbation to shed light on the impact of architecture on the performance of a deep network.
In particular, we explain analytically what deep learning practitioners have long observed empirically: the parameters of some deep architectures (e.g., residual networks, ResNets, and Dense networks, DenseNets) are easier to optimize than others (e.g., convolutional networks, ConvNets). 
Building on our earlier work connecting deep networks with continuous piecewise-affine splines, we develop an exact local linear representation of a deep network layer for a family of modern deep networks that includes ConvNets at one end of a spectrum and ResNets, DenseNets, and other networks with skip connections at the other.
For regression and classification tasks that optimize the squared-error loss, we show that the optimization loss surface of a modern deep network is piecewise quadratic in the parameters, with local shape governed by the singular values of a matrix that is a function of the local linear representation.
We develop new perturbation results for how the singular values of matrices of this sort behave as we add a fraction of the identity and multiply by certain diagonal matrices.
A direct application of our perturbation results explains analytically why a network with skip connections (such as a ResNet or DenseNet) is easier to optimize than a ConvNet: thanks to its more stable singular values and smaller condition number, the local loss surface of such a network is less erratic, less eccentric, and features local minima that are more accommodating to gradient-based optimization.
Our results also shed new light on the impact of different nonlinear activation functions on a deep network's singular values, regardless of its architecture.
\keywords{Deep Learning, Optimization Landscape, Perturbation Theory, Singular Values \\[5mm]
{\bf Dedication}~ We dedicate this paper to Prof.\ Ron DeVore to celebrate his 80th birthday and his lifetime of achievements in approximation theory.}
\end{abstract}

\newpage

\section{Introduction}
\label{sec:intro}

% DEEP LEARNING
Deep learning has significantly advanced our ability to address a wide range of difficult inference and approximation problems. 
Today's machine learning landscape is dominated by deep (neural) networks (DNs) \cite{lecun2015deep}, which are compositions of a large number of simple parameterized linear and nonlinear operators, known as {\em layers}. 
An all-too-familiar story of late is that of plugging a DN into an application as a black box, learning its parameter values using gradient descent optimization with copious training data, and then significantly improving performance over classical task-specific approaches.

% LOTS OF MODELS, WHAT MAKES THEM DIFFERENT?
Over the past decade, a menagerie of DN architectures has emerged, including Convolutional Neural Networks (ConvNets) that feature affine convolution operations \cite{lecun1995convolutional}, Residual Networks (ResNets) that extend ConvNets with skip connections that jump over some layers \cite{he2016deep}, Dense Networks (DenseNets) with several parallel skip connections \cite{huang2017densely}, and beyond.
A natural question for the practitioner is: Which architecture should be preferred for a given task?
Approximation capability does not offer a point of differentiation, because, as their size (number of parameters) grows, these and most other DN models attain universal approximation capability \cite{daubechies2021neural}.

\begin{figure}[t!]
    \centering
\begin{minipage}{1\linewidth}
\centering
\begin{minipage}{0.45\linewidth}
\centering
ConvNet\\
\includegraphics[width=\linewidth]{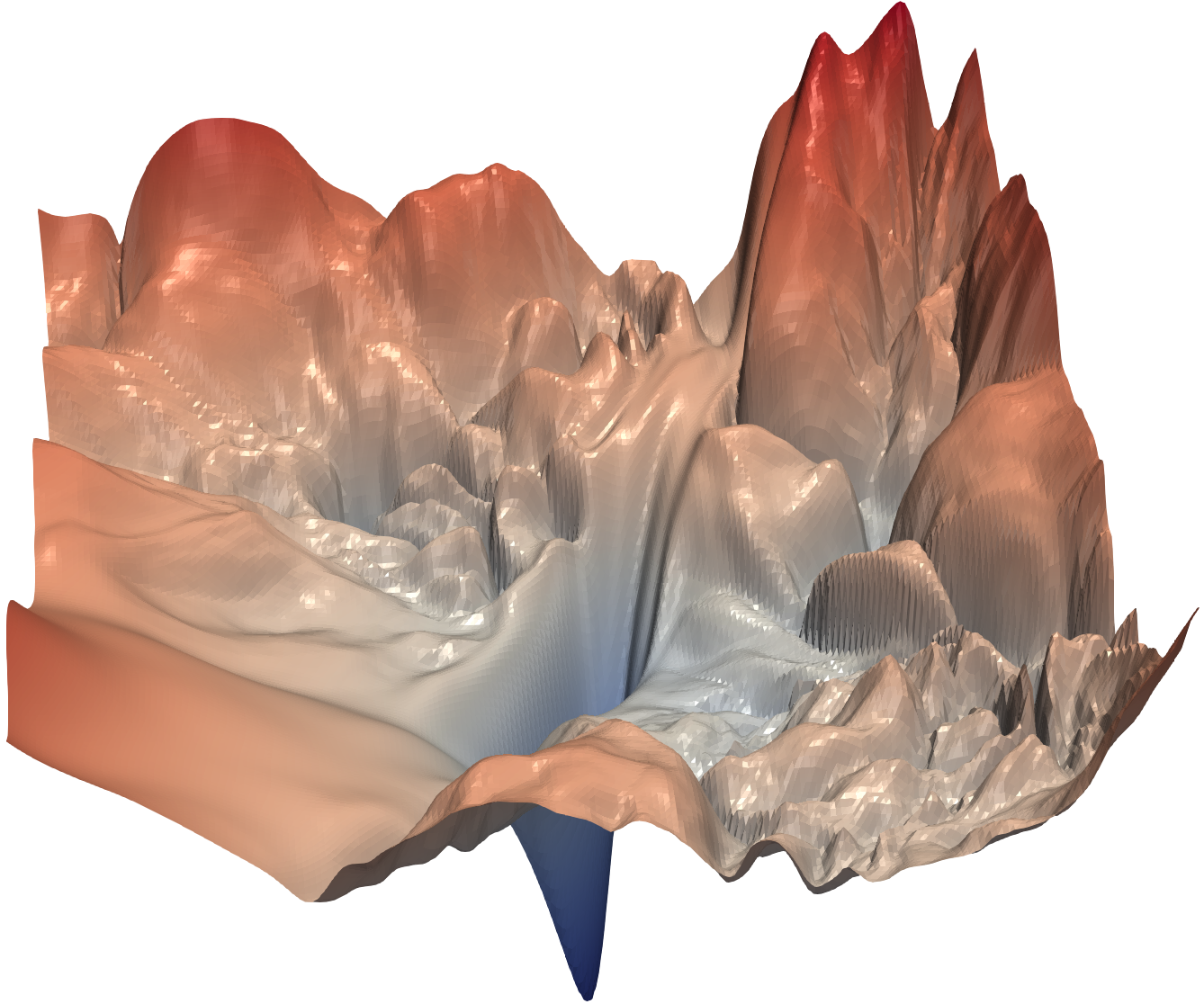}
\end{minipage}
\hfill
\begin{minipage}{0.45\linewidth}
\centering
ResNet\\[1.6cm]
\includegraphics[width=\linewidth]{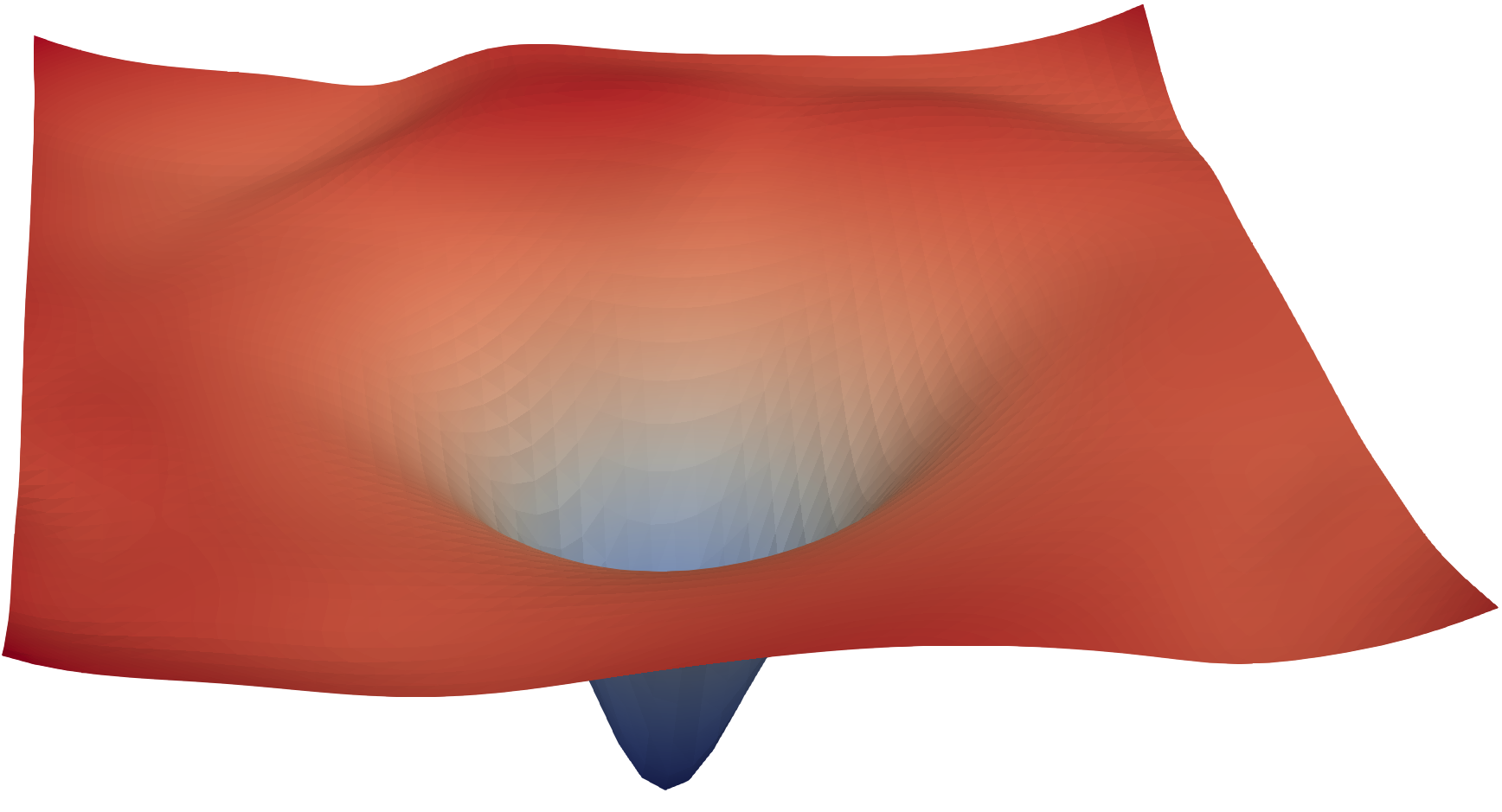}
\end{minipage}
\end{minipage}
\begin{minipage}{1\linewidth}
    \caption{\small
    Optimization loss landscape of two deep networks (DNs) along a 2D slice of their parameter space (from \cite{li2018visualizing}).
    (left) Convolutional neural network (ConvNet) with no skip connections. 
    (right) Residual network (ResNet) with skip connections.    
    {\em This paper develops new matrix perturbation theory tools to understand these landscapes and in particular explain why skip connections produce landscapes that are less erratic, less eccentric, and feature local minima that are more accommodating to gradient-based optimization.} 
    Used with permission. }
    \label{fig:loss_surface}
\end{minipage}
    \vspace*{-3mm}
\end{figure}

% DIFFERENCE IS IN LEARNING
Practitioners know that DNs with skip connections, such as ResNets and DenseNets, are much preferred over ConvNets, because empirically their gradient descent learning converges faster and more stably to a better solution.
In other words, it is not {\em what} a DN can approximate that matters, but rather {\em how it learns} to approximate. 
Empirical studies \cite{li2018visualizing} have indicated that this is because the so-called {\em loss landscape} of the objective function navigated by gradient descent as it optimizes the DN parameters is much smoother for ResNets and DenseNets as compared to ConvNets (see Figure~\ref{fig:loss_surface}).
However, to date there has been no analytical work in this direction.

% THE CRUX 
In this paper, we provide the first analytical characterization of the local properties of the DN loss landscape that enables us to quantitatively compare different DN architectures.
The key is that, since the layers of all modern DNs (ConvNets, ResNets, and DenseNets included) are continuous piecewise-affine (CPA) spline mappings \cite{reportRB,balestriero2018spline}, {\em the loss landscape is a continuous piecewise function of the DN parameters}.
In particular, for regression and classification problems where the DN approximates a function by minimizing the $\ell_2$-norm squared-error, we show that the loss landscape is a continuous piecewise quadratic function of the DN parameters.
The local eccentricity of this landscape and width of each local minimum basin is governed by the {\em singular values} of a matrix that is a function of not only the DN parameters but also the DN architecture.
This enables us to quantitatively compare different DN architectures in terms of their singular values.

% DIG DEEPER
Let us introduce some notation to elaborate on the above programme.
We study state-of-the-art DNs whose layers $f_k$ comprise a nonlinear, continuous piecewise-linear {\em activation function} $\phi$ that is applied element-wise to the output of an affine transformation.
We focus on the ubiquitous class of activations $\phi(t) = \max(\eta t,t)$, which yields the so-called rectified linear unit (ReLU) for $\eta=0$, leaky-ReLU for small $\eta\geq 0$, and absolute value for $\eta=-1$ \cite{goodfellow2016deep}.
In an abuse of notation that is standard in the machine learning community, $\phi$ can also operate on vector inputs by applying the above activation function to each coordinate of the vector input separately.
Focusing for this introduction on ConvNets and ResNets (we deal with DenseNets in Section \ref{sec:DeepLearning}), denote the input to layer $k$ by ${\inlayer}_k$ and the output by ${\outlayer}_{k+1}$.
Then we can write 
\begin{equation}
	{\outlayer}_{k+1} = f_k({\inlayer}_k) := \phi_k(W_k {\inlayer}_k+b_k) + \PertVar {\inlayer}_k,
		\label{eq:layer}
\end{equation}
where $W_k$ is an $n\times n$ {\em weight matrix} and $b_k$ is a vector of offsets.\footnote{
Without loss of generality, we assume that all $W_k$ are square. 
Rectangular $W_k$ are easily handled by extending with zeros.
}
Typically $W_k$ is a (strided) 
%circulent 
convolution matrix.
The choice $\PertVar=0$ corresponds to a ConvNet, while $\PertVar=1$ corresponds to a ResNet.

% CPA SPLINES
Previous work \cite{balestriero2018spline,reportRB} has demonstrated that the operator $f_k$ is a collection of continuous piecewise-affine (CPA) splines that partition the layer's input space into polytopal regions and fit a different affine transformation on each region with the constraint that the overall mapping is continuous.
This means that, locally around the input $\inlayeri$ and the DN parameters $W_k, b_k$, (\ref{eq:layer}) can be written as 
\begin{equation}
    {\inlayer}_{k+1} = D_k W_k {\inlayer}_k+ D_k b_k + \PertVar {\inlayer}_k,
\label{eq:linear}
\end{equation}
where the diagonal matrix $D_k$ contains 1s at the positions where the corresponding entries of $W_k {\inlayer}_k+b_k$ are positive and $\eta$ where they are negative.

% LEARNING
\sloppy
In supervised learning, we are given a set of $G$ labeled training data pairs $\{\TrainData^{(\dcounter)}, y^{(\dcounter)}\}_{\dcounter=1}^G$, and we tune the DN parameters $W_k, b_k$ such that, when datum $\TrainData^{(\dcounter)}$ is input to the DN, the output $\widehat{y}^{(\dcounter)}$ is close to the true label $y^{(\dcounter)}$ as measured by some {\em loss function} $L$.
In this paper, we focus on the $\ell_2$-norm squared-error loss averaged over a subset of the training data (called a {\em mini-batch})
\begin{equation}
    L:=
    \frac 1 G
     \sum_{g=1}^G 
%submitted:
    %\sum_{(\dcounter)} 
    \left\| y^{(\dcounter)}-\widehat{y}^{(\dcounter)} \right\|_2^2.
    \label{eq:minibatch}
\end{equation}
Standard practice is to use some flavor of gradient descent to iteratively reduce $L$ by differentiating with respect to $W_k, b_k$.
For ease of exposition, we focus our analysis first on the case of {\em fully stochastic gradient descent} that involves only a single data point per gradient step (i.e., $G=1$), meaning that we iteratively minimize  
\begin{equation}
    L^{(\dcounter)} :=
    \left\| y^{(\dcounter)}-
    \widehat{y}^{(\dcounter)} \right\|_2^2
    \label{eq:loss}
\end{equation}
for different choices of $\dcounter$.
We then extend our theoretical results to arbitrary $G>1$ in Section~\ref{subsec:batches}.
Let ${\inlayer}_k^{(\dcounter)}$ denote the input to the $k$-th layer when the DN input is $\TrainData^{(\dcounter)}$.
Using the CPA spline formulation (\ref{eq:linear}), 
it is easy to  show (see (\ref{eq:loss_squared_out}))
%in Section~\ref{sec:DeepLearning} 
that, for fixed $y^{(\dcounter)}$, 
%and $\inlayeri^{(\dcounter)}$, 
the loss function 
$L^{(\dcounter)}$ is continuous and {\em piecewise-quadratic} in 
$W_k, b_k$.

The squared-error loss $L^{(\dcounter)}$ is ubiquitous in {\em regression} tasks, where $y^{(\dcounter)}$ is real-valued, but also relevant for classification tasks, where $y^{(\dcounter)}$ is discrete-valued.  
Indeed, recent empirical work \cite{hui21} has demonstrated that state-of-the-art DNs trained for {\em classification} tasks using the squared-error loss perform as well or better than the same networks trained using the more popular cross-entropy loss.\footnote{
We note in passing that one can use the techniques developed in this paper to show that the cross-entropy loss is continuous and piecewise-cross-entropic in 
$W_k, b_k$.
Analysis of the local Hessian of this piecewise loss would lead to the same conclusion as with the squared-error loss, namely that the loss landscapes of ResNets/DenseNets are strictly better behaved than those of ConvNets.
We leave the details of this analysis for future work.}

%PIECEWISE QUADRATIC LOSS LANDSCAPE
We can characterize the local geometric properties of this piecewise-quadratic loss surface as follows.
First, without loss of generality, we simplify our notation.
For the rest of the paper, 
we label the layer of interest by $k=0$ 
and suppress the subscript $0$ (i.e., $W_0 \rightarrow W$); we assume that there are $p$ subsequent layers ($f_1,\dots,f_p$) between layer 0 and the output $\widehat{y}$.
Optimizing the weights $W$ of layer $f$ 
%with training data $\inlayert$ 
with training datum $\TrainData^{(\dcounter)}$
that produces layer input $\inlayert^{(\dcounter)}$ 
requires the analysis of the DN output $\widehat{y}$, due to the chain rule calculation of the gradient of $L$ with respect to $W$.
(As we discuss below, there is no need to analyze the optimization of $b$.)
For further simplicity, 
we suppress the superscript ${(\dcounter)}$ 
that enumerates the training data whenever possible. 
We thus write the DN output as
\begin{equation}
    \widehat{y} = M(\PertVar)DW {\inlayer} +B
\end{equation}
where 
\begin{equation}
    M(\PertVar)=D_pW_p \cdots D_1W_1+\PertVar\, {\rm Id}
    \label{EQU:M_intro}
\end{equation}
collects the combined effect of applying $W_k$ in subsequent layers $f_k$, $k=1,\dots,p$, and $B$ collects the combined offsets. 
Note that $\inlayert$ reflects the influence of the training datum 
combined with the action of the layers preceding $f$ and can be interpreted
as the input to the shallower network consisting of layers $f,f_1,\dots,f_p$.
This justifies using the index $0$ for the layer under consideration.

Using this notation and fixing $b$
as well as the input ${\inlayert}$, the piecewise-quadratic loss function $L$ can be written locally as a linear term in $W$ plus a quadratic term in $W$, which can be written as a quadratic form featuring matrix $Q$
(see Lemma~\ref{lem:standard_quadratic_form})
\begin{equation}
    {\inlayert}^TW^TD^TM(\PertVar)^TM(\PertVar)DW{\inlayert}
    =
    w^TQ^TQw.
    \label{eq:quad}
\end{equation}
Here, $w$ denotes the columnized vector version of the matrix $W$.

% SHAPE OF LOCAL LOSS SURFACE
The semi-axes of the ellipsoidal level sets of the local quadratic loss (\ref{eq:quad}) are determined by the {\em singular values} of the matrix $Q$, which we can write as $s_i \cdot |{\inlayert}[j]|$ according to Corollary~\ref{cor:standard_quadratic_form}, with $\singval_i=\singval_i(MD)$ the $i$-th singular value of the linear mapping $MD$ and ${\inlayert}[j]$ the $j$-th entry in the vector ${\inlayert}$.
The eccentricity of the ellipsoidal level sets is governed by the {\em condition number} of $Q$ which, therefore, factors as
\begin{equation}
\label{EQU:data_factor:intro}
\CondNb(Q) =
    \frac{\max_i s_i(MD)}{\min_i^* s_i(MD)}  \, \cdot \,
    \frac{\max_j {|\inlayert}[j]|}{\min_j^* |{\inlayert}[j]|}
    =
    \CondNb(MD)\cdot \CondNb(\mbox{\rm diag}({\inlayert})).
\end{equation}
Here, $\min^* $ 
denotes the minimum after discarding 
all vanishing elements (cf.~\equ{def_conditioning}). 
In this paper we focus on the effect of the DN architecture on the condition number of the matrix $MD$ representing the action of the subsequent layers, which is   
\begin{equation}
    \CondNb(MD)
    =
    \frac{\max_i s_i(MD)}{\min_i^* s_i(MD)}.
\end{equation}
However, the factorization in \equ{data_factor:intro} 
provides insights into the role played by the training datum ${\inlayert}$ (see Corollary~\ref{cor:standard_quadratic_form}).
%\rolf{In my view, we ignore since the z are the same in each step of training and independently of region. They are relevant because of the ``Data factor'' (see ): if data has a large condition number, we are already in bad shape whether we use ConvNet or ResNet. So this is an insight in addition to our comparison COnv vs Res. We focus in the paper on si and ignore z, because looking at z would be an entire new paper: how can you choose data z so that you improve condition..... Still we mention the ``data factor'' since it is relevant and leave it for someone else to study, or for our future work... whowever is faster:-) }
Extensions of these results to batches $(G>1)$ are found in Section~\ref{subsec:batches}.
(We note at this point that the optimization of the offset vector $b$ has no effect on the shape of the loss function.)

% DIFFERENCE BETWEEN CONVNET AND RESNET
For a fixed set of training data (i.e., fixed ${\TrainData}$), the difference between the loss landscapes of a ConvNet and a ResNet is determined soley by $\PertVar$ in \equ{M_intro}.
Therefore, we can make a fair, quantitative comparison between the loss landscapes of the ConvNet and ResNet architectures by studying how the singular values of the linear mapping $M(\PertVar)D$ evolve as we move infinitesimally from a ConvNet ($\PertVar=0$) towards a ResNet ($\PertVar=1$).

% NEW MATH
Addressing how the singular values and condition number of the linear mapping $M(\PertVar)D$ (and hence the DN optimization landscape) change when passing from $\PertVar=0$ towards $\PertVar=1$ requires a nontrivial extension of {\em matrix perturbation theory}.
Our focus and main contributions in this paper lie in exploring this under-explored territory.\footnote{A closely related situation occurs in Tikhonov regularization, where Id is added to $M^TM$ in order to regularize the ill-posed problem $Mx=b$.}
Our analysis technique is new, and so we dedicate most of the paper to its development, which is potentially of independent interest.

We briefly summarize our key theoretical findings from Sections~\ref{sec:101}--\ref{sec:Identity_withDiagonal} and our penultimate result in Theorem~\ref{thm:chi_bounds}.
We provide concrete bounds on the growth or decay of the singular values of any square matrix $M_o=M(0)$ when it is perturbed to $M(\PertVar) = M_o + \PertVar\,{\rm Id}$ by adding a multiple of the identity $\PertVar\,{\rm Id}$ for $0\leq \PertVar \leq 1$.
The bounds are in terms of the largest and smallest eigenvalues of the symmetrized matrix $M_o^T+ M_o$.
The behavior of the largest singular value under this perturbation can be controlled quite accurately in the sense that random matrices tend to follow the upper bound we provide quite closely with high probability.
%\richb{explain what you mean by ``hard'' upper bound}
%\rolf{hard=non-probabilistic. the term ``hard'' attempts to express the fact, that this bound hold for any matrix. But it is an upper bound on the norm of the matrix. just a bound. the statement means to say the bound is ``efficient, quite tight (=good)'' in the sense that random matrices have a norm that is quite close to the bound with high probability. in other words, the probability, that the upper bound completely overestimates the norm (=poor bound) is quite small. The term ``tend'' means that this is not an exact statement, and was observed in simulations. only in simulations, I know, but with overwhelming evidence. Short: the ``hard'' upper bound can not be improved without further assumptions. it is a ``good bound''}
For all other singular values, we require that they be non-degenerate, meaning that they have multiplicity $1$
for all but finitely many $\PertVar$, a property that holds for most matrices in a generic sense 
(see Lemma~\ref{lem:discriminant}).
Further, we establish bounds for
arbitrary singular values that do not hinge on the assumption of multiplicity $1$ but that are somewhat less tight.

%\richb{KEY: we need to state in this paragraph the key fact that the condition number is smaller when $\PertVar=1$ than when $\PertVar=0$!}\randall{yes this is key!}
%\rolf{just so we are all clear on this point:you cannot make this statement in all generality. There are exceptional cases...see Email Nov 7: \"Conv vs Res: kappa(0)>kappa(1) ?" Short: We need to be careful what we say. I see the following: 1) or Resent we have guarantees on condNb while for ConvNet all bets are off. 2) Sims that Randall is doing now (first weekend of Nov) show that in random settings this is what we observe in sims. }

Our new perturbation results enable us to draw three rigorous conclusions regarding the optimization landscape of a DN.  
First, with regards to network architecture, we establish that the addition of skip connections improves the condition number of a DN's optimization landscape.  
To this end, 
we point out that {\em no} guarantees on the piecewise-quadratic optimization landscape's condition number are available for ConvNets; in particular, $\CondNb (M(0))$ can take any value, even if the entries of $M(0)$ are bounded.
However, as we move from ConvNets ($\PertVar=0$) towards networks with skip connections such as ResNets and DenseNets ($\PertVar>0$), bounds on $\CondNb(M(\PertVar))$ become available.
For an appropriate random initialization of the weights and a wide network architecture (large $n$), we show that the condition number of a ResNet ($\PertVar=1$) is bounded above asymptotically by just $3$ 
(see \equ{conditioning_average_bound_asymptotics},  \equ{conditioning_average_bound_asymptotics_via_s1},
and
\equ{conditioning_ReLU_average_bound_via_s1}). 
Second, for the particular case of DNs employing absolute value activation functions, we show that the relevant condition number is easier to control than for ReLU activations, that is, absolute value requires less restrictive assumptions on the entries of $M$ in order to bound the condition number.
For absolute value, we also prove that the condition number of $M(1)$ (ResNet) is smaller than that of $M(0)$ (ConvNet) with probability asymptotically equal to $1$. 
Third, with regards to the influence of the training data, we prove that extreme values of the data negatively affect the performance of linear least squares and that this impact is tempered when applying mini-batch learning due to its inherent averaging.

% APPLY TO DNs
In the context of DN optimization, these results demonstrate that the local loss landscape of a DN with skip connections (e.g., ResNet, DenseNet) is better conditioned than that of a ConvNet and thus less erratic, less eccentric, and with local minima that are more accommodating to gradient-based optimization,
%and hence smoother and easier to optimize \richb{is this obvious or do we need to back this point up?}.
%Applying Theorem XXX to DNs, we see that ResNet, conditioning of $s_i$ is better than ConvNet, or at least controllable under certain assumptions on the WEIGHTS W, irrespective of the data. Since ConvNet and ResNet use --- of course --- the same data, things can only be better for ResNet.
particularly when the weights in $W$ are small. 
This is typically the case at the initialization of the optimization \cite{glorot2011deep} and often remains true throughout training \cite{gal2016dropout}. 
In particular, our results also provide new insights into the best magnitude to randomly initialize the weights at the beginning of learning in order to reap the maximum benefit from a skip connection architecture.

This paper is organized as follows.
Section~\ref{sec:101} introduces our notation and provides a review of existing useful results on the perturbation of singular values. 
Section~\ref{sec:DeepLearning} overviews our prior work establishing that CPA DNs can be viewed as locally linear and leverages this essential property to identify the singular values relevant for understanding the shape of the quadratic loss surface used in gradient-based optimization. 
Section \ref{sec:AddingIdentity} develops the perturbation results needed to capture the influence on the singular values when adding a skip connection, i.e., when passing from $M_o$ to $M_o+{\rm Id}$. 
In particular, we provide bounds on the condition number of $M_o+\PertVar{\rm Id}$ in terms of the largest singular value of $M_o$ and the largest and the smallest eigenvalue of $M_o^T+M_o$.
Section \ref{sec:PerturbationDiagonal} combines the findings of Sections \ref{sec:DeepLearning} and \ref{sec:AddingIdentity} to show that a %ResNet 
layer with a skip connection and weights bounded by an appropriate constant will have a bounded condition number. 
Clearly, this applies to weights drawn from a uniform distribution.
Section \ref{sec:Identity_withDiagonal} is dedicated to a ResNet layer with random weights of bounded standard deviation.
Here, we establish bounds on the condition number of the corresponding ResNet and on the probability with which they hold, explicitly in terms of the width $n$ of the network. 
Further, we provide convincing numerical evidence that the asymptotic bounds apply in practice for widths as modest as $n\geq 5$. 
We conclude in Section~\ref{sec:conclusions} with a synthesis of our results and perspectives on future research directions.
All proofs are provided within the main paper.
Various empirical experiments sprinkled throughout the paper support our theoretical results.
%\section{Existing and Related Work}
\section{Background}
\label{sec:101}

\subsection{Notation}
\label{subsec:notation}

{\bf Singular values.}
    Let $\AnyM$ be a $n\times n$-matrix, and let $\|u\|$ denote the $\ell_2$-norm of the vector $u$.
    All products are to be understood as matrix multiplications,
    even if the factors are vectors. Vectors are always column vectors.

    Let $u_i$ denote a unit-eigenvector of $\AnyM^T\AnyM$
    to its  \emph{eigenvalue }
    $\eigval_i=\eigval_i(\AnyM^T\AnyM)$ such that
    the vectors $u_i$ ($i=1\ldots n$) form an orthonormal basis.

    The \emph{singular values} of $\AnyM$ are defined as the square roots of the eigenvalues of ${\AnyM}^T{\AnyM}$
\begin{equation}
        \singval_i
        =
        \singval_i(\AnyM)
        =
        \sqrt{\eigval_i({\AnyM}^T{\AnyM})}
        =
        \|{\AnyM}u_i\|.
\end{equation}
    Note that the singular values are always real and non-negative valued, since
    \begin{equation}
    0\leq \|{\AnyM}u_i\|^2=u_i^T{\AnyM}^T{\AnyM}u_i=\eigval_i u_i^Tu_i=\eigval_i .
    \end{equation}
    For reasons of definiteness, we take the eigenvalues of ${\AnyM}^T{\AnyM}$, and thus the singular values of ${\AnyM}$ to be ordered by size: $\singval_1\geq\ldots\geq\singval_n$.

{\bf Singular vectors.}
The vectors $u_i$ are called \emph{right singular vectors} of ${\AnyM}$.
The \emph{left singular vectors} of ${\AnyM}$ are denoted $v_i$ and are defined as an orthonormal basis with the property that
    \begin{equation}
    \label{EQU:singular_values}
        {\AnyM}u_i=\singval_i v_i
        \qquad\mbox{and}\qquad
        {\AnyM}^Tv_i=\singval_i u_i.
    \end{equation}
    Such a basis of left-singular vectors always exists.

{\bf SVD.} Collecting the right singular vectors $u_i$ as columns into an orthogonal matrix $U$, the left singular vectors $v_i$ into an orthogonal matrix $V$, and the singular values $\singval_i$ into a diagonal matrix $\Sigma$ in corresponding order, we obtain the Singular Value Decomposition (SVD) of ${\AnyM}=V\Sigma U^T$.

{\bf Eigenvectors.}
Note that ${\AnyM}$ may have eigenvectors that differ from the singular vectors. In particular, eigenvectors are not always orthogonal onto each other. Clearly, if ${\AnyM}$ is symmetric then the eigenvectors and right-singular vectors coincide and every \emph{eigenvalue $r_i$ of ${\AnyM}$} corresponds to a
    singular value $\singval_i=|r_i|$.
However, to avoid confusion, we will refrain from referring to the {eigenvalues} and the singular
values of the same matrix
    whenever possible.

{\bf Smooth matrix deformations.}
Consider a family of $n \times n$  matrices 
${\AnyM}(\PertVar)$, where the variable $\PertVar$ in assumed 
to lie in some interval $I$.
We say that the family ${\AnyM}(\PertVar)$ is $\mathcal{C}^{m}(I)$
if all its entries are $\mathcal{C}^{m}(I)$, with $m=0$ corresponding to being continuous
and $m=1$ to continuously differentiable.
If ${\AnyM}$ is $\mathcal{C}^{m}(I)$, then so is ${\AnyM}^T{\AnyM}$.

We denote the matrix of derivatives of the entries of ${\AnyM}$ by ${\AnyM}'(\PertVar)=\frac{d}{d\PertVar} {\AnyM}(\PertVar)$.
%Note that the associated eigenvectors and eigenvalues also depend on $\PertVar$.

{\bf Multiplicity of singular values}
Denote by
$S_i\subset I$ the set of $\PertVar$ values
%in the domain $I$ of ${\AnyM}(\PertVar)$
for which $\singval_i$ is a \emph{simple} singular value:
\begin{equation}\label{EQU:def_simple}
  S_i
  =
  \{\PertVar\in I
  \,:\,
  \singval_i(\PertVar)\neq\singval_j(\PertVar)
  \mbox{ for all $j\neq i$}\}.
\end{equation}
%and let $S_i^*\subset S_i$ be the set where
%$\singval_i(\PertVar)\neq0$.
\MoreDetails{The singular vectors $u_i$ are uniquely
defined on $S_i$ up to multiplication
by $-1$; the $v_i$ follow from the
$u_i$ according to (\ref{EQU:singular_values}).
When two or more singular values coincide
then the corresponding eigenspace of ${\AnyM}^T{\AnyM}$ is at least $2$-dimensional and
there is a freedom of choice of 
eigenvectors $u_i$ as the basis of that eigenspace.}
For later use, let
\begin{equation}\label{EQU:def_nonzero}
  S_i^*
  =
  \{\PertVar\in S_i
  \,:\,
  \singval_i(\PertVar)\neq0\}.
\end{equation}

For completeness,
we mention an obvious fact that follows from the SVD:
    \begin{equation}\label{EQU:transpose}
      \singval_i({\AnyM})
      =
      \singval_i \!\left({\AnyM}^T\right).
    \end{equation}
    
\subsection{Singular values under additive perturbation}

A classical result that will prove to be key in our study is due to Wielandt and Hoffman \cite{Wielandt_Hoffman}.
It uses the notion 
of the Frobenius norm 
$\|{\AnyM}\|_F$
of the $n\times n$-matrix ${\AnyM}$.
Denoting the entries of ${\AnyM}$ by ${\AnyM}[i,j]$
and its singular values by $\singval_i({\AnyM})$,
it is well known that
\begin{equation}\label{EQU:frobenius_norm}
  \|{\AnyM}\|_F^2
  =
  \sum_{i,j=1}^{n} |{\AnyM}[i,j]|^2
  =
  \mbox{\rm trace}({\AnyM}^T{\AnyM})
  =
  \sum_{i=1}^{n}\singval_i^2({\AnyM})
  .
\end{equation}
\begin{prop}[Wielandt and Hoffman]
  \label{prop:Frobenius_bound}
    Let ${\AnyM}$ and ${\AnyMt}$ be symmetric $n\times n$-matrices
    with eigenvalues ordered by size.
    Then
    \begin{equation}
    \label{EQU:frobenius_bound}
        \sum_{i=1}^{n}
        \Big|\eigval_i({\AnyM})-\eigval_i({\AnyMt})\Big|^2
    % result not to be trusted, by appears so ChafaÃ¯:
    %    \Big|\singval_i({\AnyM})-\singval_i(B)\Big|^2
        \leq
        \|{\AnyM}-{\AnyMt}\|_F^2.
    \end{equation}
\end{prop}
This result can easily be strengthened to read as follows.
\begin{cor}[Wielandt and Hoffman]
  \label{cor:Frobenius_bound}
    Let ${\AnyM}$ and ${\AnyMt}$ be any $n\times n$-matrices
    with singular values ordered by size.
    Then
    \begin{equation}
    \label{EQU:strong_frobenius_bound}
        \sum_{i=1}^{n}
        \Big|\singval_i({\AnyM})-\singval_i({\AnyMt})\Big|^2
        \leq
        \|{\AnyM}-{\AnyMt}\|_F^2.
    \end{equation}
\end{cor}
Note that this result, as most others, holds actually also for rectangular matrices
with the usual adjustments.

\begin{proof}
  We establish Corollary~\ref{cor:Frobenius_bound}.
  The $2n$ eigenvalues of the symmetric matrix
%  \begin{equation}\label{EQU:proof_cor_Frob}
\begin{equation}
    H_{\AnyM}
    =
    \left(
        \begin{array}{cc}
          0 & {\AnyM}  \\
          {\AnyM}^T & 0
        \end{array}
    \right)
\end{equation}
%\end{equation}
  are $\singval_i({\AnyM})$ with eigenvector
%  $\frac{1}{\sqrt 2}(v_i^T u_i^T)^T$
  $\frac{1}{\sqrt 2}{ v_i \choose u_i }$
  and $-\singval_i({\AnyM})$ with eigenvector
%  $\frac{1}{\sqrt 2}(-v_i^T u_i^T)^T$.
  $\frac{1}{\sqrt 2}{ -v_i \choose u_i }$.
  Also, $\|H_\AnyM\|_F^2=2\|{\AnyM}\|_F^2$. 
  Therefore, when applying Proposition~\ref{prop:Frobenius_bound} 
  with $\AnyM$ replaced by 
  $H_\AnyM$ and $\AnyMt$ replaced by $H_\AnyMt$,
  we run through the left-hand side of (\ref{EQU:strong_frobenius_bound}) twice
  and in turn obtain twice the right-hand side.
%  \hfill$\Box$
\end{proof}

Corollary~\ref{cor:Frobenius_bound}.
  immediately implies 
  a well-known result that will be key:
   the singular values of a matrix depend continuously on its entries, as we state next. 
   
\begin{lem}
    \label{lem:continuity}
    If ${\AnyM}(\PertVar)$ is $\mathcal{C}^{0}(I)$, then
    all its singular values $\singval_i(\PertVar)$ depend continuously on $\PertVar$.
    Consequently, all $S_i$ in \equ{def_simple} and 
    $S_i^*$ in \equ{def_nonzero} are then open.
\end{lem}

\begin{proof}

To establish continuity,
choose $\PertVar_1$ and
$\PertVar_2$ arbitrary and
apply Corollary~\ref{cor:Frobenius_bound} (in slight abuse of notation) with
${\AnyM}={\AnyM}(\PertVar_1)$ and
${\AnyMt}={\AnyM}(\PertVar_2)$.
%\hfill$\Box$
\end{proof}

As will become clear in the sequel,
the main difficulty in establishing
further results on the regularities of the singular values
with elementary
arguments and estimates lies
in dealing with multiple zeros, i.e., with
multiple singular values.

Also useful are some facts about the special role of the largest singular value.
\begin{prop}[Operator norm]
  \label{prop:operator_norm}
    Let ${\AnyM}$ be a  $n\times n$-matrix. Then, its largest singular value
    is equal to its operator norm, and we have the following inequalities:
    \begin{equation}
    \label{EQU:operator_norm}
        \singval_1({\AnyM})
        =
        \|{\AnyM}\|_{\rm Op}
        :=
        \max_{|x|=1}|\AnyM x|
        \leq
        \|{\AnyM}\|_F
        \leq
        \sqrt{{\rm rank}({\AnyM})}\cdot
        \|{\AnyM}\|_{\rm Op}.
        %\singval_1({\AnyM})
    \end{equation}
    In particular, if ${\AnyM}$ is rank 1, then both norms coincide.
    Also, being a norm, the largest singular value
    satisfies 
    \begin{equation}
    \label{EQU:norm_inequality}
    \singval_1({\AnyM}+{\AnyMt})\leq \singval_1({\AnyM})+\singval_1({\AnyMt})
    \end{equation}
for any matrices ${\AnyM}$ and ${\AnyMt}$.
\end{prop}
\begin{proof}
  The first equality follows from the fact that $\singval_1$ is the largest semi-axis of the ellipsoid that is the image of the unit-sphere.
  The first inequality follows from (\ref{EQU:frobenius_norm}).
  The last inequality follows again from considering the aforementioned ellipsoid.
\end{proof}

Another proof of the continuity of singular values
is found in the well-known interlacing properties of singular values,
which we adapt to our notation and summarize for the convenience of the reader
(see, e.g. \cite[Theorem~6.1.3-5.]{Chafai}).

\begin{prop}[Interlacing]
  \label{prop:interlacing}
    Let ${\AnyM}$ and ${\AnyMt}$ be $n\times n$-matrices
    with singular values ordered by size.
    \begin{itemize}
      \item (Weyl additive perturbation) We have:
    \begin{equation}
    \label{EQU:weyl}
        \singval_{i+j-1}({\AnyM})
        \leq
        \singval_i({\AnyMt})+\singval_j({\AnyM}-{\AnyMt}).
    \end{equation}
\item (Cauchy interlacing by deletion)
      Let ${\AnyMt}$ be obtained from ${\AnyM}$ by deleting $m$ rows or $m$ columns.
      Then
    \begin{equation}
    \label{EQU:deletion}
        \singval_{i}({\AnyM})
        \geq
        \singval_i({\AnyMt})
        \geq
        \singval_{i+m}({\AnyM}).
    \end{equation}
    \end{itemize}
\end{prop}

From (\ref{EQU:weyl}) applied with $j=1$,
we note that the singular values are in fact uniformly continuous.
Indeed, we have for any $i$ that
    \begin{equation}
    \label{EQU:uniform_bound}
        |\singval_{i}({\AnyM})
        -
        \singval_i({\AnyMt})
        |
        \leq
        \singval_1({\AnyM}-{\AnyMt})
        \leq
        \|{\AnyM}-{\AnyMt}\|_F.
    \end{equation}

\subsection{Singular values under smooth perturbation}

We now turn to the study of how singular values of ${\AnyM}(\PertVar)$ change as a function of $\PertVar$, thereby leveraging the concept of differentiability.

This can be done 
in several ways; the following computation is particularly simple
and well known.
Let $N(\PertVar)$ 
denote a family of  symmetric $\mathcal{C}^{1}(I)$ matrices.
Dropping all indices and variables $\PertVar$ for ease of reading, the eigenvalues and eigenvectors satisfy
\begin{eqnarray}
    Nu=\eigval u,
    \qquad
    u^Tu=1.
\end{eqnarray}
Denoting the derivatives with respect to $\PertVar$ by
$N'$, $\eigval'$ and $u'$ and
assuming they exist, we obtain
\begin{eqnarray}
    N' u+N u'
    =
    \eigval'u+\eigval  u',
    \qquad
    u'^Tu=u^Tu'=0.
\end{eqnarray}
We then left-multiply the first equality by $u^T$ and use the second equality
to find
\begin{eqnarray}
    u^T N' u + u^TN u'
    =
    \eigval' u^T u + \eigval u^T u'
    =
    \eigval'.
\end{eqnarray}
Finally, as $N$ is symmetric we see that  $u^T Nu'=(Nu)^Tu'=\eigval u^T u' = 0$, leading to
\begin{equation}
\eigval'=u^T N' u
.
\end{equation}

In quantum mechanics, the last equation 
is known as the Hellmann-Feynman Theorem and
was discovered independently by several authors in the 1930s.
%, as well as by the present author in 2019 and many others.
It is known, though seldom mentioned,
that the formula may fail when eigenvalues coincide. For further studies in this direction, we refer the interested reader to works in the field of degenerate perturbation theory.

For a proof of the Hellmann-Feynman Theorem, it is natural to 
consider the function $\Psi:\reals^{n+2}\to \reals^{n+1}$
    \begin{equation}
        \Psi(\PertVar,u,\eigval)
        =
    \left(\begin{array}{c}
        Nu-\eigval u \\
        u^Tu-1
        \end{array}
    \right),
    \end{equation}
where $u$ is an $n$-dimensional column vector.
    %with coordinates $u[j]$.
The  Implicit Function Theorem applied to $\Psi$ then yields the following result.

\begin{thm} [Hellmann-Feynman]
\label{thm:lambda_differentiability}
Assume $N(\PertVar)$ forms a family of symmetric $\mathcal{C}^{1}(I)$ matrices.
Then, as a function of $\PertVar$,
each of its eigenvalues $\eigval_i=\eigval_i(\PertVar)$ is continuously differentiable  where  it is simple. 
%More precisely, $\eigval_i$ is
%$\mathcal{C}^{1}(S_i)$
Its derivative reads as
\begin{equation}
    \label{EQU:derivative_lambda}
        \eigval_i'(\PertVar)
        =
        u_i^T N' u_i
%        =
%        u_i^T({\AnyM}'^T {\AnyM} + {\AnyM}^T {\AnyM}')u_i
%        =
%        2u_i^T{{\AnyM}'}^T {\AnyM} u_i
%        =
%        2\singval_i u_i^T{{\AnyM}'}^T v_i
        .
\end{equation}
Its corresponding right singular vector
$u_i(\PertVar)$ can be chosen such that
it becomes
$\mathcal{C}^{1}(S_i)$
as well.
\end{thm}
If in addition to the assumptions of Theorem~\ref{thm:lambda_differentiability}
the family $N$ is actually $\mathcal{C}^{m}(I)$,
then $\eigval_i(\PertVar)$ and $u_i(\PertVar)$
are $\mathcal{C}^{m}(S_i)$. 
From Hellmann-Feynmann we may conclude the following.

\begin{figure}
    \begin{center}
    \begin{tabular}{cc}
    \includegraphics[width=0.45\linewidth]{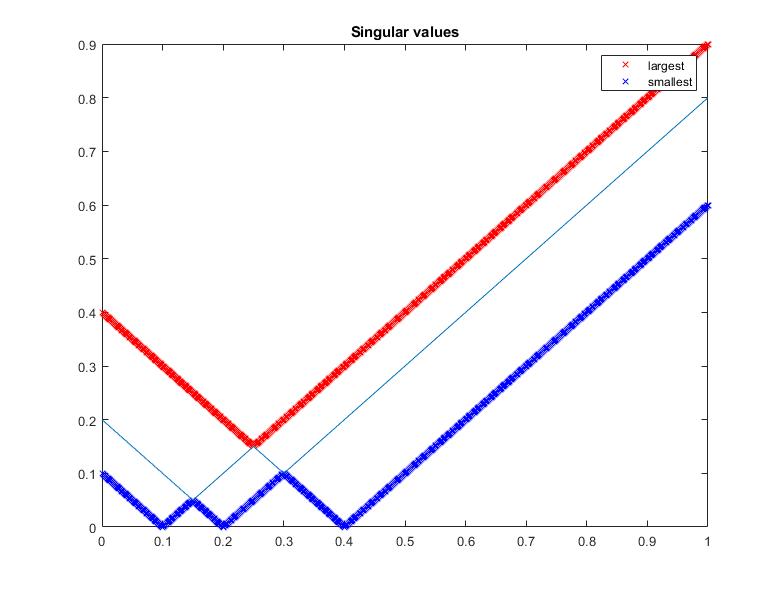}
    &
    \includegraphics[width=0.45\linewidth]{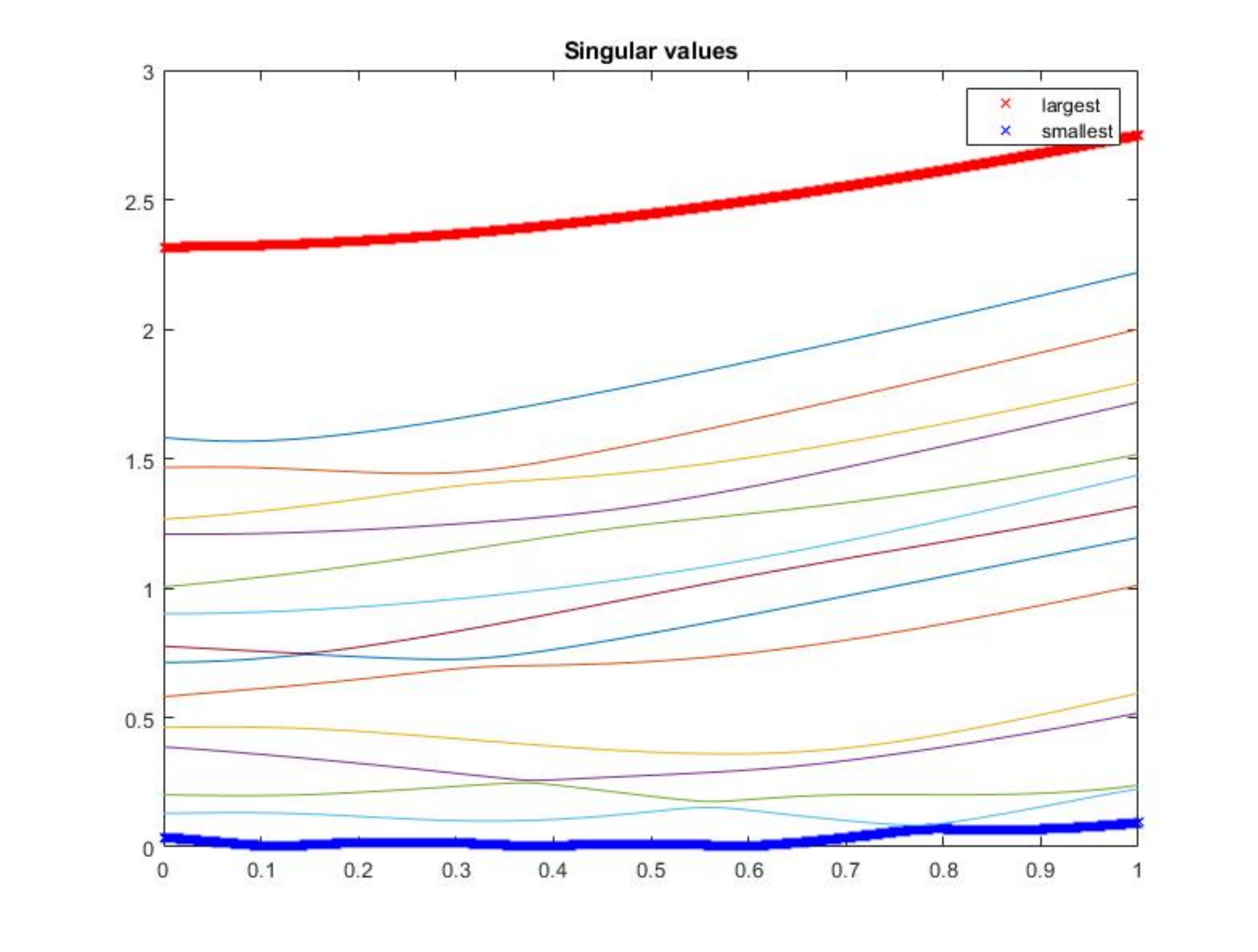}
    \\[-2mm]
    $\rho$ & $\rho$
    \\[3mm]
    (a) & (b)
    \end{tabular}
    \end{center}
    \caption{
    Singular values of two families of matrices of the form
    $M(\PertVar)=M_o+\PertVar{\rm Id}$ for $0\leq\PertVar\leq1$.
    (a) $M_o={\rm diag}(-0.4~~ -0.2 ~~-0.1)$
    demonstrates that even the largest singular value may not be
    increasing or everywhere differentiable.
    (b) The entries of the $15\times 15$ matrix $M_o$
    are drawn from a zero-mean Gaussian distribution with $\Std=0.3$. This example
    demonstrates that the singular values may be non-convex
    and may have several extrema even
    without coinciding with others.}
    \label{fig:ex_101}
\end{figure}

\begin{cor}
\label{cor:sigma_differentiability}
Let ${\AnyM}(\PertVar)$ be $\mathcal{C}^{1}(I)$.
Assume that $S_i^*$
consists of isolated points only. Then, $\singval_i(\AnyM)$
and its  left singular vector $v_i$
are $\mathcal{C}^{1}(S_i^*)$
with
\begin{equation}
    \singval'_i(\PertVar)
        =
        u_i(\PertVar)^T{{\AnyM}'(\PertVar)}^T v_i(\PertVar).
    %=\eigval_i'/(2\singval_i)=
    %=u_i^T\AnyM'^Tv_i.
\end{equation}
\end{cor}
Simple examples show that the above results cannot be improved
without stronger assumptions
(see Example 1 below and Figure~\ref{fig:ex_101}).

\begin{proof}
Apply Theorem~\ref{thm:lambda_differentiability}
to the family $N={\AnyM}^T {\AnyM}$. 
Note that
$N'={\AnyM}'^T {\AnyM} + {\AnyM}^T {\AnyM}'$
and apply the chain rule to
$\eigval_i=(\singval_i(\PertVar))^2$
to find the first equality of the following formula:
\begin{equation}
    \singval'_i
        =
        \frac{1}{2\singval_i}
        u_i^T({\AnyM}'^T {\AnyM} + {\AnyM}^T {\AnyM}')u_i
        =
        \frac{1}{2\singval_i}
        2u_i^T{{\AnyM}'}^T {\AnyM} u_i
        =
        u_i^T{{\AnyM}'}^T v_i.
    %=\eigval_i'/(2\singval_i)=
    %=u_i^T\AnyM'^Tv_i.
\end{equation}
Also, we have 
$u_i^T{\AnyM}'^T {\AnyM}u_i=(u_i^T{\AnyM}'^T {\AnyM}u_i)^T
=u_i^T{\AnyM}^T {\AnyM}'u_i$, since this is a real-valued number. This implies
the second equality in the formula above. 
Finally note that 
${\AnyM} u_i=\singval_i v_i$ to complete the proof.
%\hfill$\Box$
\end{proof}

Example 1:
The diagonal family $M=$diag$(-4 ~~ -2)+\PertVar${\rm Id} is
$\mathcal{C}^\infty(\reals)$.
When ordered by size,
the eigenvalues of $N=M^TM$
are $\max((\PertVar-4)^2,(\PertVar-2)^2)$
and
$\min((\PertVar-4)^2,(\PertVar-2)^2)$.
They are
not differentiable where they coincide.
In particular, the matrix norm of $N$ is not
differentiable at $\PertVar=3$.
The singular values of $M$ are
$\singval_1=\max(|\PertVar-4|,|\PertVar-2|)$ and
$\singval_2=\min(|\PertVar-4|,|\PertVar-2|)$; they are
not differentiable
at their respective zeros. Also, the left singular
vectors $v_i$ change sign at the zero of $\singval_i$
and so are not even continuous there
(see  Figure~\ref{fig:ex_101} for a similar example).

\section{Singular Values in Deep Learning}
\label{sec:DeepLearning}

In this section, we establish that the squared-error loss landscape of a deep network (DN) is governed locally by the singular values of the matrix of weights of a layer and that the difference between certain network architectures can be interpreted as a perturbation of this matrix. Combining this key insight with the perturbation theory of singular values for DNs, which will be developed in the sections to follow, will enable us to characterize the dependence of the loss landscape on the underlying architecture. Doing so, we will ultimately provide analytical insight into numerous phenomena that have been observed only empirically so far, such as the fact that so called ResNets \cite{he2016identity} are easier to optimize than ConvNets \cite{lecun1995convolutional}.

\subsection{The continuous piecewise-affine structure of deep networks}
\label{subsec:structure}

\begin{figure}
    \centering
    \includegraphics[width=0.3\linewidth]{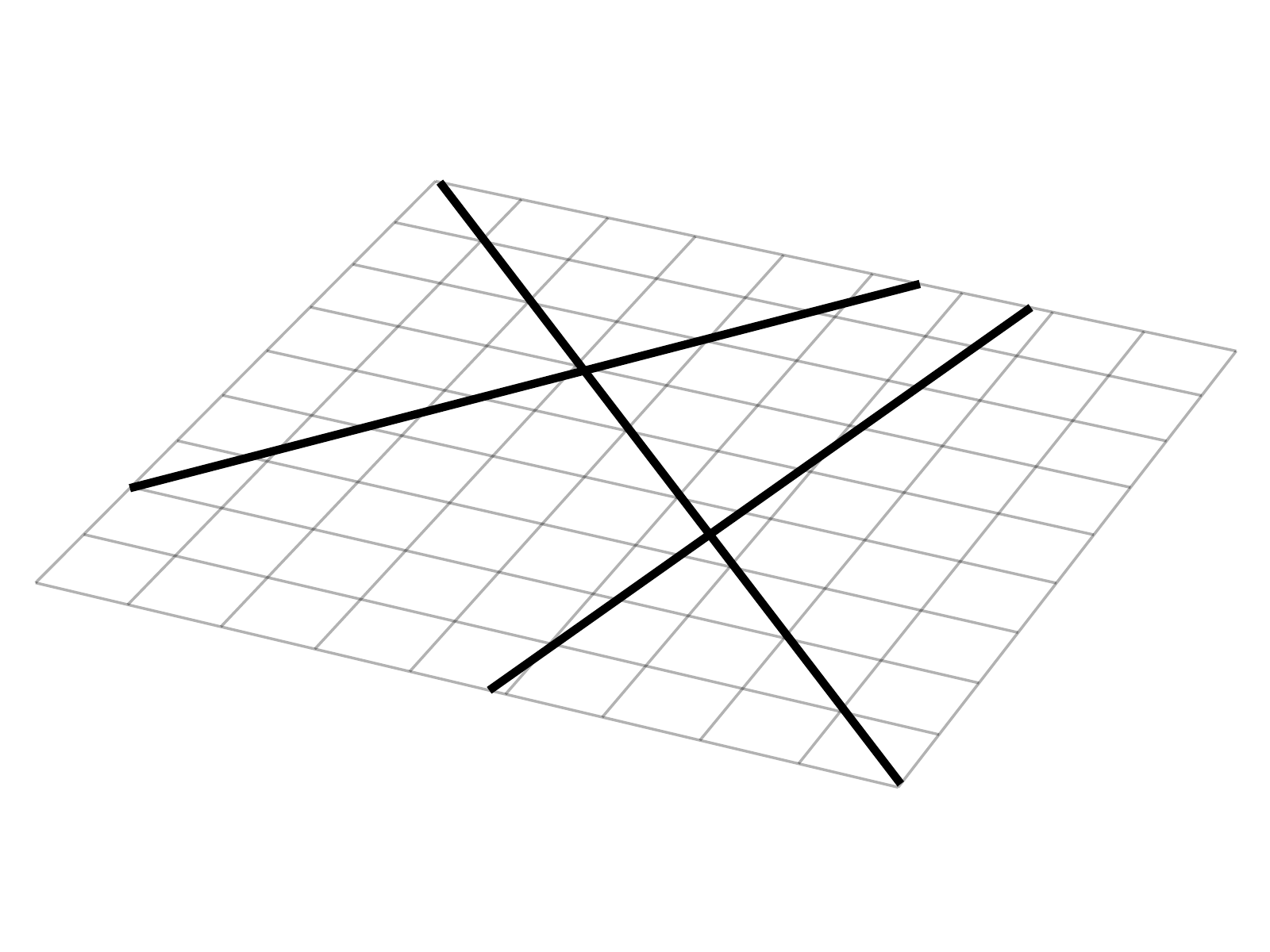}
    \includegraphics[width=0.3\linewidth]{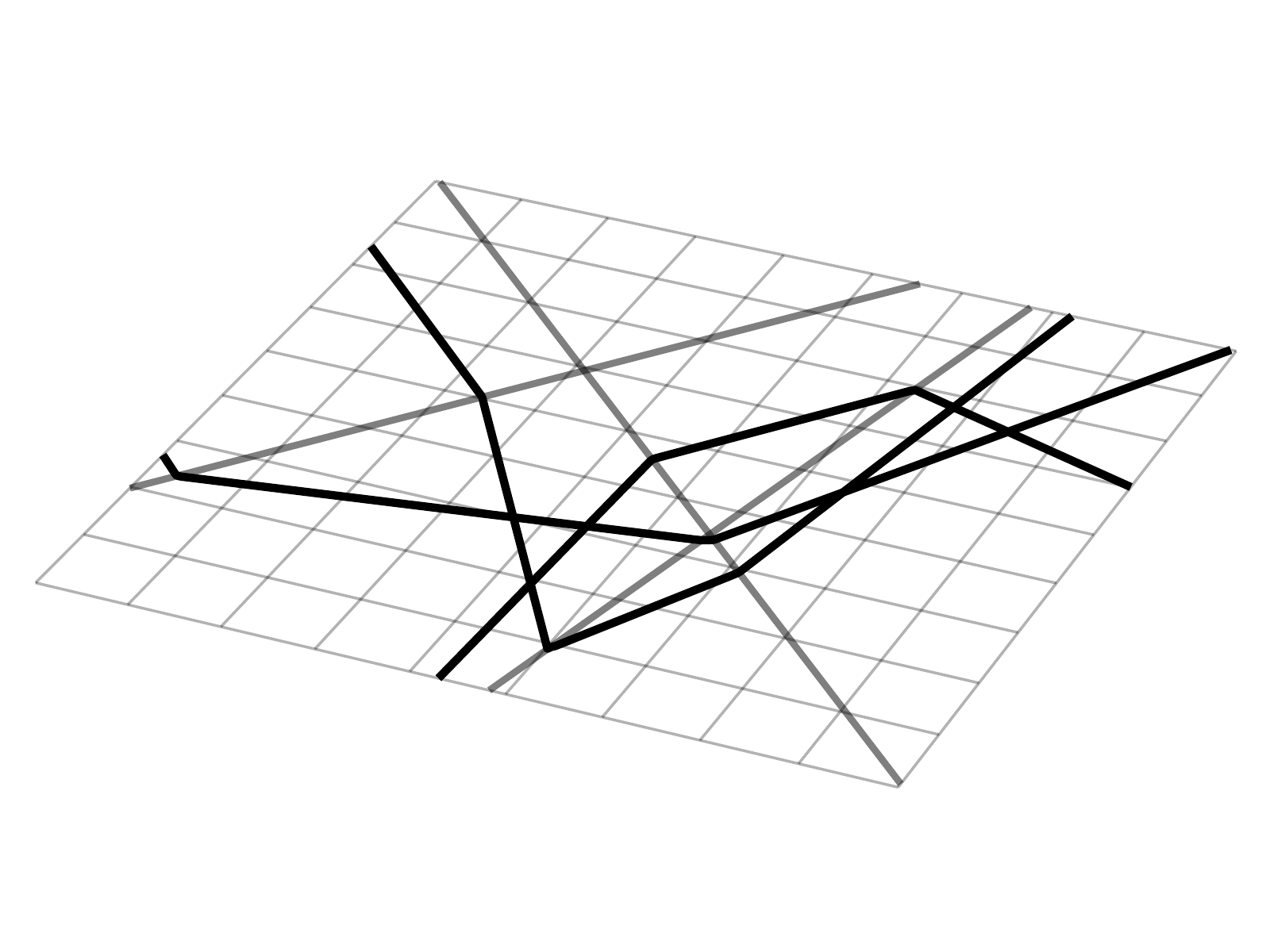}
    \includegraphics[width=0.3\linewidth]{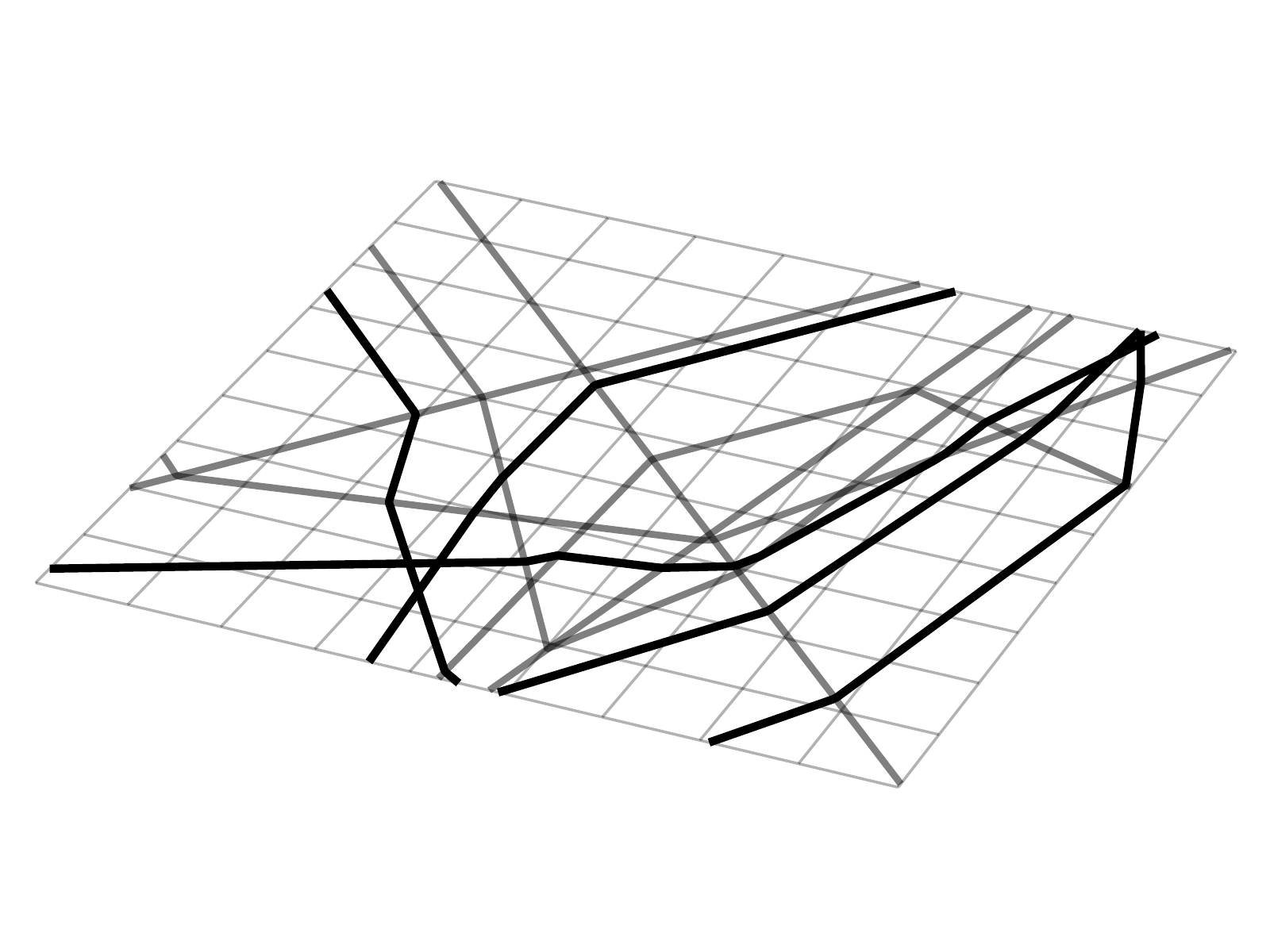}
    \caption{Iterative partitioning of input space by three consecutive layers
    of a toy deep network (DN). The first layer maps $\reals^2$ to $\reals^3$, producing three cuts
    and seven regions, with one lying outside the visualized region.
    }
    \label{fig:partitioning}
\end{figure}

The most popular DNs  
contain activation nonlinearities that are
continuous piecewise-affine (CPA), 
such as the ReLU, leaky-ReLU, absolute value, and max-pooling nonlinearities \cite{goodfellow2016deep}. 
To be more precise, let $f_{\LayerIndex}$
denote a layer of a DN. 
We will assume that it can be written as an affine map
$
    W_{\LayerIndex}{\inlayeri}+b_{\LayerIndex}
$
followed by the activation $\phi_{\LayerIndex}$
\begin{equation}\label{EQU:layer}
  {\outlayeri}
  =
  f_{\LayerIndex}({\inlayeri})
  =
  \phi_{\LayerIndex}(W_{\LayerIndex}{\inlayeri}+b_{\LayerIndex}).
\end{equation}
The entries of the matrix $W_{\LayerIndex}$ are called \emph{weights};
the constant additive term $b_{\LayerIndex}$ is called the \emph{bias}.
More general settings can easily be employed; however, the above will be sufficient for our purpose. The above-mentioned, commonly used nonlinearities can be
written in this form using
\begin{equation}\label{EQU:non_linearities}
  \phi_{\LayerIndex}(t)
  =
  \max(t,\eta t),
\end{equation}
where the max is  taken coordinate by coordinate.
Clearly, any activation of the form 
(\ref{EQU:non_linearities}) is continuous and piecewise linear,
rendering the corresponding layer CPA (Fig.~\ref{fig:partitioning}).
The following choices for $\eta$ correspond to
ReLU ($\eta=0$) \cite{glorot2011deep},
leaky-ReLU ($0<\eta<1$) \cite{DBLP:journals/corr/XuWCL15} and
absolute value \cite{bruna2013invariant}
($\eta=-1$)
(see also \cite{reportRB,balestriero2018spline}).
In summary:
\begin{equation}\label{EQU:non_linearities_eta}
  \left\{
    \begin{array}{ll}
     \eta>0  ~~~~~~& \mbox{leaky-ReLU} \\
      \eta=0  & \mbox{ReLU} \\
      \eta=-1 & \mbox{absolute value} \\
    \end{array}
    \right.
\end{equation}
For example, in two dimensions we obtain $\phi_{\LayerIndex}^{\rm (Abs)}(t_1,t_2)=(|t_1|,|t_2|)$.
We will consider only nonlinearities that can be written in the form
\equ{non_linearities}.
Such nonlinearities are in fact linear in each hyper-octant of space,
and can be written as
$$
    \phi_{\LayerIndex}(t)
    =
    D_{\LayerIndex}t,
$$
where $D_{\LayerIndex}$ is a diagonal matrix
with entries that depend on the signs
of the coordinates of the input $t=W_{\LayerIndex}{\inlayeri}+b_{\LayerIndex}$.
The diagonal entries of $D_{\LayerIndex}$ equal either $1$ or
 $\eta$ depending on the hyper-octant.

Consequently, the input space of the layer
$f_{\LayerIndex}$ is divided into \emph{regions}
that depend on $W_{\LayerIndex}$ and $b_{\LayerIndex}$
in which $f_{\LayerIndex}$ is affine
and can be rewritten as
\begin{equation}\label{EQU:locally_linear_layer}
  {\outlayeri}
  =
  f_{\LayerIndex}({\inlayeri})
  =
  \phi_{\LayerIndex}(W_{\LayerIndex}{\inlayeri}+b_{\LayerIndex})
  =
  (D_{\LayerIndex}W_{\LayerIndex}){\inlayeri}+D_{\LayerIndex}b_{\LayerIndex}.
\end{equation}
By the linear nature of ${\inlayeri}\mapsto W_{\LayerIndex}{\inlayeri}$,
these regions of the input space
(${\inlayeri}$-space) are piecewise linear.
Concatenating several layers into
a multi-layer $F({\inlayerone})=f_p\circ \ldots\circ  f_1({\inlayerone})$ then leads to a
subdivision of its input space into piecewise-linear regions (see Figure~\ref{fig:partitioning}) in which $F$
is affine, a result due to
\cite{balestriero2018spline,reportRB}.
In each linear region, this multi-layer
can be written as
\begin{equation}
    {\outlayerp}
    =
    D_pW_p\ldots D_1 W_1 {\inlayerone} +  B_p,
\end{equation}
where all constant terms are collected in
\begin{equation}
    B_p
    =
    D_pW_p\ldots D_2 W_2 b_1
    +
    D_pW_p\ldots D_3 W_3 b_2
    + \ldots +
    b_p.
\end{equation}
A multi-layer can be considered to be an entire DN or a part of one.

\subsection{From ConvNets to ResNets and DenseNets, and learning}
\label{subsec:Conv_vs_Res}

A DN with an architecture $F({\inlayerone})=f_p\circ \ldots\circ  f_1({\inlayerone})$
as described above is called a \emph{Convolutional Network}
(ConvNet) \cite{lecun1995convolutional} whenever the $W_k$ are circulant-block-circulant; otherwise it is called a multi-layer-perceptron. 
Without loss of generality, we will focus on ConvNets and their extensions, particularly ResNets and DenseNets, in this paper. 
While such architectures are responsible for the recent cresting wave of deep learning successes, they are known to be intricate to effectively employ when the number of layers $p$ large. This motivated research on  alternative architectures.

When adding the input ${\inlayeri}$ to the output $\phi_{\LayerIndex}(W_{\LayerIndex}{\inlayeri}+b_{\LayerIndex})$
of a convolutional layer (a skip connection), one obtains a \emph{residual layer}
which can be written in each region of linearity as
\begin{equation}\label{EQU:locally_residual_layer}
  {\outlayeri}
  =
  f_{\LayerIndex}({\inlayeri})
  =
  \phi_{\LayerIndex}(W_{\LayerIndex}{\inlayeri}+b_{\LayerIndex})+{\inlayeri}
  =
  (D_{\LayerIndex}W_{\LayerIndex}+{\rm Id}){\inlayeri}+D_{\LayerIndex}b_{\LayerIndex}.
\end{equation}
A DN in which all (or most) layers are residual is called a
\emph{Residual Network} (ResNet) \cite{he2016identity}. {\em Densely Connected Networks} (DenseNets) \cite{huang2017densely} allow the residual connection to bypass not just a single but multiple layers
\begin{equation}\label{EQU:locally_skip_residual}
  {\outlayerp}
  =
  (D_pW_p\ldots D_1W_1+{\rm Id}){\inlayerone}+B_p.
\end{equation}
All of the above DN architectures are universal approximators. Learning the per-layer weights of any DN architecture is performed by solving an optimization problem that involves a training data set, a differentiable loss function such as the squared-error or cross-entropy, and a weight update policy such as gradient descent. Hence, the key benefit of one architecture over another only emerges when considering the learning problem, and in particular gradient-based optimization. 

\subsection{Learning}
\label{subsec:learning}

We now consider the learning process for one particular layer of a DN.
As we make clear below, for the comparison of different network architectures, it is sufficient to study each layer independently of the others.
For ease of exposition, our analysis in Sections~\ref{subsec:learning}--\ref{subsec:cond_number} also focuses on a single training data point, which corresponds to learning via fully stochastic gradient descent ($G=1$ in the Introduction).
We extend our analysis to the general case of mini-batch gradient descent ($G>1$) in Section~\ref{subsec:batches}.

The layer being learned is $f_0$; however, for clarity we drop the index $0$ and use the notation 
\begin{equation}
    {{\outlayert}}=f(\inlayert)=\phi(W\inlayert+b)=DW\inlayert+b.
  \qquad\qquad\qquad\qquad\,\mbox{(trained layer)}
\end{equation}
The layers following $f$ are the only ones relevant in the study of learning layer $f$. In fact, the previous layers transform the DN input into a feature map that, for the layer in question, can be seen as a fixed input.
This part of the network will be denoted by
$F=f_p\circ \ldots\circ  f_1$.
In each partition region, it takes the form
\begin{eqnarray}
  \label{EQU:trained_layer}
    {\outDN}
    =
    MDW\inlayert+B.
  \qquad\qquad\qquad\,\mbox{(trained plus subsequent layers)}
\end{eqnarray}
Here, $B$ collects all of the constants, $D$ represents the
nonlinearity of the trained layer in the current region, 
and $M$ collect the action of the subsequent layers $F$. 
We use the parameter $\PertVar$ to distinguish between ConvNets and ResNets, with $\PertVar=0$ corresponding to convolution
layers and $\PertVar=1$ to residual ones. The matrix $M=M(\PertVar)$ representing the subsequent layers $F$
becomes
\begin{eqnarray}
  \label{EQU:training_one_layer}
   M=M(\PertVar) &=&   D_1W_1+\PertVar{\rm Id}
  ~\qquad\qquad\qquad\qquad\,\mbox{(one layer)}
  \\
  \label{EQU:training_multi_layer}
   M=M(\PertVar) &=&   D_pW_p\cdots D_1W_1+\PertVar{\rm Id}.
  \qquad\qquad\mbox{(multi-layer)}
\end{eqnarray}
As a shorthand, we will often write 
$M(\PertVar) = M_o+\PertVar{\rm Id}$, where
\begin{equation}
    \label{EQU:Mo}
    M_o
    =
    M(0)
    =
    D_pW_p\cdots D_1W_1.
\end{equation}
When saying ``let $M_o$ be arbitrary'' we mean that no restrictions on the factors in \equ{Mo} are imposed except for the $D_i$ to be diagonal and the $W_i$ to be square. 
When choosing $D_p=\cdots=D_1=W_p=\cdots=W_2={\rm Id}$
and $W_1$ arbitrary, we have $M_o=W_1$ arbitrary. So, there is no ambiguity in our usage of the term ``arbitrary''.

\subsection{Quadratic loss}
\label{subsec:quadr_loss}

We wish to quantify the advantage in training of a ResNet
architecture over a ConvNet one.
We consider a quadratic loss function such as the squared-error as mentioned in the Introduction (cf.\ (\ref{eq:loss}); we drop the superscript ${(g)}$):
\begin{equation}
    \label{eq:loss_squared_out}
    L^{(\dcounter)} :=
    \left\| y-\widehat{y} \right\|_2^2
    =
    (y-\widehat{y})^T(y-\widehat{y})
    =
    \left\| y \right\|_2^2
    -
    2y^T\widehat{y} 
    +  
    \left\|\widehat{y} \right\|_2^2.
\end{equation}
In the above, we used that 
$y^T\widehat{y}=\left(y^T\widehat{y}\right)^T=\widehat{y}^Ty$, since this is a real number.

Recalling that the DN is a continuous piecewise-affine operator and fixing all parameters except for $W$, 
we can locally write $\widehat{y}=MD W {\inlayert} +B$ 
for all $W$ in some region $\mathcal{R}$ of $F$
(cf.\ (\ref{eq:loss}); $\inlayert$ is the input to $f$ produced by the 
datum $\TrainData^{(g)}$).
Writing $\left\|\widehat{y} \right\|_2^2
=\widehat{y} ^T\widehat{y} $, we find
that the loss can be viewed as an 
affine term $P(W)$ with $P$ a polynomial of degree $1$
plus a quadratic term $\|MD W {\inlayert} \|_2^2$
\begin{eqnarray}
\label{EQU:quadr_term_MSE}
    L^{(\dcounter)}
    =
    P(W)+\|MD W {\inlayert} \|_2^2
    \qquad
    (W\in\mathcal{R})
    .
\end{eqnarray}
We stress once more 
that all parameters are fixed except for $W$.
Denote by $W_o=W_o(\PertVar,\mathcal{R})$ 
the minimum of the function $h(W)=P(W)+\|MD W {\inlayert} \|_2^2$
($W$ arbitrary),
which may very well lie outside $\mathcal{R}$.
Writing $\Delta W=W-W_o$ and
developing $h$ into its Taylor series at $W_o$,
we find 
$h(W)=h(W_o)+\|MD \Delta W {\inlayert} \|_2^2$,
since all first order derivatives of $h$ vanish at $W_o$.

It is worthwhile noting that the region $\mathcal{R}$ 
does not change when changing the value 
of $\PertVar$ in $M=M(\PertVar)$, since 
$D\Delta W{\inlayert}$ does not depend on $\PertVar$.
From this, writing $c=h(W_o)$ for short, we obtain for any $\PertVar$
\begin{eqnarray}
    \label{EQU:loss_centered_at_min}
    L^{(\dcounter)}
    =
    c +\|M(\PertVar)D \Delta W {\inlayert} \|_2^2
    \qquad
    (W\in\mathcal{R})
    .
\end{eqnarray}

This means that the loss surface is in each region
$\mathcal{R}$ of $F$ is a part of an exact ellipsoid 
whose shape is determined by the second-order expression
$\|M(\PertVar)D \Delta W {\inlayert} \|_2^2$ as a function of $\Delta W$. 
The value $c$ should not be interpreted as the minimum of the loss function, since $W_o$ may lie outside the region $\mathcal{R}$.

Before commenting further on the loss, however, let us untangle
the influence of the data ${\inlayert}$ (which remains unchanged during training) 
from
the weights $W$ in \equ{quadr_term_MSE} (analogous for $\Delta W$
in \equ{loss_centered_at_min}).
When listing the variables $W[i,j]$ row by row in a ``flattened''
vector $w$, the term 
\equ{quadr_term_MSE} 
becomes a quadratic form defined by the following matrix $Q$.

\begin{lem}
\label{lem:standard_quadratic_form}
In the quadratic term \equ{quadr_term_MSE} of the MSE, 
we have the following matrices:
$W$ is $n\times n$, $MD$ is $n\times n$, and ${\inlayert}$ is $n\times 1$.
Denote the flattened version of $W$ by $w$ ($n^2 \times 1$)
and the singular values of $MD$
by $\singval_i=\singval_i(MD)$.
Let
$
    Q
    =
    \text{\rm diag}({\inlayert}) \otimes (MD)
$
meaning that
\begin{equation}
\label{EQU:Q_block}
    Q =
    \begin{pmatrix}
        {\inlayert}[1] MD & 0           & \dots & 0\\
        0       & {\inlayert}[2] MD    & \dots &0 \\
               &         &  \ddots    &  \\
         0      &         \dots & 0 & {\inlayert}[n] MD
    \end{pmatrix}.
\end{equation}
Then, we have (cf.~\equ{quadr_term_MSE})
\begin{equation}
\label{EQU:flattened_M}
    \|MD W {\inlayert}\|_2^2
    =
    w^TQ^TQw,
\end{equation}
and that the singular values of $Q$ factor as
\begin{equation}
\label{EQU:sing_vals_Q}
\singval_{i,j}(Q)
=
\singval_i(MD) \cdot|{\inlayert}[j]| ,
\end{equation}
with $1 \leq i,j \leq n$.
\end{lem}

\begin{proof}
Direct computation easily
verifies \equ{flattened_M}.
For \equ{sing_vals_Q}, note that
the characteristic function 
of $\AnyM=Q^TQ$ is as follows, where in 
slight abuse of notation {\rm Id} denotes 
the identity in the appropriate dimensions
\begin{equation}
    \det(\AnyM-\eigval {\rm Id})
    =
    \prod_{j=1}^n
    \det\left(
        \left\{
            {\inlayert}[j]^2
            (MD)^T(MD)
        \right\}
            - \eigval {\rm Id}
        \right).
\end{equation}
Its zeros are easily identified as 
$\eigval_{i,j}
=
{\inlayert}[j]^2\eigval_i((MD)^T(MD))
=
{\inlayert}[j]^2\singval_i^2(MD)
$,
establishing \equ{sing_vals_Q}.
\end{proof}

\subsection{Condition number}
\label{subsec:cond_number}

We return now to the shape of the loss surface
of the trained layer $f$.
We point out that the loss surface is exactly 
equal to a portion of an ellipsoid 
whose geometry is completely 
characterized by the spectrum of the matrix $Q$ 
due to \equ{loss_centered_at_min} and
Lemma~\ref{lem:standard_quadratic_form}.
Its curvature and higher-dimensional eccentricity
has a considerable influence on the numerical behavior of gradient-based optimization. 
In essence, the larger the eccentricity, the less accurately 
the gradient points towards the maximal change of the loss 
when making a non-infinitesimal change in $W$.

The condition number $\CondNb$ of the matrix $Q$ provides a measure of its
higher-dimensional eccentricity and, therefore,
of the difficulty of optimization by linear least squares.
It is defined as
the ratio of the largest singular value $\singval_1$
to the smallest non-zero singular value. Denoting 
the latter by $\singval_*$ for convenience, the
condition number of
an arbitrary matrix $A$ is given by
\begin{equation}\label{EQU:def_conditioning}
  \CondNb(A)
  :=
  \frac{\singval_1(A)}{
  \min\{\singval_i(A)\,:\,\singval_i(A)>0\}}
  =
  \frac{\singval_1(A)}{
  \min_i^* \singval_i(A)
  }=
  \frac{\singval_1(A)}{\singval_*(A)}.
\end{equation}

Having identified $\CondNb(Q)$ as a most relevant
quantity towards understanding the performance of
linear least squares, we can apply 
Lemma~\ref{lem:standard_quadratic_form} 
to identify and separate the influence
of data and subsequent layers on 
the loss landscape as follows.
For an efficient notation we 
recall that the singular values of 
${\rm diag}({\inlayert})$ are $|z[j]|$
and that vanishing $z[j]$ do not
contribute to 
        $\CondNb(\mbox{\rm diag}({\inlayert}))$
        (cf.~\equ{def_conditioning}).

\begin{cor}[The Data Factor]
\label{cor:standard_quadratic_form}
    The condition number of the matrix $Q$ 
    of Lemma~\ref{lem:standard_quadratic_form} 
    factors as
    \begin{equation}
        \label{EQU:data_factor}
        \CondNb(Q)
        =
        \CondNb(MD)
        \cdot
        \CondNb(\mbox{\rm diag}({\inlayert})).
    \end{equation}
 
\end{cor}

Our main interest lies in the impact of the architecture of the $p$ layers 
following $f$ (adding a residual link or not), leaving the rest of the DN
as is. 
Since the input data ${\inlayert}$ does not depend on this architectural choice, the performance of the linear least-squares optimization we focus on here 
is governed by $\CondNb(MD)=\CondNb(M(\PertVar)D)$.

Consequently, we are faced
with two types of perturbation settings: 
First, the evolution of the loss landscape of DNs 
as their architecture is continuously interpolated from a ConvNet ($\PertVar=0$) to a ResNet ($\PertVar=1$), that is, when moving from
$M(0)$ to $M(1)$. 
In this case, since the region $\mathcal{R}$ of $F$ does not depend on $\PertVar$, such an analysis is meaningful for understanding the optimization of the loss via 
\equ{loss_centered_at_min}.
Second, the effect of multiplying $M$
by a diagonal matrix $D$.
Before studying the perturbations in the 
subsequent sections, we make a remark regarding the use of mini-batches.

\subsection{Learning using mini-batches}
\label{subsec:batches}

In deep learning, one works typically with mini-batches of data ${\TrainData}^{(g)}$
($g=1,\ldots, G$) that produce the inputs $\inlayert^{(g)}$ to the trained layer $f$. 
The loss of a mini-batch is obtained by averaging the losses of the individual data points (cf.\ (\ref{eq:minibatch}), \equ{quadr_term_MSE} and \equ{flattened_M}).
Using the superscript ${(g)}$ in a self-explanatory way and letting $w$ be as in Lemma~\ref{lem:standard_quadratic_form}, we have 
\begin{equation}
    \label{EQU:loss_batch}
    \Lbatch
    =
    \frac 1 G
     \sum_{g=1}^G  P^{(g)}(W)
     +
    \frac 1 G
     \sum_{g=1}^G  \| Q^{(g)} w \|_2^2,
     \qquad\qquad
  W\in\bigcap_{g=1}^G\mathcal{R}^{(g)}.
\end{equation}

Letting ${\inlayert}^{(g)} [k]$ denote the $k$-th coordinate of the input
to the trained layer produced by the $g$-th datum 
${\TrainData}^{(g)}$ we can rewrite the mini-batch loss
in a form analogous to \equ{quadr_term_MSE}, which enables us
to extend our analysis 
of the shape of the loss surface  to mini-batches.

\begin{lem}
  \label{lem:loss_batch_form}
  There exists a linear polynomial $\Pbatch$ and a symmetric matrix 
  $\Qbatch$ such that 
\begin{equation}
    \label{EQU:loss_batch_form}
    \Lbatch
    =
    \Pbatch(W)
     +
     w^T \Qbatch^T \Qbatch w. 
\end{equation}
The singular values of $\Qbatch$ 
can be bounded as follows:
\begin{eqnarray}
    \label{EQU:batch_singval_upper_bound}
    \singval_{i,j}^2(\Qbatch)
    &\leq&
    \frac 1 G \sum_{g=1}^G  
    |{\inlayert}^{(g)} [j]|^2
    \singval_1^2 (   M^{(g)}D^{(g)}  ),
    \\
    \label{EQU:batch_singval_lower_bound}
    \singval_{i,j}^2(\Qbatch)
    &\geq&
    \frac 1 G \sum_{g=1}^G  
    |{\inlayert}^{(g)} [j]|^2
    \singval_n^2(   M^{(g)}D^{(g)}  ) 
    .
\end{eqnarray}
Notably, 
for any $j$ with 
${\inlayert}^{(g)} [j]=0$ $(1\leq g \leq G)$,
we have that
$
\singval_{i,j}(\Qbatch)=0
$ for $1 \leq i \leq n$.

\end{lem}

\begin{proof}
The linear polynomial is obviously
$
    \Pbatch(W)
    =
    %\frac 1 G
    ( 1 / G )
     \sum_{g=1}^G  P^{(g)}(W)
.$
To clarify the choice of $\Qbatch$, set 
$
    \AnyM
    =
    %\frac 1 G
    ( 1 / G )
    \sum_{g=1}^G  (Q^{(g)})^TQ^{(g)}
.$
Then, 
$
    \Lbatch
    =
    \Pbatch(W)
    +
    w^T \AnyM w
$,
and $A$ is symmetric and positive semi-definite.

Notably, the matrix $\AnyM$ inherits the diagonal 
block structure 
of the $Q^{(g)}$ as given in \equ{Q_block}.
In a slight abuse of notation, we can write $\AnyM={\rm diag}(\AnyM_1\ldots,\AnyM_n)$,  where 
the $j$-th block is a symmetric $n\times n$ matrix
\begin{eqnarray}
    \label{EQU:Aj_blockform}
    \AnyM_j
    =
    \frac 1 G
    %( 1 / G )
    \sum_{g=1}^G  
    {\inlayert}^{(g)} [j]^2
    ( M^{(g)}D^{(g)} )^T( M^{(g)}D^{(g)} )
.
\end{eqnarray}
Following the argumentation of the proof of Lemma~\ref{lem:standard_quadratic_form}, meaning
that we exploit the fact that the blocks do not interact, it becomes apparent that the eigenvalues of $\AnyM$ are those of the blocks 
$\AnyM_j$. In short, we can list the eigenvalues
of $\AnyM$ as
$
    \eigval_{i,j}(\AnyM)
    =
    \eigval_{i}(\AnyM_j)
     \geq 0,
$
where we arrange eigenvalues  
by size in descending order within each block.
Arranging these eigenvalues by block into a diagonal
matrix $\Lambda$,
there is an orthogonal matrix $U$ such that 
$\AnyM= U^T \Lambda U$. 
Now arranging their square roots 
$\sqrt{\eigval_{i,j}(\AnyM)}$ in the same order 
into the diagonal matrix $S
$,
we can set $\Qbatch=U^TSU$.
Then, 
$  \AnyM   = \Qbatch^T \Qbatch$ and 
$  w^T \AnyM w  = \| \Qbatch w \|_2^2$ as desired.
While we could choose $\Qbatch$ to be any matrix
of the form $V^TSU$ with $V$ orthogonal and find 
identical singular values,
we chose a symmetric matrix  for reasons of simplicity and convenience.

Clearly, 
$
    \singval_{i,j}^2(\Qbatch)
    =
    \eigval_{i,j}(\AnyM)
    =
    \eigval_{i}(\AnyM_j)
.$
For the upper bound of these singular values,
we  proceed as follows
\begin{equation}
    \singval_{i,j}^2(\Qbatch)
    =
    \eigval_{i}(\AnyM_j)
    \leq  
        \|{\AnyM_j}\|_{\rm Op}
    \leq  
    \frac 1 G
    %( 1 / G )
    \sum_{g=1}^G  
    {\inlayert}^{(g)} [j]^2
    ~
    \| M^{(g)}D^{(g)} \|_{\rm Op}^2.
\end{equation}
From this, \equ{batch_singval_upper_bound} 
follows easily. For the lower bound
\equ{batch_singval_lower_bound}
let $u$ denote the normalized eigenvector 
to the smallest
eigenvalue $\eigval_{n,j}$ of 
$\AnyM_j$. 
Using \equ{Aj_blockform} we obtain
\begin{equation}
    \singval_{i,j}^2(\Qbatch)
    =
    \eigval_{i}(\AnyM_j)
    \geq  
    \eigval_{n}(\AnyM_j)
    =
    u^T \AnyM_j u
    =
    \frac 1 G
    \sum_{g=1}^G  
    {\inlayert}^{(g)} [j]^2
    ~
    \| ( M^{(g)}D^{(g)} )u \|_2^2.
\end{equation}
From this, using
$\| ( M^{(g)}D^{(g)} )u \|_2
\geq  
\singval_n( M^{(g)}D^{(g)} ) \|u \|_2
=
\singval_n( M^{(g)}D^{(g)} ) 
$,
the bound \equ{batch_singval_lower_bound} follows.
\end{proof}

For a convenient bound of the condition number
of $\Qbatch$, we set
$\overline{\inlayert} [j]
=
\sqrt{(1/G)
    \sum_{g=1}^G  
    |{\inlayert}^{(g)} [j]|^2
    }
$,
which can be interpreted as the $j$-th
mean square feature of the batch.
From Lemma~\ref{lem:loss_batch_form} we can conclude
that $\overline{\inlayert} [j]=0$ implies
$\singval_{i,j}(\Qbatch)=0$ for all $i$.
Since vanishing singular values are disregarded
in the computation of the condition number 
as defined in this paper, we obtain the following 
result.
\begin{cor}
  \label{cor:loss_batch_form}
  In the notation of  Lemma \ref{lem:loss_batch_form}, assume that
  there are constants $C$ and $c>0$ such that, 
  for  $1\leq g \leq G$,
\begin{equation}
\singval_1( M^{(g)}D^{(g)} ) 
\leq 
C
\qquad {\rm and} \qquad
\singval_n( M^{(g)}D^{(g)} ) 
\geq
c.
\end{equation}
Assume that there is at least
one {non-vanishing} $\overline{\inlayert} [j]$,
and denote by 
$\min_j^*\overline{\inlayert} [j]$
the minimum over all \emph{non-vanishing}
$\overline{\inlayert} [j]$.
Then
\begin{equation}
\CondNb(\Qbatch)  \leq    \frac C c   
    \cdot    
    \frac{  
        \max_j\overline{\inlayert} [j]
    }{   
    \min_j^*\overline{\inlayert} [j]
    }.
\end{equation}
\end{cor}

\section{Perturbation of a DN's Loss Landscape When Moving from a ConvNet to a ResNet}
\label{sec:AddingIdentity}

In this section we exploit differentiability of the family  
$M(\PertVar)$ in order to study the perturbation of
its singular values when moving from
ConvNets with $\PertVar=0$  to ResNets with $\PertVar=1$    (see Section~\ref{subsec:Conv_vs_Res}).

We start with a general result that will prove most useful.

\begin{lem}
    \label{lem:discriminant}
    Let $N(\PertVar)$ be a family of symmetric $n\times n$ matrices
    with  entries that are polynomials in $\PertVar$.
    Let $\Mult$ 
    denote the set of $\PertVar\in\reals$ 
    for which $N(\PertVar)$ has multiple eigenvalues.
    Then the following dichotomy holds: 
    \begin{itemize}
        \item 
        \label{dicho_A}
        Either $\Mult$ is finite
            or $\Mult=\reals$.
    \end{itemize}
Also, 
if the entries of $N$ are real analytical functions of $\PertVar$ 
in some open set containing the compact interval $[\PertVar_1,\PertVar_2]$,
then the set of $\PertVar\in[\PertVar_1,\PertVar_2]$
with multiple eigenvalues of $N(\PertVar)$ is either finite or all of $[\PertVar_1,\PertVar_2]$.
\end{lem}

Clearly, this result extends easily to the multiplicity
the singular values of any family 
of square matrices $\AnyM(\PertVar)$ 
by considering $N(\PertVar)=\AnyM^T\AnyM$.

\begin{proof}
Fixing $\PertVar$ for a moment,
the eigenvalues of $N(\PertVar)$ are the zeros of the 
characteristic polynomial $h(\eigval)=\det(N(\PertVar)-\eigval\,{\rm Id})$.
This polynomial $h$ has multiple zeros $\eigval$ 
if and only if its discriminant  
is zero. It is well known that the
discriminant $\discr_h$ of a polynomial $h$ 
is itself a polynomial in the coefficients of $h$.

If the entries of $N$ are  polynomials in $\PertVar$, then
the discriminant of $h$ is, as a function
of $\PertVar$, in fact a polynomial $\discr_h(\PertVar)$.
In other words, the set 
$\Mult$ is the set of zeros of the polynomial $\discr_h(\PertVar)$. This set is finite unless the
polynomial $\discr_h(\PertVar)$ vanishes identically, in which case $\Mult$ becomes all of $\reals$.

Finally, if the entries of $N$ are 
real analytical, then so is 
$\discr_h(\PertVar)$, and
the identity theorem implies the claim.
\end{proof}

Combining with Lemma~\ref{lem:continuity} and
Theorem~\ref{thm:lambda_differentiability}
we obtain the following. 

\begin{cor}
    \label{cor:discriminant}
    Let $N$ be as in Lemma~\ref{lem:discriminant}.
    In the real analytical case, assume that $\PertVar_1=0$
    and $\PertVar_2=\PertVar$ for some fixed $\PertVar$.
Assume that $N(0)$ has no multiple eigenvalues.
Then $\Mult$ is finite. Thus, for any $i$
the set $[0,\PertVar]\setminus S_i$ is finite and
\begin{equation}
    \label{EQU:FundThmCalc}
    \eigval_i(\PertVar)
    -\eigval_i(0)
    =
    \int_0^\PertVar \eigval'_i(t) dt
    =
    \int_0^\PertVar 
        u_i(t)^T N'(t) u_i(t) dt.
\end{equation}
\end{cor}

After these general remarks, let us now consider 
$M(\PertVar)=M_o+\PertVar$\,{\rm Id}
\equ{Mo} 
and use \equ{FundThmCalc}  to bound the growth of the singular values of $M$. 
To this end, we note the following.

\MoreDetails{
        A simple result concerns  with the sum of the eigenvalues
        $\eigval_i$ of $N=M^TM$ or equivalently the sum of squared singular values of $M$:
        \begin{equation}
            \label{EQU:def_sum_eigenvalues}
            \sumSQsing(\PertVar)
            :=
            \sum_{i=1}^n \eigval_i(\PertVar)
            =
            \sum_{i=1}^n \singval^2_i(\PertVar)
            =
            {\rm  tr}(N(\PertVar))
            =
            \|M(\PertVar)\|_F^2
            .
        \end{equation}
        Since the trace is the sum of the diagonal entries of $N$, a simple direct computation yields immediately the following result.
        
        \begin{lem}
            \label{lem:trace}
            Let $M_o$ be arbitrary (see \equ{Mo}).
            Then, $M'=$Id and 
            %$\mathcal{C}^{1}(\reals)$ 
            \begin{equation}
                \label{EQU:derivative_trace}
                \frac{d}{d\PertVar} \sumSQsing(\PertVar)
                =
                \sumSQsing'(\PertVar)
                =
                %\sum_i \eigval_i'(\PertVar)=
                {\rm tr}(N'(\PertVar))
                %={\rm tr}({M'}^T M + M^T M')(\PertVar)
                =
                2{\rm tr}({M'}^T M)
                =
              2{\rm tr}(M_o)+2n\PertVar
                .
            \end{equation}
        \end{lem}
}

\begin{lem}[Derivative]
  \label{lem:derivative_lambda_plusId}
    Let $M_o$ be arbitrary (see \equ{Mo}).
Writing  $u_i=u_i(\PertVar)$ for short, the derivative of the  eigenvalue $\eigval_i(\PertVar)$ of $M^TM$ becomes, on $S_i$ (see \equ{def_simple})
\begin{equation}
    \label{EQU:derivative_lambda_plusId}
        \eigval_i'(\PertVar)
        =
        u_i^T(M_o^T+M_o) u_i+2\PertVar
        =
        2u_i(\PertVar)^TM_o u_i(\PertVar)+2\PertVar.
\end{equation}
On $S_i^*$ (see \equ{def_nonzero}) we have
\begin{equation}
    \label{EQU:derivative_sigma_plusId}
        \singval_i'(\PertVar)
        =
        \eigval_i'(\PertVar)/(2\singval_i)
        =
        u_i^T v_i
        =
        \cos(\alpha_i(\PertVar))
        ,
\end{equation}
where $\alpha_i(\PertVar)$ denotes the angle between the $i$-th left and right singular vectors.
\end{lem}

Using \equ{derivative_sigma_plusId} it is possible 
to compute the derivative
$\singval_i'(\PertVar)$ numerically
with ease.

\begin{proof}
Apply 
Theorem~\ref{thm:lambda_differentiability}
to $N(\PertVar)=M^TM$.
Use
$N'=(M')^TM+M^T(M')=M+M^T=M_o^T+M_o+2$Id
and $u_i^Tu_i=1$ to obtain
\begin{equation}
        \eigval_i'(\PertVar)
        =
        u_i^T(M^T+M) u_i
        =
        u_i^T(M_o^T+M_o) u_i+2\PertVar.
\end{equation}
Also, note that $u_i^TM_o^Tu_i=(u_i^TM_o^Tu_i)^T
=u_i^TM_ou_i$, since this is a real-valued number. 
This establishes \equ{derivative_lambda_plusId}.
In a similar way
\begin{equation}  
    \eigval_i'(\PertVar)
    =
    u_i^T(M^T+M) u_i=2(u_i^TM u_i)=2\singval_i u_i^Tv_i
.
\end{equation}
The chain rule implies
\equ{derivative_sigma_plusId}.
Alternatively, \equ{derivative_sigma_plusId}
follows from Corollary~\ref{cor:sigma_differentiability}.
~\end{proof}

\begin{defi}
\label{def:tau}
For a family of matrices of the form $M(\PertVar)=M_o+\PertVar\,{\rm Id}$,
denote the smallest eigenvalue of the symmetric matrix $M_o^T+M_o$ by $\underline{\tau}$ and
its largest by $\overline{\tau}$.
\end{defi}
Note that, due to
$\underline{\tau}\leq \overline{\tau}$,
we have
\begin{equation}\label{EQU:OpNorm_MT+M}
  \|M_o^T+M_o\|_{\rm Op}
  =
  \max(
  |\underline{\tau}|,|\overline{\tau}|
    )
  =
  \max(
  -\underline{\tau},\overline{\tau}
    )
  \leq
  2\singval_1(M_o)
  .
\end{equation}
Using 
Corollary~\ref{cor:discriminant}
with $N=M^TM$
and \equ{derivative_lambda_plusId} from Lemma~\ref{lem:derivative_lambda_plusId},
we have the following immediate result.

\begin{thm}
\label{thm:lambda_bounds}
Assume that $M_o$ has no multiple singular values
but is otherwise  arbitrary (see \equ{Mo}). 
Then, for all $i$,
        \begin{equation}
            \label{EQU:lambda_increase_bounds}
            \PertVar\underline\tau+\PertVar^2
            \leq
            \singval_i(\PertVar)^2-\singval_i(0)^2
            \leq
            \PertVar\overline\tau+\PertVar^2.
        \end{equation}
\end{thm}

We mention that a random matrix 
with iid uniform or Gaussian entries has 
almost surely no 
multiple singular values (see
%\cite[Sect. 6.2.3.]{Chafai} or %\cite{wishart1928generalised}
Theorem~\ref{thm:Mo_simple}).

We continue with some observations 
on the growth of singular values that
leverage the simple form of $M(\PertVar)$ 
(see \equ{Mo}) explicitly
and that apply even in the case of $M_o$ having 
multiple zeros.

Obviously, \equ{lambda_increase_bounds}
holds also if for any $\PertVar$ the matrix
$M(\PertVar)$ is known to have no multiple singular values.
Even without any such knowledge, we have
the following  useful consequence of Weyl's additive perturbation bound.

\begin{lem}
    \label{lem:Weyl_bound}
    Let $M_o$ be arbitrary (see \equ{Mo}).
    Then we have
\begin{equation}
    \label{EQU:Weyl_bound}
    \PertVar-\singval_1(M_o)
    \leq
    \singval_i(M(\PertVar))
    \leq
    \PertVar+\singval_1(M_o).
\end{equation}  
\end{lem}

\begin{proof}
Apply \equ{uniform_bound} 
with $\AnyM=M(\PertVar)=M_o+\PertVar\,{\rm Id}$
and $\AnyMt=\PertVar\,{\rm Id}$ to obtain
\begin{equation}
    |\singval_i(M(\PertVar))-\PertVar|
    \leq
    \singval_1(\AnyM-\AnyMt)
    =
    \singval_1(M_o)
    .
\end{equation}
\end{proof}
For a comparison, note that 
\equ{lambda_increase_bounds} implies that for any $i$ 
\begin{equation}\label{EQU:max_value_stronger}
    \sqrt{\singval_i^2(M_o)
            +
    \PertVar\underline{\tau}  
            +\PertVar^2}
    \leq
    \singval_i(\PertVar)
    \leq
    \sqrt{\singval_i^2(M_o)
            +
    \PertVar\overline{\tau}  
            +\PertVar^2}
\end{equation}
whenever it holds.
Since $\overline\tau\leq 2 \singval_1(M_o)$         
and $\singval_i(0)\leq \singval_1(0)$,
the upper  bound is clearly sharper than
that of \equ{Weyl_bound}.

%end more details on the largest eigenvalue

Continuing our observations,
the following fact is obvious
due to the simple form of $M(\PertVar)$.

\begin{lem}[Eigenvector]
  \label{lem:eigenvector}
    Let $M_o$ be arbitrary (see \equ{Mo}).
  Assume that, for some $\PertVar_o$,
  the vector $w=u_i(\PertVar_o)$ is 
  a right singular vector as well as an eigenvector of $M(\PertVar_o)$  with eigenvalue $t$.
  Then, for all $\PertVar$, the vector $w$ is an eigenvector and right singular vector of $M(\PertVar)$ with eigenvalue $t+\PertVar-\PertVar_o$
  and singular value $|t+\PertVar-\PertVar_o|$.
\end{lem}
Note that the singular value of the fixed vector $w$
may change order, as other singular values may increase or decrease at
a different speed.

Further, 
we have that $|\singval_i'(\PertVar)|\leq1$ on $S_i^*$ which is obvious from  geometry
and 
from \equ{derivative_sigma_plusId}.
The eigenvectors of $M_o$, if they exist,
achieve extreme growth or decay.
\begin{lem}[Maximum Growth]
  \label{lem:max_growth}
    Let $M_o$ be arbitrary (see \equ{Mo}).
  Assume that
  %$\alpha_i(\PertVar_o)=0$
  $\singval'_i(\PertVar_o)=1$
  for some $\PertVar_o$.
  Then, $w=u_i(\PertVar_o)$ is eigenvector and right singular vector of $M(\PertVar)$
  with eigenvalue $t+\PertVar-\PertVar_o$ and singular value $|t+\PertVar-\PertVar_o|$
  for $t=\singval_i(\PertVar_o)>0$.
  % and all $\PertVar$.
%
  The analogous results holds for
%  $\alpha_i(\PertVar_o)=\pi$
  $\singval'_i(\PertVar_o)=-1$
with $t=-\singval_i(\PertVar_o)<0$.
\end{lem}
%@@@ double text under PrintAll
Note that the singular value of the fixed vector $w$
may change order, since other singular values may increase or decrease at
a different speed.

Of interest are also the zeros of the singular values
since differentiability may fail there (c.f.~Lemma~\ref{lem:derivative_lambda_plusId}).

\begin{lem}[Non-Vanishing Singular Values]
  \label{lem:vanishing_sing_value}
  Let $M=M_o+\PertVar{\rm Id}$.
  Assume that $\singval_i(\PertVar_o)=0$ for some $\PertVar_o$.
  Then, $w=u_i(\PertVar_o)$ is an eigenvector and right singular vector of $M(\PertVar)$ for all 
  $\PertVar$
  with eigenvalue $\PertVar-\PertVar_o$ and singular value $|\PertVar-\PertVar_o|$.
  Consequently, if $M_o$ has no eigenvectors, then $\singval_i(\PertVar)\neq 0$
  for all $\PertVar$ and $i$.
  % and $\min\{\singval_i(\PertVar)\,:\,0\leq\PertVar \leq 1\}>0$,  for all $i$.
\end{lem}

Theorem~\ref{thm:lambda_bounds} shows that 
under mild conditions all 
singular values should  be expected to increase
when adding a multiple of the identity. 
However, when concerned only with the condition number,
the following result suffices, requiring no technical assumptions at all.

\begin{lem}
  \label{lem:wasem}
    Let $M_o$ be arbitrary (see \equ{Mo}).
    Then
    \begin{equation}
        \label{EQU:wasem}
            \singval_n(0)^2+\PertVar\underline\tau+\PertVar^2
            \leq
            \singval_n(\PertVar)^2
            \leq
            \singval_1(\PertVar)^2
            \leq
            \singval_1(0)^2+\PertVar\overline\tau+\PertVar^2.
    \end{equation}
\end{lem}

\begin{proof}
The argument consists of a direct computation, leveraging the special form of $M(\PertVar)=M_o+\PertVar{\rm Id}$.
For simplicity, let us denote the normalized  
eigenvector of 
$N(\PertVar)=M^TM$ with eigenvalue 
$\eigval_i(\PertVar)=\singval_i(\PertVar)^2$ 
by $u=u_i(\PertVar)$.
Then,
\begin{eqnarray}
    \nonumber
    \eigval_i(\PertVar)
    =
    u^TN(\PertVar)u
    &=&
    u^T(M_o^TM_o+\PertVar(M_o^T+M_o)+\PertVar^2{\rm Id})u
    \\    \nonumber
    &=&
    u^T(M_o^TM_o)u +\PertVar u^T(M_o^T+M_o)u+\PertVar^2
    \\
%    \nonumber
    &\leq&
     \eigval_1(0)+\PertVar\overline{\tau}+\PertVar^2.
\end{eqnarray}
Similarly, we can estimate the second to last expression
from below by
$\eigval_n(0)+\PertVar\underline{\tau}+\PertVar^2
$.
Since $i$ is arbitrary, the claim follows.
~\end{proof}

The advantage of \equ{Weyl_bound} is its simple form,
while \equ{wasem} is sharper (cf.~\equ{max_value_stronger}).
        As we will see, certain random matrices of interest 
        tend to achieve the  upper bound 
        \equ{wasem} 
        quite closely with high probability.
Combining this observation with
Lemmata~\ref{lem:Weyl_bound}
and~\ref{lem:wasem},
we obtain the following.

\begin{thm}
\label{thm:chi_bounds}
Let $M_o$ be arbitrary (see \equ{Mo}).
Assume that
$\singval_1(M_o)< 1$.
Then
\begin{equation}
    \label{EQU:cond_bound_via_OpNorm}
    \CondNb(M_o+{\rm Id})
    \leq
    \frac{\sqrt{1+\overline\tau+\singval_1(M_o)^2}}{1-\singval_1(M_o)}
    \leq
    \frac{1+\singval_1(M_o)}{{1-\singval_1(M_o)}}.
\end{equation}
Alternatively, assume that
$\underline{\tau} > -1$.
Then
\begin{equation}
    \label{EQU:cond_bound_via_Tau}
    \CondNb(M_o+{\rm Id})
    \leq
    \sqrt{\frac{1+\overline\tau+\singval_1(M_o)^2}{1+\underline\tau+\singval_n(M_o)^2}}
    \leq
    \frac{1+\singval_1(M_o)}{\sqrt{1+\underline\tau}}.
\end{equation}
\end{thm}
We list all four bounds due to their different advantages.
Due to \equ{OpNorm_MT+M}, the first bounds in both
\equ{cond_bound_via_OpNorm} as well as \equ{cond_bound_via_Tau}
are in general sharper than the second. 
Indeed, for random  matrices as used in Section~\ref{sec:probabilities}
we have typically 
$\overline\tau\simeq\sqrt{2}\singval_1(M_o)
<2\singval_1(M_o)$.
The second bounds are simpler and  sufficient in certain
deterministic settings.

The first bound of \equ{cond_bound_via_Tau} 
is the tightest
of all four bounds for the random matrices of interest 
in this paper. 
On the other hand, since we typically have 
$-\underline\tau\simeq\sqrt{2}\singval_1(M_o)$
for such matrices,
the stated sufficient condition for 
\equ{cond_bound_via_OpNorm} is somewhat easier to meet than that of \equ{cond_bound_via_Tau}.

\begin{proof}
From Lemmata~\ref{lem:wasem} and~\ref{lem:Weyl_bound}
we obtain the first inequality of \equ{cond_bound_via_OpNorm};
the second part follows from 
$\overline\tau\leq 2\singval_1(0)$.
Using Lemma~\ref{lem:wasem} we obtain \equ{cond_bound_via_Tau}.
\MoreDetails{
            Simple algebra yields
            $$
                \singval_n^2(0)
                +
                \underline{\tau}  
                +1
                \geq
                \singval_n^2(0)
                +
                (\overline\tau+1)/\CondNb(M_o)^2
                =
                \left[\singval_1^2(0)
                +
                (\overline\tau+1)\right]
                /\CondNb(M_o)^2
            $$
            The first inequality of \equ{cond_bound_via_Tau} completes the proof.
}
%~
\end{proof}

\section{Influence of the Activation Nonlinearity on a Deep Network's Loss Landscape }
\label{sec:PerturbationDiagonal}

As we put forward in Section~\ref{sec:DeepLearning}, each layer of a DN can be represented locally as a product of matrices $D_iW_i$, where 
$W_i$ contains the weights of the layer while the diagonal
$D_i$ 
with entries equal to $\eta$ or $1$ (compare \equ{non_linearities_eta} and \equ{locally_linear_layer})
captures the effect of the activation nonlinearity.

In the last section we explored how the singular values of a matrix $M_o$
change when adding a fraction of the identity, i.e., when
moving from a ConvNet to a ResNet.
In this section we explore the effect of the activation nonlinearities on the singular values.
To this end, let $\mathcal D(m,n,\eta)$ denote the set of all $n \times n$ diagonal matrices
with exactly $m$ diagonal entries equal to $\eta$ and $n-m$ entries equal to $1$:
    \begin{eqnarray}\label{EQU:Dk(eta)}
        \!\!\!\!\!
        \mathcal{D}(m,n,\eta)=
        \{D={\rm diag}(\eigval_1,\ldots,\eigval_n)
        :
        \eigval_i\in\{\eta,1\},
        {\rm trace}(D)=m\eta+n-m
        \}.
    \end{eqnarray}
We will always assume that $0\leq m \leq n$.
Note that
$\mathcal D(0,n,\eta)=\{{\rm Id}\}$
and
$\mathcal D(n,n,\eta)=\{\eta{\rm Id}\}$.

\subsection{Absolute value activation}

Consider the case $\eta=-1$, corresponding
to an activation of the type ``absolute value'' (see (\ref{EQU:non_linearities_eta})).

\begin{prop}[Absolute Value]
  \label{prop:absolute_value}
  \sloppy
    Let $M_o$ be arbitrary (see \equ{Mo}).
  For any $D\in\mathcal{D}(m,n,-1)$, we have
  \begin{equation}
    \label{EQU:absolute_value}
    \singval_i(M)
    =
    \singval_i(MD)
    =
    \singval_i(DM).
  \end{equation}
\end{prop}
\begin{proof}
  This is a simple consequence of the SVD.
  Since $D$ is orthogonal due to $\eta=-1$,
  $DV$ and $DU$ are orthogonal as well.
  Multiplication by $D$ only changes the singular vectors and
  not the singular values: $DM=D(V\Sigma U^T)=(DV)\Sigma U^T$
  and $MD=(V\Sigma U^T)D=V\Sigma (DU)^T$.
~\end{proof}

\subsection{ReLU activation}

The next simple case is the ReLU activation, corresponding to $\eta=0$ in (\ref{EQU:non_linearities_eta}). As we state next, the singular values of a linear map $M$ decrease when the map is combined with a ReLU activation, from left or right.

\begin{lem}[ReLU Decreases Singular Values]
  \label{lem:ReLU_decrease}
  ~\\
    Let $M_o$ be arbitrary (see \equ{Mo}).
  For any $D\in\mathcal{D}(m,n,0)$ we have
  \begin{equation}
    \label{EQU:ReLU_decreases}
    \singval_i(M)
    \geq
    \singval_i(MD)
    \qquad\mbox{and}\qquad
    \singval_i(M)
    \geq
    \singval_i(DM).
  \end{equation}
\end{lem}
Note that a similar result holds for leaky ReLU ($\eta\neq0$), however only for $i=1$.
This follows from the fact that $\singval_1(MD)\leq \singval_1(M)\singval_1(D)$ for any two matrices $M$ and $D$.

\begin{proof}
This is a simple consequence of the Cauchy interlacing property by deletion (see (\ref{EQU:deletion})).
~\end{proof}

As a consequence, we have for ReLU that
$
    |
    \singval_i(M)
    -
    \singval_i(DM)
    |
    =
    \singval_i(M)
    -
    \singval_i(DM)
$
and the analogous for $MD$.
This motivates the  computation of the difference
of traces which equals the sum of squares of the singular values
(recall (\ref{EQU:frobenius_norm})).
\begin{prop}[ReLU]
  \label{prop:ReLU_output}
  \label{prop:ReLU_multiple_output}
  ~\\
    Let $M_o$ be arbitrary (see \equ{Mo}).
    Let $\PertVar$ be arbitrary and write $M$ instead of $M(\PertVar)$ for short.
  For any $D\in\mathcal{D}(m,n,0)$, the following holds:
  Let $k_i$ ($i=1\ldots m$) be such that
    $D[k_ik_i]=0$.
      Denote by ${\singval_i}$ the singular values of $M$ and
      by $\tilde{\singval_i}$ those of
      $MD$.
  Then
  \begin{eqnarray}
    \label{EQU:ReLU_multiple_trace_output}
    \sum_{i=1}^n \singval_i^2
    -
    \sum_{i=1}^n \tilde\singval_i^2
    =
    \sum_{i=1}^n
        |\singval_i^2-\tilde\singval_i^2|
    &=&
    \|M\|_F^2-\|MD\|_F^2
    =
    \sum_{i=1}^m
    \sum_{j=1}^{n} M[j k_i]^2
    \\
    \label{EQU:ReLU_multiple_singval_output}
    \sum_{i=1}^n
    |\singval_i-\tilde\singval_i|^2
    &\leq&
    \sum_{i=1}^m
    \sum_{j=1}^{n} M[j k_i]^2.
  \end{eqnarray}
\end{prop}

\begin{proof}
Write $\tilde M=MD$ for short.
The following equality simplifies our computation and holds in this special setting,
because $D$ amounts to an orthogonal projection enabling us to appeal to Pythagoras:
$ \|M\|_F^2
    =
    \|\tilde M\|_F^2
    +
    \|M-\tilde M\|_F^2
.
$
This can also be verified by direct computation.
Lemma~\ref{lem:ReLU_decrease} implies
\begin{equation}
%    \sumSQsing(M)-\sumSQsing(\tilde M)
%%%@    \sumSQsing-\tilde\sumSQsing
    \sum_{i=1}^n
        |\singval_i^2-\tilde\singval_i^2|
    =
    \sum_{i=1}^n \singval_i^2
    -
    \sum_{i=1}^n \tilde\singval_i^2
    =
    \|M\|_F^2-\|\tilde M\|_F^2
    =
    \|M-\tilde M\|_F^2
    =
    \sum_{i=1}^m
    \sum_{j=1}^{n} M[j,k_i]^2,
\end{equation}
yielding \equ{ReLU_multiple_trace_output}.
Using Corollary~\ref{cor:Frobenius_bound}
with ${\AnyM}=M$ and ${\AnyMt}=\tilde M$, we obtain \equ{ReLU_multiple_singval_output}.
~\end{proof}

Bounds on the effect of nonlinearities at the input can be obtained easily from
bounds at the output using the following fact.

\begin{prop}[Input-Output via Transposition]
  \label{prop:input-output}
  For $D$ diagonal,    \begin{equation}\label{EQU:input-output}
    \singval_i(MD)
    =
    \singval_i(DM^T).
  \end{equation}
\end{prop}

\begin{proof}
This follows from the fact that matrix transposition does not alter singular values.
%combined with $D=D^T$.
~\end{proof}

\section{Bounding  the Weights of a ResNet Bounds its Condition Number}
\label{sec:Identity_withDiagonal}

In this section, we combine the results of the preceding two sections in order to quantify 
how bounds
on the entries of the matrix $W_i$
translate into bounds on the condition number. 
As we have seen, this is of relevance in the context of optimizing
DNs, especially when comparing ConvNets to ResNets (recall
Section~\ref{subsec:Conv_vs_Res}).

To this end, we will proceed in three steps:
\begin{enumerate}
    \item Bounds on the weights $W_i[j,k]$ translate into bounds of $M(0)$ (ConvNet)
    \item Perturbation when adding a residual link, moving from  $M(0)$ (ConvNet) to  $M(1)$ (ResNet)
    \item Perturbation by the activation nonlinearity of the trained layer (passing from $M(1)$ to $M(1)D$)
\end{enumerate}

We treat the two cases of deterministic weights (hard bounds)  and random weights (high probability bounds) separately.

\subsection{Hard bounds}
\label{sec:hard_bounds}

We start with a few facts on how hard bounds on the entries of a matrix
are inherited by the entries and singular values of
perturbed versions of the matrix.
We use the term ``hard bounds'' for 
the following results on weights $W_i$ bounded by a constant
in order to distinguish them
from ``high probability bounds'' for random weights $W_i$
with bounded standard deviation.
Clearly, for uniform random weights we can apply both types of results.

\begin{lem}
  \label{lem:op_norm_hard_bound}
  Assume that all entries of the $n\times n$ matrix $W_i$
  are bounded by the same constant: $|W_i[j,k]|\leq c$.
  Then
  \begin{equation}\label{EQU:sigma1_hard_bound}
    \singval_1(W_i)
    \leq
    cn,
  \end{equation}
  with equality, for example, for the matrix $W_i$ with all entries equal to $c$.
\end{lem}

\begin{proof}
  The bound is a simple consequence of (\ref{EQU:operator_norm}):
    $\singval_1(W_i) \leq \|W_i\|_F \leq cn$.
  If all entries of $W_i$ are equal to $c$, then the unit vector
  $u_1=(1/\sqrt n) (1,\ldots,1)^T$ achieves $W_i\, u_1=c\,n\, u_1$.
~\end{proof}

\begin{lem}
  \label{lem:product_hard_bound}
  Let $W_i$ ($i=1,\ldots, p$) be
  matrices as in Lemma~\ref{lem:op_norm_hard_bound}.
  Assume that $c\leq 1/n$.
  Let $D_i$ ($i=1\ldots p$) be diagonal matrices with entries
  bounded by $1$.
  Then all
  the entries of $M_o=M(0)=D_pW_p\cdots D_1W_1$ are bounded by $c$
  as well.
\end{lem}

\begin{proof}
  Each entry of $D_iW_i$ is bounded by $c$
  and each entry of $(D_2W_2)(D_1W_1)$ is bounded by
  $nc^2=(nc)c\leq c$. Iterating the argument establishes
  the second claim.
~\end{proof}

Recall that $M(0)=M_o$ corresponds to a multi-layer $F=f_p\ldots f_1$
with arbitrary nonlinearities.
Let us add a skip-residual link by choosing $\PertVar=1$ in \equ{training_multi_layer}.

\begin{lem}
  \label{lem:increase_singval_hard_bound}
  Let $W_i$ be as in Lemma~\ref{lem:product_hard_bound}
  and recall \equ{training_multi_layer}.
  Assume that $c<1/n$.
  Then
  \begin{equation}
  \label{EQU:increase_singval_hard_bound}
    0
    <
    1-cn
    \leq
    \singval_n(M(1))
    \leq
    \singval_1(M(1))
    \leq
    1+cn
    <
    2
  \end{equation}
  and
  \begin{equation}
  \label{EQU:conditioning_hard_bound}
    \CondNb(M(1))
    \leq
    \frac{1+cn}{1-cn}
    .
  \end{equation}
\end{lem}

\begin{proof}
To establish (\ref{EQU:increase_singval_hard_bound}) apply
Lemma~\ref{lem:Weyl_bound} with $i=1$ and $i=n$ and 
$\PertVar=1$, and note
that $\singval_1(M_o)\leq cn$ by Lemmata~\ref{lem:op_norm_hard_bound} and~\ref{lem:product_hard_bound}.
\MoreDetails{
    Alternatively, apply
    (\ref{EQU:uniform_bound}) with ${\AnyM}=M(1)$ and ${\AnyMt}={\rm Id}$.
    Note that $\singval_i({\AnyMt})=1$ and ${\AnyM}-{\AnyMt}=M(1)-{\rm Id}=M(0)$ which yields
    $\singval_i({\AnyM})-\singval_i({\AnyMt})=\singval_i(M(1))-1\leq\|{\AnyM}-{\AnyMt}\|_F
    =\singval_1(M(0))$.
}
The bound (\ref{EQU:conditioning_hard_bound}) follows immediately.
~\end{proof}

Proposition~\ref{prop:absolute_value} implies the following bound for the absolute value activation.

\begin{thm}[Hard Bound, Absolute Value]
  \label{thm:increase_singval_hard_bound}
    Under the assumptions of 
    Lemma~\ref{lem:increase_singval_hard_bound}
    and for 
    $D\in\mathcal{D}(m,n,-1)$,
    we have
    \begin{equation}\label{EQU:hard_abs}
%%%@        \CondNb((M(1))D^{(\sboldk)}(-1))
        \CondNb(M(1)D)
        \leq
        \frac{1+cn}{1-cn}.
    \end{equation}
\end{thm}

For a residual multi-layer \equ{training_multi_layer}
with ReLU nonlinearities, we find the following.

\begin{thm}[Hard Bound, ReLU]
  \label{thm:conditioning_ReLU_hard_bound}
    Under the assumptions of 
    Lemma~\ref{lem:increase_singval_hard_bound}
    and for 
    $D\in\mathcal{D}(m,n,0)$, the following statements hold:
  
  \noindent
  Provided $2cn<1/(1+n)$,
  we have
  \begin{equation}
  \label{EQU:conditioning_ReLU_hard_bound}
    \CondNb(M(1)D)
    \leq
    \frac{1+cn}{\sqrt{1-2cn(1+n)}}.
  \end{equation}
  Provided $m\leq n(1-cn)^2$, we have
  \begin{equation}
  \label{EQU:conditioning_ReLU_hard_bound_small_m}
    \CondNb(M(1)D)
    \leq
    \frac{1+cn}{1-cn-\sqrt{m/n}}
    .
  \end{equation}
  Provided $2cn<1/(1+2m)$,
  we have
  \begin{equation}
  \label{EQU:conditioning_ReLU_hard_bound_alt}
    \CondNb(M(1)D)
    \leq
    \frac{1+cn}{\sqrt{1-2cn(1 +2m)}}
    .
  \end{equation}

\end{thm}

\begin{proof}
Applying Proposition~\ref{prop:ReLU_multiple_output} with $\PertVar=1$, i.e., $M=M(1)$, we have by
(\ref{EQU:ReLU_multiple_trace_output}) that
\begin{equation}
    \sum_{i=1}^{n}
        |\singval_i^2 - \tilde\singval_i^2 |
%    =
%    \|M\|_F^2-\|\tilde M\|_F^2
    =
    \|M(1)- M(1)D\|_F^2
    \leq
    m\cdot (c^2n+2c+1)
    .
\end{equation}
Note that this bound
cannot be made arbitrarily small
by choosing $c$ small and
that it amounts to at least $m$.
However, the bulk of the sum of deviations
stems from some $m$ singular values of $M$
being changed to zero by $D$,
compensating the constant term $m$ in the above bound.
More precisely,
$\singval_i\geq 1-cn$ by (\ref{EQU:increase_singval_hard_bound})
for all $i$
while
$\tilde\singval_i=0$ for $i=n-m+1,\ldots, n$.
Thus,
\begin{equation}
    \sum_{i=n-m+1}^{n}
        |\singval_i^2 - \tilde\singval_i^2 |
    =
    \sum_{i=n-m+1}^{n}
        \singval_i^2
    \geq
    m\cdot (1-cn)^2
    ,
\end{equation}
implying that for any $1\leq j \leq n-m$
\begin{equation}
    |\singval_j^2 - \tilde\singval_j^2 |
    \leq
    \sum_{i=1}^{n-m}
        |\singval_i^2 - \tilde\singval_i^2 |
    \leq
    m\cdot (c^2n+2c+1)
    -
    m\cdot (1-cn)^2
%    =
%    mc(cn+2+2n-cn^2)
%%    =
%%    mc(2+n(2+c-cn))
    \leq
    2mc(1+n)
\end{equation}
(using $cn-cn^2<0$).
Therefore,
\begin{eqnarray}
    \tilde\singval_j^2
    &\geq&
    \singval_j^2 - 2mc(1+n)
    \geq
    (1-cn)^2 - 2mc(1+n)
    =
    1-c(2n+ 2m(1+n)-cn^2)
%    \geq
%    1-c(2n^2 + 2n)
\nonumber
\\ 
    &\geq&
    1-2c(n^2+n)
\end{eqnarray}
(using $m(1+n) \leq n^2-1\leq n^2$).
Since this bound holds for all non-zero singular values of $\tilde M$,
we obtain
\begin{equation}
    \tilde\singval_*
    \geq
    \sqrt{1-2c(n^2+n)}
%    \geq
%    {1-2c(n^2+n)}
    .
\end{equation}
Finally, we have $\tilde\singval_1\leq\singval_1\leq 1+cn$
by Lemma~\ref{lem:ReLU_decrease} and (\ref{EQU:increase_singval_hard_bound}).
This proves (\ref{EQU:conditioning_ReLU_hard_bound}).
Alternatively, we can estimate as follows:
\begin{equation}
    \tilde\singval_j^2
    \geq
    1-c(2n+ 2m(1+n)-cn^2)
    \geq
    1-cn(2 +2m(1+1/n)-cn)
    \geq
    1-2cn(1 +2m)
\end{equation}
using $1/n\leq 1$, which proves \equ{conditioning_ReLU_hard_bound_alt}.
At last, we can use \equ{ReLU_multiple_singval_output}
instead of (\ref{EQU:ReLU_multiple_trace_output}). 
Analogous to before, we find
\begin{equation}
    \sum_{i=1}^{n}
        |\singval_i - \tilde\singval_i |^2
    \leq
    \|M(1)-M(1)D\|_F^2
    \leq
    m\cdot (c^2n+2c+1),
\end{equation}
implying for any $1\leq j \leq n-m$ that
$
    |\singval_j - \tilde\singval_j |^2
    \leq
    2mc(1+n)
    .
$
Therefore, provided $2cn<1/(1+n)$ or  $2c(1+n)<1/n$, we obtain
\equ{conditioning_ReLU_hard_bound_small_m} by observing that
\begin{equation}
    \tilde\singval_j
    \geq
    \singval_j - \sqrt{2mc(1+n)}
    \geq
    (1-cn) - \sqrt{2mc(1+n)}
    \geq
    1-cn -\sqrt{m/n}
    .
\end{equation}
~\end{proof}

For leaky-ReLU, the situation is more intricate.

\begin{thm}[Hard Bound, Leaky-ReLU]
  \label{thm:conditioning_Leaky_ReLU_hard_bound}
  Let $W_i$ be as in Lemma~\ref{lem:product_hard_bound}
  and recall \equ{training_multi_layer}.
  Assume that $c<1/n$.
  Let $D\in\mathcal{D}(m,n,\eta)$ with $0\leq \eta\leq1$.
%%%@  Let $D=D^{(\sboldk)}(0)$
  %which  corresponds to a
  %nonlinearity of the type ReLU
%%%@  with $\boldk=(k_1\ldots k_m)$ and $0\leq m\leq n-1$.
  Provided $n\geq5$ and $\eta<1$ large enough to ensure
    $(1-cn)^2
        >
    4(1-\eta)\sqrt{m}
    $,
  we have
  \begin{equation}
  \label{EQU:conditioning_Leaky_ReLU_hard_bound}
    \CondNb(M(1)D)
    \leq
    \frac{1+cn}{
        \sqrt{
        (1-cn)^2
        -
        4(1-\eta)\sqrt{m}
        }}.
  \end{equation}
  Alternatively, if $\eta<1$ is such that
  $1-cn>(1-\eta)m(3c+1)$, then
  we have
  \begin{equation}
  \label{EQU:conditioning_Leaky_ReLU_hard_bound_alt}
    \CondNb(M(1)D)
    \leq
    \frac{1+cn}{{1-cn-(1-\eta)m(3c+1)}}
    .
  \end{equation}

\end{thm}

This situation corresponds to a ResNet \equ{training_multi_layer}
with leaky-ReLU nonlinearities, because $\eta$ can take any value between 0 and 1 (see \equ{non_linearities_eta}).

\begin{proof}
  We proceed along the lines of the arguments of
  Theorem~\ref{thm:conditioning_ReLU_hard_bound}.
  To this end, we need to generalize the results of
  Proposition~\ref{prop:ReLU_multiple_output}.

  In order to replace \equ{ReLU_multiple_trace_output} we
  write for short 
  $M=M(1)=M_o+$Id with singular values $\singval_i$
  and $\tilde M=M(1)D$ 
  with singular values $\tilde\singval_i$.
  Applying Proposition~\ref{prop:Frobenius_bound} with
  ${\AnyM}=M^TM$ and ${\AnyMt}= \tilde M^T\tilde M=
  (MD)^TMD=D{\AnyM}D$, we obtain
\begin{equation}
  \sum_{i=1}^n
  |\singval_i^2-\tilde\singval_i^2 |^2
  \leq
          \|{\AnyM}-{\AnyMt}\|_F^2
  .
\end{equation}
  Since all entries of $M_o$ are bounded by $c$ due to
  Lemma~\ref{lem:product_hard_bound} and using $cn<1$,
   it follows
  that the entries of ${\AnyM}$ are bounded as
  follows
\begin{equation}
    |{\AnyM}[i,j]|
    \leq
    \twocases{
        (n-2)c^2+2(1+c)c\leq3c}{if $i\neq j$}{
        (n-1)c^2+(1+c)^2\leq3c+1~~~}{if $i= j$
        .}
  \end{equation}
  Using these bounds, 
  direct computation shows that
  $
      {\AnyM}-{\AnyMt}={\AnyM}-D{\AnyM}D
  $ contains $2m(n-m)$ terms off the diagonal bounded by $3c(1-\eta)$,
  $m$ terms in the diagonal bounded by $(3c+1)(1-\eta^2)$,
  further $m^2-m$ terms off the diagonal bounded by $3c(1-\eta^2)$,
  and the remaining $(n-m)^2$ terms equal to zero.
  Using $cn<1$ whenever apparent
  as well as $(1-\eta^2)\leq 2(1-\eta)$, we can estimate as follows:
  \begin{eqnarray}
    \|{\AnyM}-{\AnyMt}\|_F^2
    &\leq&
    (1-\eta)^2 18c^2 m(n-m)
        +
    (1-\eta^2)^2\left(
                9c^2 (m^2-m)+(3c+1)^2 m
                \right)
    \nonumber \\
    &\leq&
    (1-\eta)^2
        18c^2 m(n-m)
        +
    4(1-\eta)^2\left(
                9c^2 m^2+6cm+ m
                \right)
    \nonumber \\
    &\leq&
    (1-\eta)^2(42cm+18c^2m^2+4m)
    \leq
    4(1-\eta)^2m(15/n+1)
    .
  \end{eqnarray}
  We conclude that,
  for $n\geq5$, we have
  $|\singval_j^2-\tilde\singval_j^2 |\leq (1-\eta)4\sqrt{m}$.
  Together with $\singval_j\geq (1-cn)$,
  we obtain
   \equ{conditioning_Leaky_ReLU_hard_bound}.

Alternatively, we may strive at adapting
\equ{ReLU_multiple_singval_output} (which is based on
Corollary~\ref{cor:Frobenius_bound})
for $\eta\neq0$.
Recomputing
\equ{ReLU_multiple_singval_output} using
\equ{strong_frobenius_bound}
with ${\AnyM}=M$ and ${\AnyMt}=\tilde M$  for non-zero $\eta$,
we obtain
\begin{equation}
\sum_{i=1}^n
    |\singval_i-\tilde\singval_i|^2
    \leq
        \|M-\tilde M\|_F^2
    \leq
%%%@
%    (1-\eta)^2
%    \sum_{i=1}^m
%    \sum_{j=1}^{n} M[j k_i]^2
%    \leq
    (1-\eta)^2
    m(c^2n+2c+1).
\end{equation}
We obtain
$|\singval_j-\tilde\singval_j |
\leq (1-\eta)\sqrt{m(c^2n+2c+1)}
\leq (1-\eta)\sqrt{m(3c+1)}
$
and \equ{conditioning_Leaky_ReLU_hard_bound_alt}.
~\end{proof}

We note that the earlier Theorem~\ref{thm:chi_bounds}
will not lead to strong results in this context.
However, it will prove
 effective in the random setting of the
next section.

\subsection{Bounds with high probability for random matrices}
\label{sec:probabilities}

While the results of the last section provide absolute guarantees,
they are quite restrictive on the matrix $M_o$ since they cover the worst cases.
Working with the average behavior of a randomly initialized matrix $M_o$
enables us to relax the restrictions
from bounding the actual values of the weights to
only bounding their standard deviation.

The strongest results in this context concern the operator norm
and exploit cancellations in the entries of $M_o$
(see \cite[Proposition 3.1]{BAL}, \cite{Tracy_Widom,Tracy_Widom2}).
Most useful for this study is the well-known
result due to Tracy and Widom \cite{Tracy_Widom,Tracy_Widom2}.
Translated into our setting, it reads as follows.

\begin{thm}[Tracy-Widom]
  \label{thm:op_norm_edge_bound}
    Let $X_{i,j}$ ($1\leq i,j$) be an infinite family of
    iid copies of a Uniform or Gaussian random variable $X$
    with zero mean and variance $1$.
    %%%$\Std^2$.
    Let $K_n$ be a sequence of $n\times n$ matrices with entries
    $K_n[i,j]=X_{i,j}$ ($1\leq i,j\leq n$).

    Let $\overline{t_n}$ denote the largest eigenvalue of
    $K_n^T+K_n$ and let $\underline{t_n}$ denote its smallest.

  Then, as $n\to\infty$, both $\overline{t_n}$ and $-\underline{t_n}$
  lie  with high probability in the interval
  \begin{equation}\label{EQU:tau_edge_bounds}
    %%%\Std\cdot
    \left[ \sqrt{8n}- O(n^{-1/6}),\sqrt{8n}+O(n^{-1/6}) \right]
    .
  \end{equation}
  More precisely, the distributions of
\begin{eqnarray}
    \label{EQU:TAU_edge_bound_sharper}
   Z_n
   &:=&
   (\overline{t_n}%%%/\Std
        -\sqrt{8n})\, n^{1/6}
   \\
    \label{EQU:tau_edge_bound_sharper}
   \tilde Z_n
   &:=&
   (-\underline{t_n}%%%/\Std
        -\sqrt{8n})\, n^{1/6}
\end{eqnarray}
   converge both (individually) to a Tracy-Widom law $P_{TW}$. 
   The Tracy-Widom law is well concentrated
   at values close to $0$.
\end{thm}

The random variables $Z_n$ and $\tilde Z_n$
are equal in distribution due to the
symmetry of the random variable~$X$. However,
they are not independent.

\begin{figure}
    \centering
    (a)
    \includegraphics[width=0.45\linewidth]{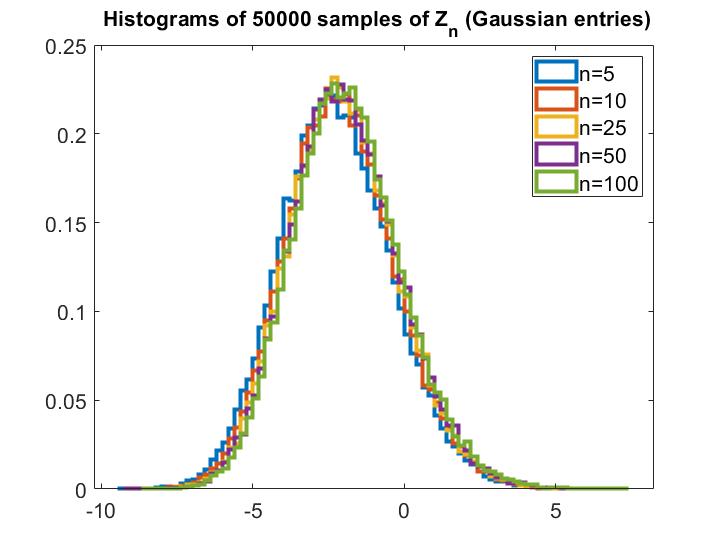}
    (b)
    \includegraphics[width=0.45\linewidth]{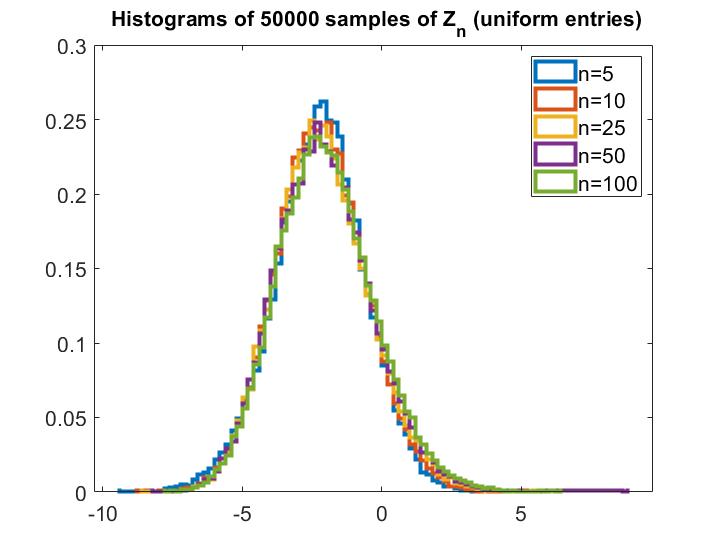}
    \caption{
    Empirical demonstration of the Tracy-Widom law.
    Note the collapsing histograms of
    $Z_n=(\overline{t_n}%%%/\Std
    -\sqrt{8n})\, n^{1/6}$
    for $n=5, 10, 25, 50, 100$
    where $\overline{t_n}$ denotes the largest eigenvalue of the
    $n\times n$ matrix of the form $K_n+K_n^T$ 
    and where the entries of $K_n$ are drawn from
    a (a) Gaussian or (b) Uniform distribution with zero mean and
    variance $1$.
    %%%$\Std^2$.
    Note that the right-side tails $P[Z_n>t]$ decay very quickly,
    leading to negligible probabilities at moderate values of $t$.}
    \label{fig:tracy-widom}
\end{figure}

It is easy to verify that the theory of Wigner matrices and the
law of Tracy-Widom apply here and to establish Theorem~\ref{thm:op_norm_edge_bound}.
In particular, $K_n^T+K_n$ is symmetric with iid\ entries above the diagonal
and iid\ entries in the diagonal, all with finite non-zero second moment.
The normalization of $\overline{t_n}$ leading to $Z_n$ (see \equ{TAU_edge_bound_sharper}) is quite
sharp, meaning that the distributions of $Z_n$ are close to their limit starting at
values as low as $n=5$ (see Figure~\ref{fig:tracy-widom}).

The Tracy-Widom distributions
are well studied and documented, with a right tail that
decays exponentially fast \cite{TW_Tail}.
Moreover, they are well concentrated at values close to $0$.
Combined with the fast convergence, this implies that $\overline{t_n}$ is close to
$ \sqrt{8n}$ with high probability even for modest $n$.

With the appropriate adjustments, Theorem~\ref{thm:op_norm_edge_bound}
provides tight control on $\overline{\tau}$
and $\underline{\tau}$ (see Definition~\ref{def:tau}). In order to exploit
this control toward a bound on the growth of singular values via Theorem~\ref{thm:lambda_bounds}, the following proves useful.
\begin{thm}
  \label{thm:Mo_simple}
  Assume that the $n^2$ entries 
  of the random matrix $M_o$
  are  jointly continuous with some joint probability
  density function (e.g.,
  Uniform or jointly Gaussian entries).
  Then, almost surely, $M_o$ possesses no
  multiple singular value, and
  the growth of the singular values of
  $M_o+\PertVar{\rm Id}$ is bounded as in 
  \equ{lambda_increase_bounds}.
\end{thm}

\begin{proof}
    As in the proof of
    Lemma~\ref{lem:discriminant}, we argue
    that the discriminant $\chi$ of the 
    characteristic polynomial of the $n\times n$
    matrix $M_o^TM_o$ 
    is itself a polynomial of the 
    entries of $M_o$. Viewing $M_o$ as a point
    in $\reals^{n^2}$ defined by its entries, 
    the matrices $M_o$ with multiple singular values
    form a set that is 
    identical to the zero-set of the
    polynomial $\chi$ 
    and form, therefore, a set of Lebesgue measure $0$. The statement is now obvious.
%\hfill$\Box$
\end{proof}

\begin{figure}
    \centering
    (a)
    \includegraphics[width=0.45\linewidth]{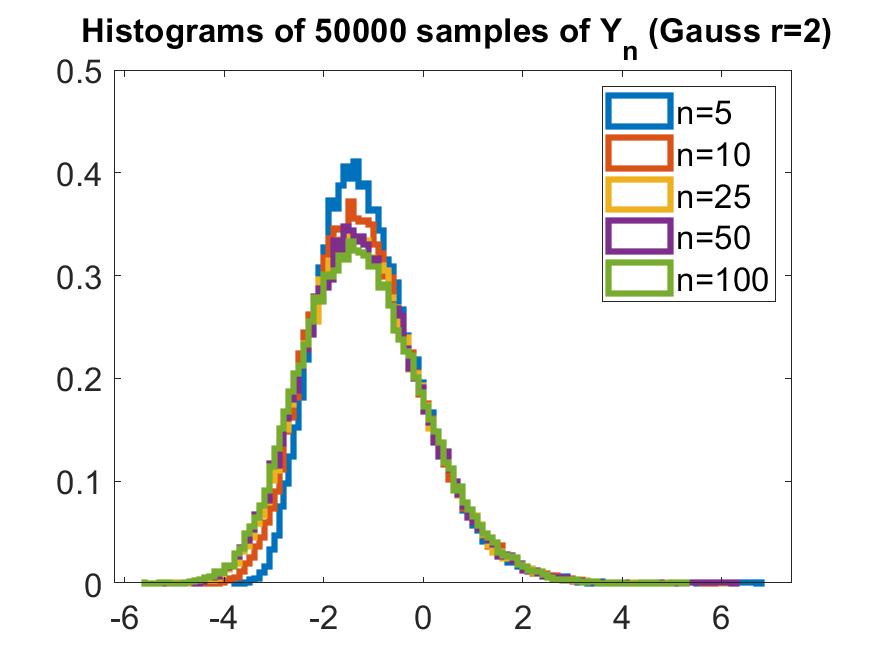}
    (b)
    \includegraphics[width=0.45\linewidth]{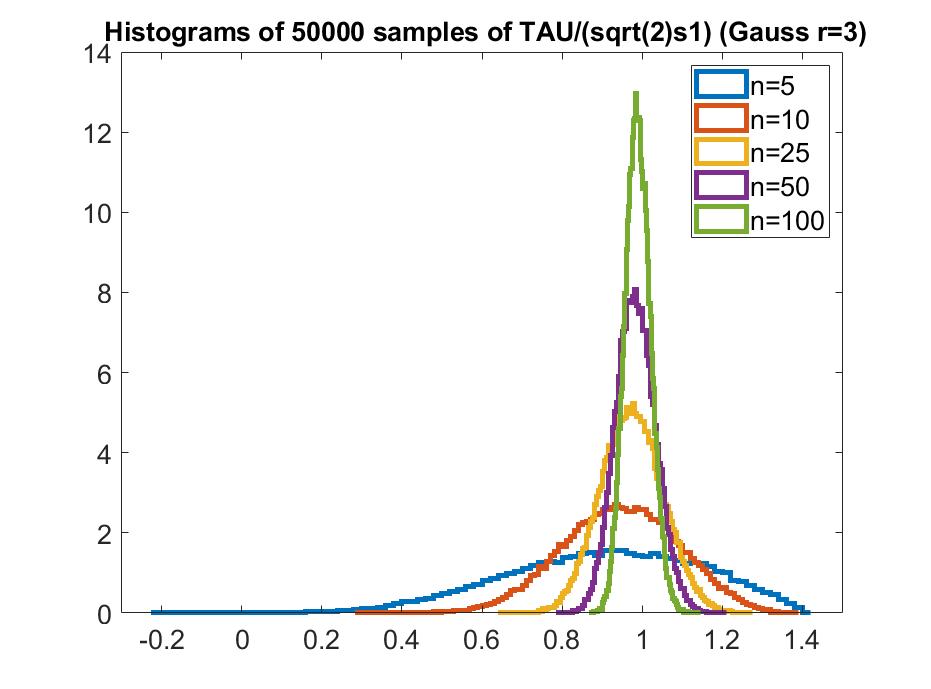}
    \caption{
    (a) Empirical demonstration of the convergence of $Y_n$ in
    \equ{Yn_def}
to a Tracy-Widom law.
Note that the right-side tails $P[Y_n>t]$ decay very quickly,
    leading to negligible probabilities at moderate values of $t$ and for $n$
    as small as $5$
    similar to $Z_n$ from Figure~\ref{fig:tracy-widom}.
    (b) Empirical demonstration of the convergence of $\overline{\tau}/(\sqrt{2}\singval_1)$
    to $1$: The ratio lies indeed with high probability between $0.6$ and $1.4$ for 
    $n$ as small as $10$.}
    \label{fig:Yn_convergence}
\end{figure}

A related useful result reads as follows
(see \cite[Theorem~6.2.6 and Section 6.2.3]{Chafai} and the references therein).
\begin{thm}
  \label{thm:Op_norm_Gauss}
  Let $K_n$ be as in Theorem~\ref{thm:op_norm_edge_bound}, but Gaussian. Then, as $n\to\infty$
  \begin{equation}
      \label{EQU:Op_norm_Gauss_limit}
      \frac {\singval_1(K_n)}{\sqrt{n}}
    \;\stackrel{\rm a.s.}{\rightarrow}\;
      2 .
  \end{equation}
  Moreover, letting $\mu_n=(\sqrt{n-1}+\sqrt{n})^2$
  and $q_n=\sqrt{\mu_n}(1/\sqrt{n-1}+1/\sqrt{n})^{1/3}$, 
  the distribution of 
\begin{equation}
      \label{EQU:Yn_def}
      Y_n
      :=
      ({\singval_1(K_n)^2-\mu_n})/{q_n}
  \end{equation}
    converges narrowly to a Tracy-Widom law.
%    \\
%    Also, almost surely, $K_n$ has no multiple  singular values.
\end{thm}
Comparing with Theorem~\ref{thm:op_norm_edge_bound},
we should expect that $\singval_1(K_n)\simeq 2\sqrt{n}\simeq \overline t_n/\sqrt{2}$
(compare to Figure~\ref{fig:Yn_convergence}).
Combining
%Proposition~\ref{prop:max_value} as well as
Theorem~\ref{thm:chi_bounds} (with $M_0=W_1$)
%, i.e., $D_1=$Id), 
and Theorem~\ref{thm:op_norm_edge_bound},
and using that (for $n\geq2$) $\mu_n\leq 4n$ and
$q_n
\leq 
\sqrt{(4n)}2^{1/3}(n-1)^{-1/6}
\leq 
\sqrt{(4n)}2^{1/3}(n/2)^{-1/6}
\leq 
2^{3/2}n^{1/3}
$,
we obtain the following.  

\begin{lem}[High Probability Bound]
  \label{lem:Abs_Value_average_bound}
  In the notation of Theorem~\ref{thm:op_norm_edge_bound},
  let $r>1$ and $n\geq 2$, set
  \begin{equation}
  \label{EQU:Sigma_n_Abs_Value}
    \Std_n
    =
    \frac{1}{r\, \sqrt{8n}},
  \end{equation}
  and let
  $
    W_1 %M_n
    =
    \Std_n K_n
    .
    $
  Denote the largest and smallest eigenvalues of $W_1^T+W_1$ % M_n^T+M_n$
  by $\overline{\tau}_n=\overline{\tau}_n(r)$ and $\underline{\tau}_n=\underline{\tau}_n(r)$
  (compare with \equ{OpNorm_MT+M}).
 
 \noindent
  (i) We have
  \begin{eqnarray}
  \label{EQU:TAU=renormZn}
    % proof:
    % \overline{\tau} &=& \Std_n\cdot t_n =  \Std_n\cdot (\sqrt{8n}+Z_n n^{-1/6}\\
    \overline{\tau}_n(r) 
    &=& \frac 1 r \, \left(1 +
            Z_n \frac {1}{\sqrt{8}\, n^{2/3}}\right)
    \qquad
    -\underline{\tau}_n(r)
    = \frac 1 r \, 
    \left(1 +
        \tilde   Z_n \frac {1}{\sqrt{8}\, n^{2/3}}
    \right)
  \end{eqnarray}
  and
  $$
    P[\underline{\tau}_n(r)>-1]
    =
    P\left[\tilde Z_n\leq (r-1)\sqrt{8} n^{2/3}\right]
    .
  $$
  
   \noindent
  (ii) In the Gaussian case, we have that
  $
    \singval_1(W_1)
    {\rightarrow}
%    \stackrel{\rm a.s.}{\rightarrow}
    1/(\sqrt{2}r)
$
%and \equ{lambda_increase_bounds}, both
%with probability $1$, i.e., 
almost surely.
Also
\begin{equation}
      \label{EQU:renorm_s1(W1)}
      \singval_1(W_1)^2
      =
      \frac{1}{2r^2}
      \left( 
      \frac{\mu_n}{4n}
      +
      n^{-2/3}\frac{q_n}{4n^{1/3}}
      Y_n
      \right)
      \leq
      \frac{1}{2r^2}
      \left( 
      1
      +
      n^{-2/3}\frac{1}{\sqrt{2}}
      Y_n
      \right)
  \end{equation}

 \noindent  (iii)
  For any realization of $W_1$ with
%%%   $S_1$  dense in $[0,1]$ and
  $\singval_1(W_1)<1$
  \begin{equation}
  \label{EQU:conditioning_average_bound_via_s1}
    \CondNb(W_1+{\rm Id})
    \leq
    \frac{\sqrt{\singval_1^2(W_1)+\overline\tau_n+1
        }}{1-\singval_1(W_1)}
    \leq
    \frac{1+\singval_1(W_1)}{1-\singval_1(W_1)}
    .
  \end{equation}

  \noindent (iv)
  For any realization of $W_1$
  with 
  $\underline{\tau}_n>-1$ 
   we have
  the slightly tighter bounds 
  \begin{equation}
  \label{EQU:increase_singval_average_bound}
    0
    <
    \singval_n^2(W_1)+\underline{\tau}_n+1
    %1-\Std\sqrt{8n}
    \leq
    \singval_n^2(W_1+{\rm Id})
    \leq
    \singval_1^2(W_1+{\rm Id})
    \leq
    \singval_1^2(W_1)+\overline{\tau}_n+1.
  \end{equation}
  The condition number of
  $W_1+{\rm Id}$
   is then bounded as
  \begin{equation}
  \label{EQU:conditioning_average_bound}
  \CondNb_n
  :=
    \CondNb(W_1+{\rm Id})
    \leq
    \sqrt{\frac{\singval_1^2(W_1)+\overline\tau_n+1}{\underline\tau_n+1}}
    \leq
    \frac{\singval_1(W_1)+1}{\sqrt{\underline\tau_n+1}}
    .
  \end{equation}
%\fi

\end{lem}

\begin{figure}
    \centering
\begin{tabular}{ccc}
    \includegraphics[width=0.3\linewidth]{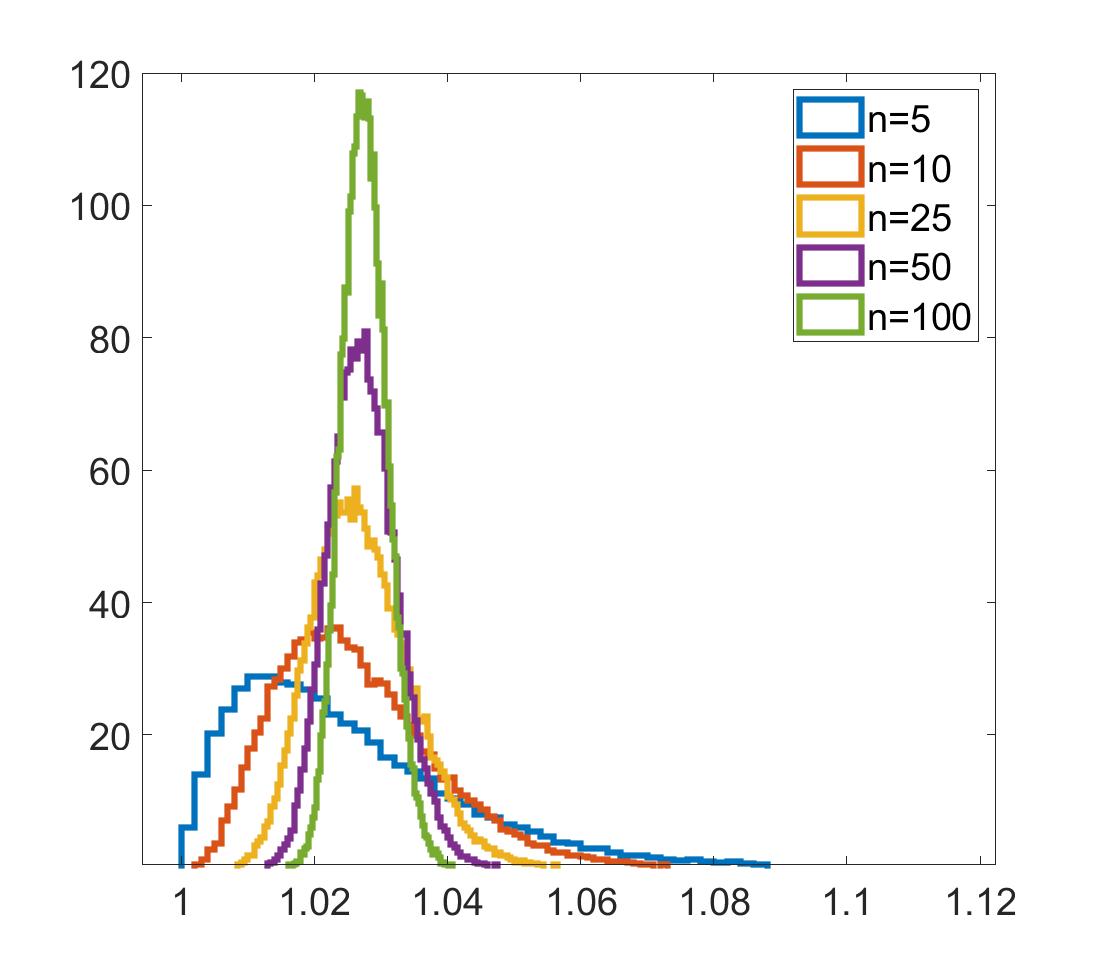}
    &
    \includegraphics[width=0.3\linewidth]{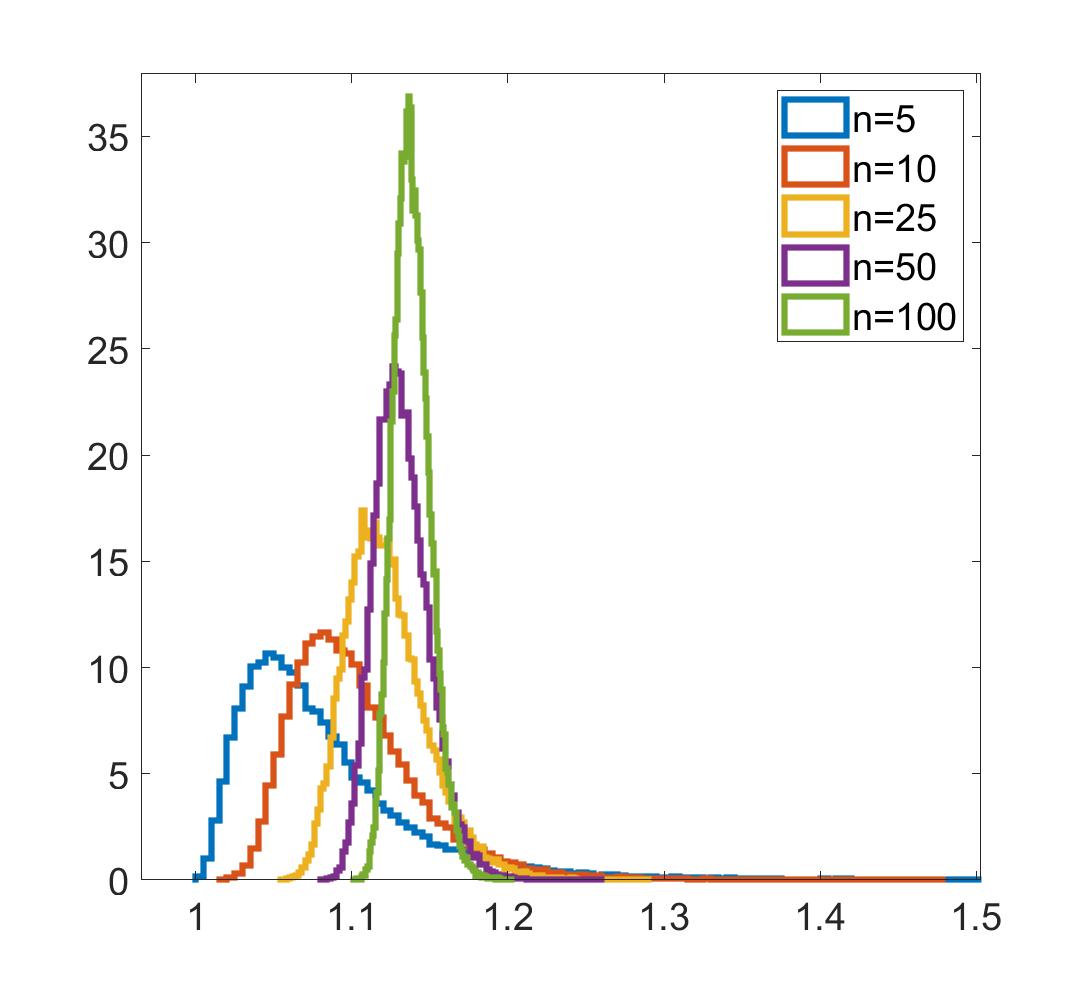}
    &
    \includegraphics[width=0.3\linewidth]{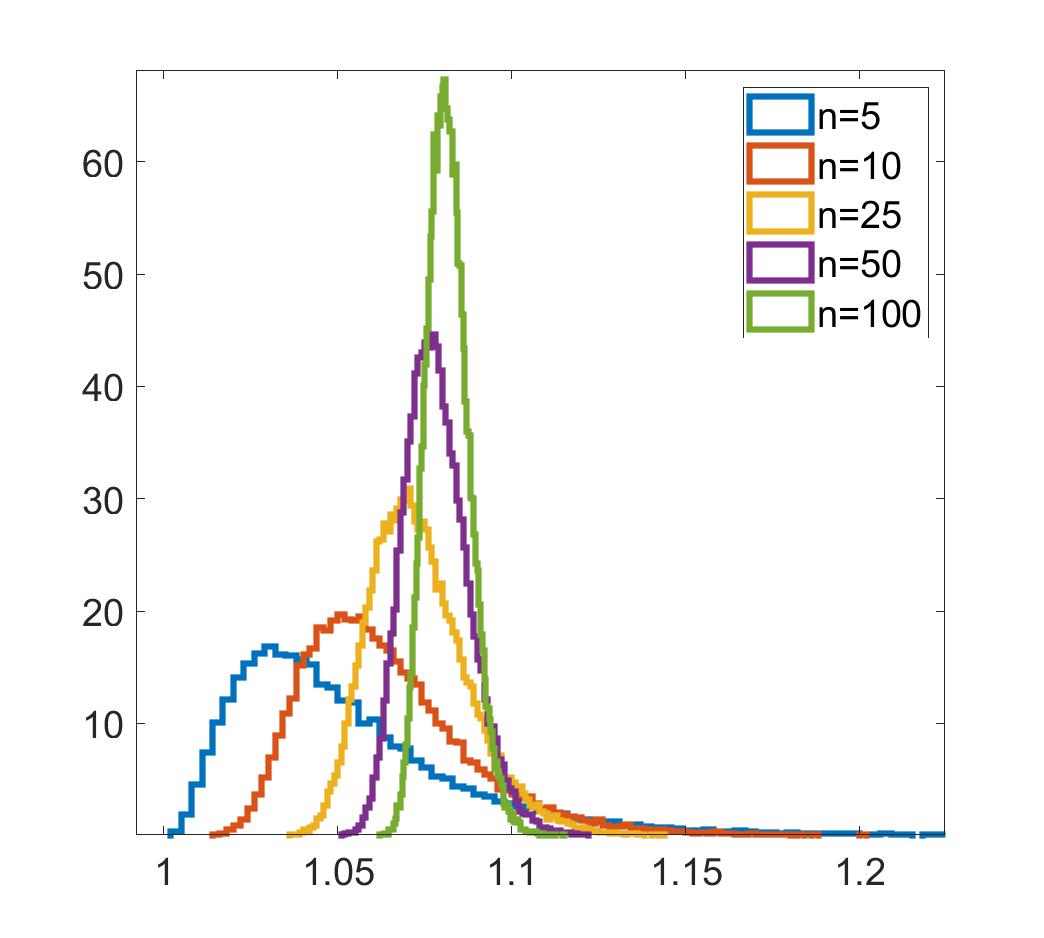}
    \\
    (a) & (b) & (c)
    \end{tabular}
    \caption{
    Empirical demonstration of the efficiency of the bounds
    \equ{increase_singval_average_bound} and \equ{conditioning_average_bound}
    using $r=2$.
    The figures show histograms of the fractions 
    (a)
    $
    [\singval_n^2(W_1+{\rm Id})]/
    [\singval_n^2(0)+\underline{\tau}_n+1]
$,
(b)
$
    [\singval_1^2(W_1)+\overline{\tau}_n+1]
    /
    [\singval_1^2(W_1+{\rm Id})]
$,
and
(c)
$
    \sqrt{({\singval_1^2(W_1)+\overline\tau_n+1)
%%%        }{\singval_n^2(W_1)+\underline\tau_n+1}}
        }/({\underline\tau_n+1}})
    /
        \CondNb(W_1+{\rm Id})
$
.
    Note that all values are above $1$ and actually quite close to $1$.}
    \label{fig:efficient_bounds}
\end{figure}

When comparing (i) and (ii),
it becomes apparent that the stated sufficient condition in (iii) is more easily satisfied 
than that of (iv).

For a a single residual layer \equ{training_one_layer}
with weights $W_1$
and a ResNet with absolute value
nonlinearities, we obtain the following.

\begin{cor}[High Probability Bound, Absolute Value]
  \label{cor:Abs_Value_average_bound}
  Let $W_1$ be as in Lemma~\ref{lem:Abs_Value_average_bound}.
  Let $D,D_1\in\mathcal{D}(m,n,-1)$ for some fixed $m$.
  Set $M_o=D_1W_1$ and recall Definition~\ref{def:tau}.

\noindent
  (i)
  In the Gaussian case, we have 
  $
    \singval_1(M_o)
    {\rightarrow}
%    \stackrel{\rm a.s.}{\rightarrow}
    1/(\sqrt{2}r)
$
%and \equ{lambda_increase_bounds}, both
almost surely.
%with probability $1$.

\noindent
(ii)  For any realization of $M_o$
  for which 
%%%  $S_1$ is dense in $[0,1]$ and 
$\singval_1(M_o)<1$, we have
  \begin{equation}
  \label{EQU:conditioning_average_bound_AbsValue_via_s1}
    \CondNb((M_o+{\rm Id})D)
    \leq
    \frac{\sqrt{\singval_1^2(M_o)+\overline\tau+1
        }}{1-\singval_1(M_o)}
    \leq
    \frac{1+\singval_1(M_o)}{1-\singval_1(M_o)}
    .
  \end{equation}

\noindent
    (iii)
    For any realization of $W_1$ with
%%%   $S_1$  dense in $[0,1]$, 
  $\underline{\tau}>-1$ 
%%%  and for which 
%%%  %the smallest singular value 
%%%  $ \singval_n(M_o+\PertVar {\rm Id})$
%%%  is piecewise simple for $\PertVar$ in $[0,1]$,
  the following bound holds
  \begin{equation}
  \label{EQU:conditioning_average_bound_AbsValue}
    \CondNb((M_o+{\rm Id})D)
    \leq
    \sqrt{\frac{\singval_1^2(M_o)+\overline\tau+1
%%%        }{\singval_n^2(M_o)+\underline\tau_n+1}}
        }{\underline\tau+1}}
    \leq
    \frac{\singval_1(M_o)+1}{\sqrt{\underline\tau+1}}
    .
  \end{equation}
\end{cor}

\begin{proof}
    Note that $M_0=D_1W_1$ is a random matrix with the same properties as $W_1$ itself, since $D_1$ is deterministic and changes only the sign of some of the entries of $W_1$. We may, therefore,
    apply Lemma~\ref{lem:Abs_Value_average_bound} to $M_o$ (replacing $W_1$). 
    % we obtain the bound on $\CondNb(D_1W_1+{\rm Id})$ in terms of $s_1(D_1W_1)$ and $\overline{\tau}(D_1W_1)$ etc.
    %Problem: $\overline{\tau}(D_1W_1)$ is the  largest eigenvalue of $(D_1W_1)^T(D_1W_1)=D_1(W_1^T+W_1)$ which may change...only its singvals remain unchanged, but the eigenvals may change!
    Also, $D$ does not alter the singular values:
    $\singval_i(M_o)=\singval_i(W_1)$ and
    $\singval_i((M_o+{\rm Id})D)=\singval_i(M_o+{\rm Id})$.
%\hfill$\Box$
\end{proof}

\noindent{\bf Probabilities of exceptional events}
\begin{itemize}
  \item{Clarification regarding $\underline{\tau}$ vs. $\underline{\tau}_n$ in the setting of 
  Corollary~\ref{cor:Abs_Value_average_bound}}:
  
  While we have $\singval_1(M_o)=\singval_1(W_1)$, one should note that $\overline\tau$
  (largest eigenvalue of $M_o^T+M_o$)
  may differ from $\overline\tau_n$
  (largest eigenvalue of $W_1^T+W_1$)
  for any particular realization. 
  However, since $M_o$ and $W_1$ are equal in distribution due to the special form of $D_1$,
  we have that
  $\overline\tau$ and $\overline\tau_n$, as well as 
  $-\underline{\tau}$ and $-\underline{\tau}_n$
  are all equal in distribution with distribution well concentrated around $1/r$.

  \item{Exception 
  $
  E_{\underline{\tau}_n}
  =
  E_{\underline{\tau}_n(r)}
  =
  \{\underline{\tau}_n(r)
  \leq-1\}$}:
  
  We have 
  \begin{equation}
  P[E_{\underline{\tau}_n}]\to 0
  \quad\mbox{as $n\to\infty$ for $r>1$.}
  \end{equation}
  Moreover,
  $P[E_{\underline{\tau}_n}]\approx0$ 
  for values as small as
  $r\geq 2$ and $n\geq 5$. 
  We are able to support this claim through the following 
  computations and simulations:
  Clearly, on the one hand,
  $P[\underline{\tau}_n>-1]
      =
    P[\tilde Z_n < (r-1)\sqrt{8} n^{2/3}]
    \to 1$ for $r>1$.
  On the other hand, for $r\geq2$ and 
  $n\geq 5$, we have $(r-1)\sqrt{8} n^{2/3}>8$
  and may estimate
  $P[\underline{\tau}_n>-1]
      =
    P[\tilde Z_n < (r-1)\sqrt{8} n^{2/3}]
    \geq
    P[\tilde Z_n < 8]
    =
    P[ Z_n < 8].
  $ 
  From Figure~\ref{fig:tracy-widom} it appears that 
  $P[Z_n<8]=1$ for $n\geq 5 $
  for all practical purposes.
  This is confirmed by our simulations 
  in the following sense:
  Every single one of the 500,000 random matrices simulated
  with $r=2$ satisfied $\underline{\tau}_n>-1$.

  \item{Exception
  $
  E_{\singval_1(W_1)}
  =
  \{\singval_1(W_1)\geq 1\}$}:
  
  We have that
  \begin{equation}
  P[
    E_{\singval_1(W_1)}
        ]\to 0
    \quad\mbox{as $n\to\infty$ for $r>1$.}
 \end{equation}
  Moreover,
  $P[
    E_{\singval_1(W_1)}
        ]\approx0$ 
  for values as small as
  $r\geq 2$ and $n\geq 5$.
  According to \equ{renorm_s1(W1)} and Theorem~\ref{thm:Op_norm_Gauss}, we have
 $       
    P[\singval_1(W_1)^2\geq 1]
      \leq
     P[
      Y_n
      \geq
      \sqrt{2}({2r^2}-1)
    n^{2/3}
    ]
    \to 0
    .
$
As demonstrated in Figure~\ref{fig:Yn_convergence},
$Y_n$ converges rapidly to a Tracy-Widom law
whence $P[E_{\singval_1(W_1)}]$ is small for $r\geq 2$ and modest $n$.
In the  setting of  Corollary~\ref{cor:Abs_Value_average_bound},
the probability of the event 
$
  E_{\singval_1(M_o)}
  =
  \{\singval_1(M_o)\geq 1\}$
tends to zero rapidly in a  similar fashion.

\end{itemize}

The bounds
\equ{increase_singval_average_bound} and \equ{conditioning_average_bound}
are surprisingly efficient.
The first upper bound of \equ{conditioning_average_bound}, for example,
is only about $10\%$ larger than $\CondNb(W_1+{\rm Id})$ for large $n$
(see Figure~\ref{fig:efficient_bounds}).

We see an explanation for this efficiency of the bounds to be rooted
in the fact that the singular values of random matrices tend to be
as widely spread as possible, meaning that the largest singular values
grow as fast as possible while the smallest singular values grow as slowly as possible.
While this maximal spreading is well known for symmetric
random matrices called ``Wigner matrices,'' we are not aware of any
reports of this behavior for non-symmetric matrices such as $W_1$.

\begin{figure}
    \centering
    (a)
    \includegraphics[width=0.45\linewidth]{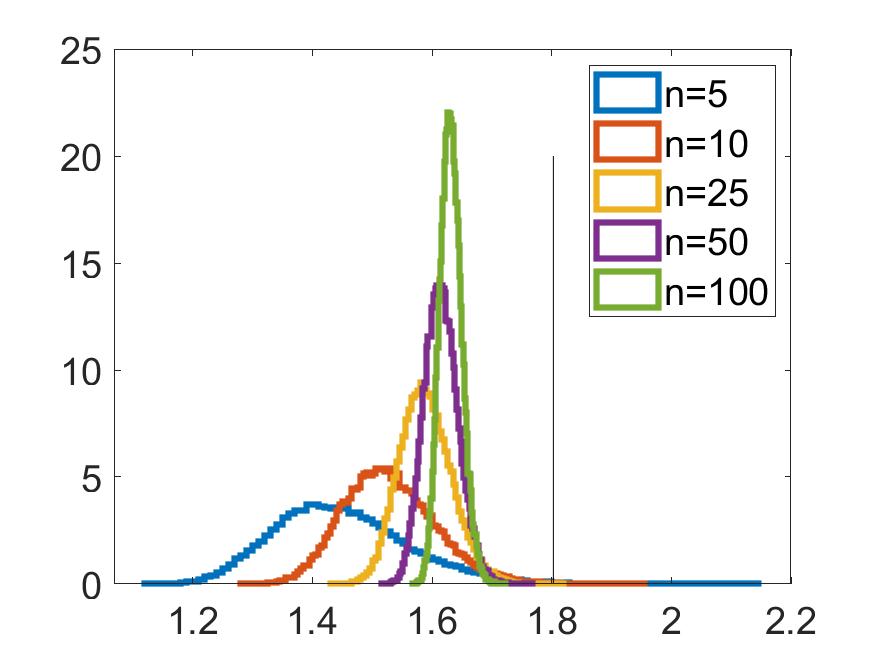}
    (b)
    \includegraphics[width=0.45\linewidth]{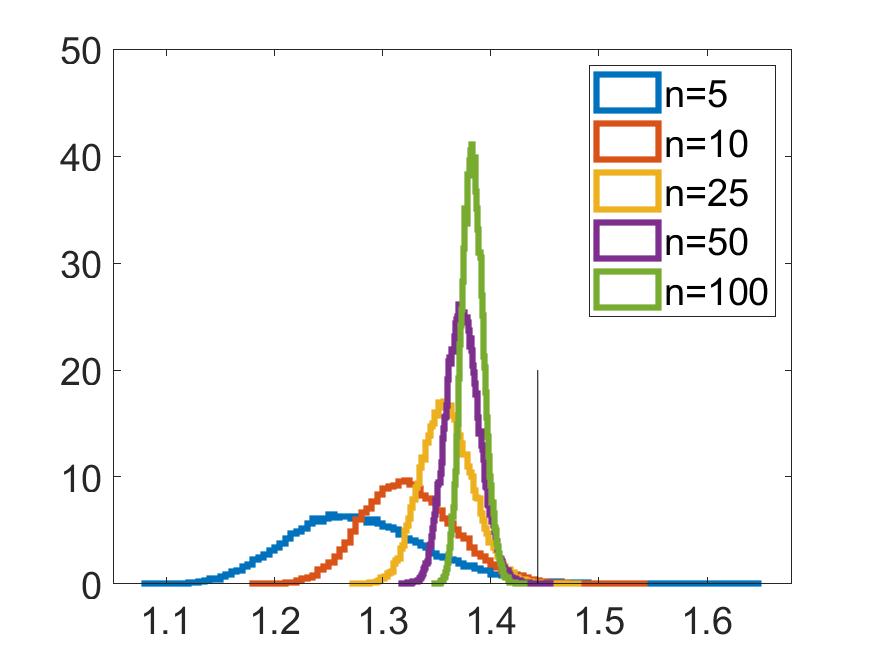}
    \caption{
    Empirical demonstration of the asymptotic bound \equ{conditioning_average_bound_asymptotics}.
    Histograms of the $\CondNb(W_1+{\rm Id})$ for (a) $r=2$ and (b) $r=3$
    together with the upper bound in form of a deterministic expression 
    from \equ{conditioning_average_bound_asymptotics} in black.
    Note that the values of $\CondNb(W_1+{\rm Id})$ lie asymptotically below the deterministic expression as indicated in
    \equ{conditioning_average_bound_asymptotics}.
    Also, recall that $\CondNb(W_1+{\rm Id})$ and
    $\CondNb(D_1W_1+{\rm Id})$ are equal in distribution in the setting of 
    Corollary~\ref{cor:Abs_Value_average_bound}.}
    \label{fig:asymptotic_bounds}
\end{figure}

For design purposes, an asymptotic bound for
the condition number $\CondNb(M(1)D)$ 
in terms of the network parameters might be more useful
than those in terms of the random realization as given 
in Corollary~\ref{cor:Abs_Value_average_bound}.

\begin{thm}[High Probability Bound in the Limit]
  \label{thm:chi_asymptotics}
  With the notation and assumptions of Theorem~\ref{thm:op_norm_edge_bound}
  and Lemma~\ref{lem:Abs_Value_average_bound},
  the upper bounds given there
  converge in distribution to constants as follows
  \begin{equation}
  \label{EQU:conditioning_average_bound_asymptotics}
    \sqrt{\frac{\singval_1^2(W_1)+\overline\tau_n+1
        }{\underline\tau_n+1}}
    \;\stackrel{\rm distr}{\rightarrow}\;
    \sqrt{
        \frac{r^2+r+1/2}{r^2-r}
    }
  \end{equation}
  and
  \begin{equation}
  \label{EQU:conditioning_average_bound_asymptotics_via_s1}
    \frac{\sqrt{\singval_1^2(W_1)+\overline\tau_n+1
        }}{1-\singval_1(W_1)}
    \;\stackrel{\rm distr}{\rightarrow}\;
    \frac{\sqrt{r^2+r+1/2}}{r-1/\sqrt{2}}
    .
  \end{equation}
\end{thm}

For a demonstration of the deterministic expression bounding
the condition number as in 
\equ{conditioning_average_bound_asymptotics}, see
Figure~\ref{fig:asymptotic_bounds}.
For $r=2$, for example, the right-hand side of \equ{conditioning_average_bound_asymptotics} becomes $1.8028$, while the right-hand side of 
\equ{conditioning_average_bound_asymptotics_via_s1} becomes the slightly larger $1.9719$. 

\begin{proof}
  Since $Z_n$ converges in distribution to a real-valued random variable $Z$,
  we can conclude that $Z_n/n^{2/3}$ converges in distribution to $0$.
  Indeed, for any $\eps>0$ and any $n_o\geq 1$, we have for $n\geq n_o$ that
  \begin{equation}
    P\left[|Z_n/n^{2/3}|\geq \eps\right]
    \leq
    P\left[|Z_n|\geq \eps n_o^{2/3}\right]
    \stackrel{n\to\infty}{\to}
%%%    P_{TW}[Z\geq \eps n_o^{2/3}]
    P\left[|Z|\geq \eps n_o^{2/3}\right]
    .
  \end{equation}
  From this we obtain for any $n_o\geq 1$ that
  \begin{equation}
    \limsup_{n\to\infty} P\left[|Z_n/n^{2/3}|\geq \eps\right]
    \leq
    P_{TW}\left[|Z|\geq \eps n_o^{2/3}\right],
  \end{equation}
  and letting $n_o\to \infty$ we conclude that
  $
  P[|Z_n/n^{2/3}|\geq \eps]
  \to
  0
  .
  $
  This establishes that $Z_n/n^{2/3}\stackrel{\rm distr}{\rightarrow} 0$.
  Similarly, we obtain $\tilde Z_n/n^{2/3}\stackrel{\rm distr}{\rightarrow} 0$.

  By the continuous mapping theorem, and since $r>1$, we obtain 
  \begin{equation}
    \sqrt{\frac{\singval_1^2(W_1)+\overline\tau_n+1
        }{\underline\tau_n+1}}
    \stackrel{\rm distr}{\rightarrow}
    \sqrt{
        \frac {1/(2r^2)+1/r+1
                }{-1/r+1}
                }.
    \end{equation}
Similarly,
\begin{equation}
\frac{\sqrt{\singval_1^2(W_1)+\overline\tau_n+1
        }}{1-\singval_1(W_1)}
    \stackrel{\rm distr}{\rightarrow}
    \frac {\sqrt{1/(2r^2)+1/r+1}
                }{1-1/(\sqrt{2}r)}.
    \end{equation}
Simple algebra completes the proof.
  ~\end{proof}

Our simulations suggest that actually  
$\overline{\tau}_n / \singval_1(W_1)$ converges in distribution to $\sqrt{2}$ as $n\to\infty$, implying that we should see 
$\overline{\tau}_n \simeq \sqrt{2} \singval_1(W_1)$
for most matrices in real world applications.

We turn now to the condition number of the relevant matrix $(D_1W_1+{\rm Id})D$
for a ResNet with ReLU activations
that are represented by
$D_1\in\mathcal{D}(\tilde m,n,0)$
and
$D\in\mathcal{D}(m,n,0)$.
We address the case of a single  residual layer
following the trained layer (cf.~\equ{training_one_layer} with $p=1$).

To this end, we let
  \begin{equation}\label{EQU:def_nu}
    \nu
    =
    m+\tilde m-\frac{m\,\tilde m}n,
  \end{equation}
  which can be considered as a measure of
  the extent to which these two
  nonlinearities affect the output in combination.

\begin{thm}[High Probability Bound, ReLU]
  \label{thm:ReLU_average_bound}
~\\
  Let
  $D_1\in\mathcal{D}(\tilde m,n,0)$
  and
      $D\in\mathcal{D}(m,n,0)$
  with
  $1 \leq m + \tilde m $.
    Then, $1\leq \nu \leq n$.
 Choose $\theta$ such that $4<\theta\leq 2 \sqrt n$,
 and let $\Std_n$ and $W_1$ be as in 
  Lemma~\ref{lem:Abs_Value_average_bound},
  with $r$ defined as below and 
   with Gaussian entries for $W_1$.

    (1) Choose $r>(2+2\nu+\theta)$. Then,
    with 
  probability at least
  $1-2\exp(-\theta^2/2)
  - 
  P[E_{\singval_1(W_1)}]$, we have
  \begin{equation}
  \label{EQU:conditioning_ReLU_average_bound_via_s1}
    \CondNb((D_1W_1+{\rm Id})D))
    \leq
        2\frac{
            1+1/r
            }{
            \left(
            {1-(2+2\nu+\theta)/r}
            \right)
            }
    .
  \end{equation}

  (2) Choose $\rr>1$ and $r>({\rr}+{\rr}\nu+\theta)$.
  Then,
  with probability
  at least
  $1-2\exp(-\theta^2/2)
  - P[E_{\underline{\tau}_n(\rr)}]
  $,
we have
  \begin{equation}
  \label{EQU:conditioning_ReLU_average_bound}
    \CondNb((D_1W_1+{\rm Id})D))
    \leq
        2\frac{
            1+\singval_1(W_1)
            }{
            \left(
            {1-({\rr}+{\rr}\nu+\theta)/r}
            \right)
            }
    .
  \end{equation}
\end{thm}

We recall that, according to our simulations,
 $       
P[  E_{\singval_1(W_1)}]
=
P[\singval_1(W_1)^2\geq 1]
      \leq
     P[
      Y_n
      \geq
      \sqrt{2}({2r^2}-1)
    n^{2/3}
    ]
$
and
$
    P[E_{\underline{\tau}_n(\rr)}]
    =
    P[Z_n > ({\rr}-1)\sqrt{8} n^{2/3}]
$
are negligible for $r,{\rr}\geq 2$ and modest $n$.

We mention further that the restriction $4\leq \theta\leq 2 \sqrt n$
is solely for simplicity of the representation
and can be dropped at the cost of more complicated formulae (see proof).
In the given form, useful bounds on the probability are obtained
even with modest $n$ as low as $n=7$.

Clearly, the bound \equ{conditioning_ReLU_average_bound_via_s1} on the condition number $\CondNb$ can be forced arbitrarily close to $2$
by choosing $r$ sufficiently large.
In fact, bounds can be given that come
arbitrarily close to $1$, as becomes
apparent from \equ{stronger_bound_Gauss_via_s1}
and (\ref{EQU:stronger_bound_Gauss}) in the proof.

\begin{proof}
Let ${\AnyM}=W_1+{\rm Id}$ with singular values $\singval_i$
and ${\AnyMt}=(D_1W_1+{\rm Id})D$  with singular values $\tilde\singval_i$.
Clearly, ${r}>{\rr}>1$.

We proceed in several steps, addressing first \equ{conditioning_ReLU_average_bound}.

\emph{(i) Sure bound on the sum of perturbations.}
We cannot apply Proposition~\ref{prop:ReLU_multiple_output}
as we did in Theorem~\ref{thm:conditioning_ReLU_hard_bound},
since we need to take into account also the effect of the
left-factor $D_1$.
Due to Corollary~\ref{cor:Frobenius_bound} we have
\begin{equation}
    \sum_{i=1}^{n}
        |\singval_i - \tilde\singval_i |^2
    \leq
    \|{\AnyMt}-{\AnyM}\|^2_2
    =
    \|(D_1W_1D-W_1)+(D-{\rm Id})\|^2_2.
\end{equation}
We find $k=n(m+\tilde m)-m\tilde m$
non-zero terms in the first matrix $(D_1W_1D-W_1)$
that are of the form $W_1[i,j]$.
The second matrix $(D-{\rm Id})$
is diagonal
with exactly $m$ non-zero terms in the diagonal
that are all equal to $-1$.
Thus, the Frobenius norm $\|{\AnyMt}-{\AnyM}\|^2_2$
consists of exactly $k$ terms of the form $W_1[i,j]^2$,
which we collect into the variable $V_1$;
several terms of the form $2W_1[i,j]$, which
we collect into the variable $V_2$;
and finally $m$ terms equal to $1$.
Note that $V_2$ consists of at most $m$
terms, depending on how many diagonal positions are non-zero
in both $(D_1W_1D-W_1)$ and $(D-{\rm Id})$.
We obtain
\begin{equation}
    \sum_{i=1}^{n}
        |\singval_i - \tilde\singval_i |^2
    \leq
    \|{\AnyMt}-{\AnyM}\|^2_2
    =
    V_1+V_2+m.
\end{equation}
By elementary calculus we have
$n\leq k\leq n^2$.
Together with 
$m,\tilde m \leq n$ and $m+\tilde m\geq 1$,
this implies that $1\leq \nu\leq n$.

\emph{(ii) Concentration of the values of $V_1$, $V_2$ and $\underline{\tau}_n$.}
The variable $V_2$ is zero-mean Gaussian with a variance of
at most $4m\Std_n^2$.
To formulate where the values of $V_2$ concentrate, we will use
a parameter $\theta$.
A standard result says that
\begin{equation}\label{EQU:GaussTail}
  P[V_2>\theta \cdot 2\Std_n\sqrt{m}]
  \leq
  P[X>\theta \cdot \Std_n]
  \leq
  \exp(-\theta^2/2)
  .
\end{equation}

Since $W_1[i,j]/\Std_n$ are independent standard normal variables,
the variable $Y=V_1/\Std_n^2$ follows a Chi-square distribution with
$k$ degrees of freedom.
One of the sharpest bounds known for the tail
of a Chi-squared distribution
is that of Massart and Laurent \cite[Lemma 1, pg.~1325]{Laurent-Massart}.
A corollary of their bound yields
\begin{equation}
    P[Y\geq k + 2 \sqrt{kq} +2q]
    \leq \exp(-q).
\end{equation}
Choosing $q=\theta^2/2$ for symmetry with $V_2$, we obtain
\begin{equation}\label{EQU:ChiTail}
  P[V_1>\Std_n^2(k+2\theta\sqrt{k/2}+ \theta^2)]
  \leq
  \exp(-\theta^2/2)
  .
\end{equation}

Finally, since $-\underline{\tau}_n$ and $\overline{\tau}_n $
are equal in distribution, we obtain
\begin{equation}\label{EQU:tauTail}
\mbox{\small
$
    P[-\underline{\tau}_n > {\rr} \Std_n\sqrt{8n}]
    =
    P[-\underline{\tau}_n > {\rr} /  r]
    =
    P[Z_n > ({\rr}-1)\sqrt{8} n^{2/3}]
    =
    P[E_{\underline{\tau}_n(\rr)}],
$}
\end{equation}
which is negligibly small for ${\rr}\geq 2$.
%\rolf{replace with the s1(W1) version ... (74) 
%could be tons of changes}  

In summary, 
using 
(\ref{EQU:GaussTail}), (\ref{EQU:ChiTail})
and \equ{tauTail}
we find that, with probability at least 
$
1
-2\exp(-\theta^2/2)
-P[E_{\underline{\tau}_n(\rr)}]
%-P[Z_n > ({\rr}-1)\sqrt{8} n^{2/3}]
%%%-P[E_{\rm \singval_n(W_1)}]
$,
%\rolf{replace with the s+(W1) version ... (74) }  
we have  simultaneously  (\ref{EQU:increase_singval_average_bound})
as well as
\begin{eqnarray}
    \label{EQU:concentration}
    V_2&\leq&\theta \, 2\Std_n\sqrt{m}
    \nonumber
    \\
    \nonumber
    V_1&\leq&\Std_n^2(k+2\theta\sqrt{k/2}+ \theta^2)
    \\
    -\underline{\tau}_n &\leq& {\rr} \Std_n\sqrt{8n}=\rr/r.
\end{eqnarray}

\emph{(iii) Sure bound on individual perturbations.}
As in earlier proofs, we note that the bulk
of the sum of deviations
$|\singval_i - \tilde\singval_i |^2$
stem from some $m$ singular values of ${\AnyM}$
being changed to zero by $D$,
compensating the constant term $m$
on the right.
%in the bound (\ref{EQU:bound-GaussCase}).
More precisely,
%\rolf{replace with the s+(W1) version ... (74) 
%$\singval_i^2\geq \singval_n^2\geq 1-s1(W1)$
%}  
$\singval_i^2\geq \singval_n^2\geq 1+\underline{\tau}_n$
%%$\singval_i^2\geq 1-\Std_n\sqrt{8n}$
by (\ref{EQU:increase_singval_average_bound})
while
$\tilde\singval_i=0$ for $i=n-m+1,\ldots, n$.
Thus,
\begin{equation}
    \sum_{i=n-m+1}^{n}
        |\singval_i - \tilde\singval_i |^2
    \geq
    m\, (1+\underline{\tau}_n)
%%    m\cdot (1-\Std_n\sqrt{8n})
    ,
\end{equation}
implying for any $1\leq j \leq n-m$ that
\begin{eqnarray}
    |\singval_j - \tilde\singval_j |^2
    &\leq&
    \sum_{i=1}^{n-m}
        |\singval_i - \tilde\singval_i |^2
    \leq
    V_1+V_2+m
    -
    m\cdot (1+\underline{\tau}_n)
%%    m\cdot (1-\Std_n\sqrt{8n})
    \nonumber \\
    &=&
    V_1+V_2 - m\underline{\tau}_n.
%%    V_1+V_2+m\Std_n\sqrt{8n}
\end{eqnarray}
%$$
%    \leq
%    ((m+\tilde m)n\Std_n^2 + \theta \Std_n^2\sqrt{2(m+\tilde m)n})
%    +
%    2  \theta \Std_n\sqrt{m}
%    +
%    m
%    -
%    m\cdot (1-\Std_n\sqrt{8n})
%$$

{\em (iv) Probabilistic bound on the perturbation.}
Combining the three bounds \equ{concentration} that
hold simultaneously with the indicated probability,
we have
\begin{eqnarray}
    |\singval_j - \tilde\singval_j |^2
    &\leq&
    \Std_n^2(k+2\theta\sqrt{k/2} + \theta^2)
    +
    \theta \cdot 2\Std_n\sqrt{m}
    +m{\rr}\Std_n\sqrt{8n}
    \nonumber \\
    &=&
    \Std_n\sqrt{8n}
    \left[
        \Std_n\sqrt{8n}
        \frac{k+\sqrt{2k}\theta+\theta^2}{8n}
        +
        2\theta \sqrt{\frac{m}{8n}}
        +
        m {\rr}
    \right].
\end{eqnarray}
Here, we factored $\Std_n\sqrt{8n}$,
since it simplifies to $1/r$.

{\em (v) Simplifications for convenience.}
There are many ways to simplify this bound.
For convenience of presentation, here is one. Assuming that $\theta\leq 2\sqrt n$,
we have $\theta\leq 2\sqrt k$ and
\begin{equation}
        \frac{k+\sqrt{2k}\theta+\theta^2}{8n}
        \leq
        \frac{k+\sqrt{8}k+4k}{8n}
        \leq
        \frac{8k}{8n}
        =
        \nu
        .
\end{equation}
Using $\Std_n\sqrt{8n}\, \nu\leq 1$,
$m /n \leq 1$
as well as
$\nu=m+\tilde m(1-m/n)\geq m$
and
%$1/4+1/\sqrt{2}\leq 1$,
$1+\theta/\sqrt 2\leq \theta$,
we obtain
\begin{eqnarray}
    |\singval_j - \tilde\singval_j |^2
   &\leq&
    \Std_n\sqrt{8n}
    \left[
    1+\theta/\sqrt 2 + {\rr}\nu
    \right]
   \leq
    \Std_n\sqrt{8n}
    \left[
    {\rr}\nu +\theta
    \right]
   .
\end{eqnarray}
Noting that $\sqrt a-\sqrt b=(a-b)/(\sqrt a + \sqrt b)
\geq (a-b)/(1+\sqrt b)
\geq (a-b)/2$
whenever $0<a,b<1$ and using
(\ref{EQU:increase_singval_average_bound}), $s_j\geq s_n$
and \equ{tauTail}, we have
\begin{eqnarray}
    \nonumber
    \tilde\singval_j
    &\geq&
    \singval_j
    -
    |\singval_j - \tilde\singval_j |
    \geq
    \sqrt{1-{\rr}\Std_n\sqrt{8n}}
    -
    \sqrt{  \Std_n\sqrt{8n}({\rr}\nu+\theta)}
    \\
    &\geq&
        \frac {1-\Std_n\sqrt{8n}({\rr}+{\rr}\nu+\theta)}
            {
                1
                +
                \sqrt{ \Std_n\sqrt{8n} ({\rr}\nu+\theta)}
            }
            \geq
        \frac 1 2
            \left(
            {1-\Std_n\sqrt{8n}({\rr}+{\rr}\nu+\theta)}
            \right)
    .
\end{eqnarray}
Replacing $\Std_n\sqrt{8n}$ by $1/r$,
this becomes
\begin{eqnarray}
    \tilde\singval_j
    &\geq&
    \label{EQU:stronger_bound_Gauss}
        \frac {1-({\rr}+{\rr}\nu+\theta)/r}
            {
                1
                +
                \sqrt{ ({\rr}\nu+\theta)/r}
            }
            \geq
        \frac 1 2
            \left(
            {1-({\rr}+{\rr}\nu+\theta)/r}
            \right)
    .
\end{eqnarray}

This simple form is for convenience. Tighter bounds can be chosen along the lines above.
This takes care of the denominator of $\CondNb$.
For the numerator, we estimate using well-known
rules of operator norms (or alternatively
Lemma~\ref{lem:ReLU_decrease} and \equ{Weyl_bound}):
\begin{equation}
    \tilde\singval_1
    \leq
    (\|D_1\|\,\|W_1\|+\|{\rm Id}\|)\|D\|
    \leq
    \|W_1\|+1
    =
    \singval_1(W_1)+1
    .
\end{equation}

{\em (vi) Let us now turn to \equ{conditioning_ReLU_average_bound_via_s1}.}
We may proceed as in steps (iii) through (v) with minor modifications.
First, replace \equ{tauTail} by
\begin{equation}
    P[\singval_1(W_1)\geq 1/r]
    =
    P[\singval_1(W_1)^2\geq 1/r^2]
      \leq
     P[
      Y_n
      \geq
      \sqrt{2}  n^{2/3}
    ]
    \to 0
    .
\end{equation}
Thus, we have
\begin{eqnarray}
    \label{EQU:concentration_via_s1}
    V_2&\leq&\theta \cdot 2\Std_n\sqrt{m}
    \nonumber
    \\
    \nonumber
    V_1&\leq&\Std_n^2(k+2\theta\sqrt{k/2}+ \theta^2)
    \\
    \singval_1(W_1)&\leq& 1/r = \Std_n\sqrt{8n}
\end{eqnarray}
with probability at least 
$
1
-2\exp(-\theta^2/2)
- P[\singval_1(W_1)\geq 1/r]
\geq
1
-2\exp(-\theta^2/2)
- P[
      Y_n
      \geq
      \sqrt{2}  n^{2/3}
]
$,
where we use the last expression for convenience of presentation,
as it relates to the exceptional event $E_{\singval_1(W_1)}$.

Replacing 
$\singval_i^2\geq \singval_n^2\geq 1+\underline{\tau}_n$
by
$\singval_i^2\geq \singval_n^2\geq (1-\singval_1(W_1))^2
\geq 1-2\singval_1(W_1)$, we obtain the sure bound similar to (iii)
\begin{eqnarray}
    |\singval_j - \tilde\singval_j |^2
    &\leq&
    V_1+V_2 +2m\singval_1(W_1)
\end{eqnarray}
and from this, computing as in (iv) and (v), the bound
\begin{eqnarray}
    |\singval_j - \tilde\singval_j |^2
   &\leq&
    \Std_n\sqrt{8n}
    \left[
    1+\theta/\sqrt 2 + 2\nu
    \right]
   \leq
    \Std_n\sqrt{8n}
    \left[
    2\nu +\theta
    \right]
   ,
\end{eqnarray}
which holds with probability at least 
$1
-2\exp(-\theta^2/2)
- P[\singval_1(W_1)\geq 1]
$.

Noting that $ a-\sqrt b=(a^2-b)/( a + \sqrt b)
\geq (a^2-b)/(a+1)
\geq (a^2-b)/2$
for $0<a,b<1$ 
and replacing $\Std_n\sqrt{8n}$ by $1/r$,
we obtain, using
$s_j\geq s_n\geq 1-\singval_1(W_1)\geq 1-1/r$
\begin{eqnarray}
    \nonumber
    \tilde\singval_j
    &\geq&
    \singval_j
    -
    |\singval_j - \tilde\singval_j |
    \geq
    (1-1/r)
    -
    \sqrt{  (2\nu+\theta)/r}
    \\&\geq&
    \label{EQU:stronger_bound_Gauss_via_s1}
        \frac {(1-1/r)^2-(2\nu+\theta)/r}
            {
                1-1/r
                +
                1
            }
            \geq
        \frac 1 2
            \left(
            {1-(2+2\nu+\theta)/r+1/r^2}
            \right)
    .
\end{eqnarray}
Note that the very first bound 
$    (1-1/r)
    -
    \sqrt{  (2\nu+\theta)/r}
$
in \equ{stronger_bound_Gauss_via_s1}
can be made arbitrarily close to $1$ by increasing $r$.
Dropping the last term $1/r^2$ for convenience and noting that
\begin{equation}
    \tilde\singval_1
    \leq
    \|W_1\|+1
    =
    \singval_1(W_1)+1
    \leq 1+1/r
\end{equation}
in the setting of step (vi),  the proof is complete.
%\hfill$\Box$
\end{proof}

A similar argument can be given for Uniform random entries instead of Gaussian ones
using the Central Limit Theorem and the Berry-Esseen Theorem instead of the
tail behavior of Gaussian and Chi-Square variables.
The statements and arguments will be more involved.

Also, similarly to the hard bounds it is possible to find bounds on the condition number
for leaky-ReLU, implying again that choosing $\eta$ close to $1$ enables us to bound
the condition number.

\subsection{Advantages of residual layers}
\label{subsec:advantages}

\noindent
{\bf Volatility of the piecewise-quadratic loss landscape.}~
The preceding sections provide results on the condition number $\CondNb$ of the relevant matrix $M(1)D$ for ResNets
(recall Corollary~\ref{cor:standard_quadratic_form}). They establish conditions under which $\CondNb(M(1)D)$ can be bounded. To be more specific, let us provide some concrete examples assuming just one subsequent layer after the trained one, i.e., set $p=1$ in \equ{training_multi_layer}.

Consider first a \emph{deterministic setting}.
Assume that the entries of the $n \times n$ weight matrix $W_1$ are bounded by $1/(3n)$.
Then the condition number of the relevant matrix $\CondNb(M(1)D)$ amounts to at most $2$
for an absolute value activation nonlinearity according to (\ref{EQU:conditioning_hard_bound}). 
If in addition not more than $1/4$ of the outputs of the trained layer are affected by the activation,  then the condition number $\CondNb(M(1)D)$ is bounded by $8$
for a ReLU activation according to (\ref{EQU:conditioning_ReLU_hard_bound_small_m}).
Under more technical assumptions, bounds on the condition number are also available for the leaky-ReLU activation.
Further bounds are available without restrictions on the outputs but with more severe restrictions on the entries.
Note that no a priori bounds on $\CondNb(M(0)D)$ are available,
even if a bound on the entries of $M(0)$ is imposed, 
as the matrices
$M(0)={\rm diag}(1/n,1/(2n),\ldots,1/n^2)$ 
and similar examples demonstrate.

Consider now a \emph{random weight setting} and assume that the standard deviation of the entries of the weight matrix $W_1$ are bounded on the order of $1/\sqrt n$.  Then the condition number $\CondNb(M(1)D)$  can be bounded for the absolute value activation with known probability using (\ref{EQU:conditioning_average_bound}).
For the ReLU activation,  similar bounds are available as for the absolute value.
Again, as in the deterministic case, the restrictions on the standard deviation of the weights become more severe the more outputs are affected by the ReLU (see (\ref{EQU:conditioning_ReLU_average_bound})).

To the best of our knowledge, similar bounds are not available for the random matrix $M(0)D$ corresponding to a ConvNet.
Rather, the relevant condition number increases with $n$,
as we put forward in the proof of the next result
(see Theorem~\ref{thm:Adding_Id_improves_CondNb}).
%\rolf{please verify the following claim:}

To better appreciate the impact of this difference between ConvNets and ResNets,
recall that the loss surface is \emph{piecewise} quadratic, meaning that, close to the optimal values $W$, regions with quite different activities of the nonlinearities and thus with different $D$ and $D_k$ may exist. 
Thus, lacking any bounds, the condition number $\CondNb(M(0)D)$ of a
ConvNet may very well exhibit large fluctuations, resulting in a volatile shape for the loss landscape.
For ResNets, the relevant condition number $\CondNb(M(1)D)$ is not only bounded via 
Theorem~\ref{thm:ReLU_average_bound}, but
the bounds can also easily be rendered 
independent of the activities of the activation nonlinearities, i.e., independent of $D$ and $D_1$. 
Indeed, obviously we can replace $\nu$ everywhere by $n$
and the statements still hold
(cf.~\equ{stronger_bound_Gauss_via_s1}).

\begin{cor}[High Probability Bound, ReLU, for all regions]
{  \label{cor:ReLU_average_bound}
~\\
 Choose $\theta$ such that $4<\theta\leq 2 \sqrt n$,
 and   $r>(2+2n+\theta)$. 
 Let $\Std_n$ and $W_1$ be as in 
  Lemma~\ref{lem:Abs_Value_average_bound},
   with Gaussian entries for $W_1$.
 Set $C=1+1/r$ and
$c=(1-(2+2n+\theta)/r)/2$.

Then,  
   for any $D_1\in\mathcal{D}(\tilde m,n,0)$
  and
      $D\in\mathcal{D}(m,n,0)$
  with
  $1 \leq m + \tilde m $, 
    with 
  probability at least
  $1-2\exp(-\theta^2/2)
  - 
  P[E_{\singval_1(W_1)}
]
  $,
  we have
  \begin{equation}
  \label{EQU:conditioning_ReLU_average_bound_indep_of_nu}
  c\leq \singval_i((D_1W_1+{\rm Id})D)\leq C
  \quad\mbox{and}\quad
    \CondNb((D_1W_1+{\rm Id})D))
    \leq
    \frac C c.
  \end{equation}
}
\end{cor}

The guarantees provided for the relevant matrix for a ResNet in terms of (\ref{EQU:conditioning_ReLU_average_bound_indep_of_nu}) assure that the resulting piecewise-quadratic loss surface cannot change shape too drastically.

{\bf Influence of the training data.}~
As mentioned earlier, we focus in our results and discussions 
mainly on how  the weights and nonlinearities of the 
layers, i.e., $M(\PertVar)D$  influence  the loss shape, with the exception of Subsection~\ref{subsec:batches}. This makes sense, since
the influence of the data is literally factored out via
Corollary~\ref{cor:standard_quadratic_form} with a contribution
to the condition number 
that is independent of the design choice of ConvNet versus ResNet.

Using mini-batches for learning affects the loss shape in two ways. First, 
it leads to smaller regions as put forward 
in Subsection~\ref{subsec:batches}
(cf.~\equ{loss_batch}) and thus to potentially increased 
volatility of the loss landscape. 
Second, Corollary~\ref{cor:loss_batch_form}
implies that using mini-batches alleviates the 
influence of data and features 
${\inlayert}^{(g)}$ (input to the trained layer $f$) in the sense that 
the ``data factor'' of the condition number $\CondNb(Q)$
has a less severe impact, since feature values are averaged
over $G$ data points.

As mentioned above, these two kinds of influences of the data
on the condition number are the same for both ConvNets and ResNets.
Nevertheless, some difference prevails.
For ConvNets, the ``layer factor'' $\CondNb(M(0)D)$ 
of the condition number $\CondNb(Q)$ leads mini-batches inducing even more 
loss landscape volatility due to the smaller regions.
For ResNets, the uniform bound
(\ref{EQU:conditioning_ReLU_average_bound_indep_of_nu}) 
carries over to mini-batches, and hence the smaller regions do not cause
increased condition numbers and loss landscape volatility.
{\em Again, no such guarantees are available for ConvNets.}

\noindent
{\bf Eccentricity of the loss landscape.}~
Intuition from our singular values results says that {\em the loss landscape of a ResNet should be less eccentric than that of a ConvNet.} We now provide a theoretical result as well as a numerical validation to support this point of view.

\begin{lem}
  \label{lem:Adding_Id_improves_CondNb}
  Let $W_1$ be as in Lemma~\ref{lem:Abs_Value_average_bound},
  and let $D_1\in\mathcal{D}(m,n,-1)$ for some $m$.
  Let $M_o=D_1W_1$.
Then, asymptotically as $n\to\infty$
\begin{equation}
\label{EQU:Adding_Id_improves_CondNb}
    P[
\CondNb(M_o)
\geq    
(1+\singval_1(M_o))/(1-\singval_1(M_o)) 
>0
    ]
    \to 1.
\end{equation}
\end{lem}

Recalling that absolute value nonlinearites do not change singular values (recall Corollary~\ref{cor:Abs_Value_average_bound}), we obtain the following.

\begin{thm}[Residual Links Improve Condition Number]
  \label{thm:Adding_Id_improves_CondNb}
  Let $W_1$ be as in Lemma~\ref{lem:Abs_Value_average_bound},
  and let $D,D_1\in\mathcal{D}(m,n,-1)$ 
  for some $m$,
  corresponding to an absolute value activation nonlinearity.
Consider a realization of $W_1$ such that \equ{Adding_Id_improves_CondNb} holds.
Then,  adding a residual link improves the condition number of the matrix relevant for learning
\begin{equation}
\label{EQU:Res_beats_Conv_AbsValue}    
    \CondNb(D_1W_1D)\geq    
    \CondNb((D_1W_1+{\rm Id})D)
.
\end{equation}
\end{thm}
By Lemma~\ref{lem:Adding_Id_improves_CondNb}, the condition on $W_1$ is satisfied with probability asymptotically equal to $1$.

\begin{proof}
To establish  Lemma~\ref{lem:Adding_Id_improves_CondNb},
it suffices to observe that 
$\singval_1(M_o)$ converges almost surely to $1/(\sqrt{2}r)<1$, as well as
\equ{conditioning_average_bound_AbsValue_via_s1}
and that $\CondNb=\CondNb(D_1W_1)$ is asymptotically of the order
$n$.

To provide a rigorous argument, 
note that $D_1W_1$ is a Gaussian random matrix just like 
$W_1$ itself, since $D_1$ only changes some of the signs of $W_1$'s entries. Write $\CondNb=\CondNb(D_1W_1)$ for short.

Applying  a result due to Edelman \cite[Theorem 6.1]{Edelman} to the matrix $D_1W_1$ (multiplying a matrix by a constant does not change its condition number), we find that
$\CondNb/n$ converges in distribution to a random variable $K$ with density (for $t>0$)
\begin{equation}
    f_K(t)
    =
    \frac{2t+4}{t^3}\exp(-2/t-2/t^2).
\end{equation}
Note that $f_K$ is $\mathcal{C}^{1}(\reals^+)$ and $f_K(t)\to 0$ as $t\to0$. It follows that $P[K<\eps]\to 0$
as $\eps\to 0$.
We conclude that
\begin{equation}
    P[\CondNb<n\eps]
    =
    P[\CondNb/n<\eps]
    \to
    P[K<\eps]
\end{equation}
as $n\to\infty.$

Now let $\delta>0$ be arbitrary. Choose $\eps>0$ small enough such that 
$    P[K<\eps]<\delta/2
$, and choose $n$ large enough such that 
$P[\CondNb<n\eps]<\delta$. Increase $n$ if necessary and choose $r'<r$ close enough to $1$ such that 
$n\eps=(1+1/r')/(1-1/r')$.

Since $D_1W_1$ has the same properties as $W_1$, it follows that
$\singval_1(M_o)$
is equal in distribution to 
$\singval_1(W_1)$ from  \equ{renorm_s1(W1)}.
Then, since $Y_n$ converges in distribution, we have that
\begin{eqnarray}
    P[\singval_1(M_o)>1/r']
    &=&
    P[\singval_1(M_o)^2>(1/r')^2]
    \leq
    P[1+Y_n\frac{1}{\sqrt{2}\cdot n^{2/3}}>2(r/r')^2]
    \nonumber\\
    &=&
    P[Y_n>(2(r/r')^2-1){\sqrt{2}\cdot n^{2/3}}]
    \to 0
\end{eqnarray}
as $n\to\infty$.
Note that if $\singval_1(M_o)\leq 1/r' $ 
then 
$ (1+\singval_1(M_o))/(1-\singval_1(M_o)) 
\leq (1+1/r')/(1-1/r')$. 
As we just showed,  the probability of this happening tends to $1$ as $n\to\infty$.

Similarly, as in our other proofs, we note that
$
P[\CondNb<n\eps
\mbox{ or } 
(1+\singval_1(M_o))/(1-\singval_1(M_o)) 
 >(1+1/r')/(1-1/r')
]
$
is at most equal to the sum of both probabilities and thus at most $2\delta$ as $n\to\infty$. 
We conclude that 
\begin{equation}
P[\CondNb\geq n\eps
~\mbox{ and }~ 
(1+\singval_1(M_o))/(1-\singval_1(M_o)) < n\eps]
\end{equation}
is asymptotically at least $1-2\delta$. 
With $\delta>0$ arbitrary, this establishes Lemma~\ref{lem:Adding_Id_improves_CondNb}.

The theorem follows now from \equ{conditioning_average_bound_AbsValue_via_s1}.
%\hfill$\Box$
\end{proof}

\begin{figure}%[t]
\begin{center}
\begin{tabular}{ccc}
largest singular value &\hspace*{5mm}& smallest non-zero singular value
\\ 
$s_1(\PertVar)=s_1(M_o+\PertVar\,{\rm Id})$ && 
$s_*(\PertVar)=s_*(M_o+\PertVar\,{\rm Id})$ 
\\
\includegraphics[width=0.40\linewidth]{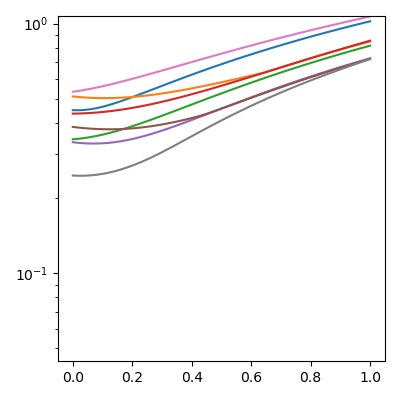} &&
\includegraphics[width=0.40\linewidth]{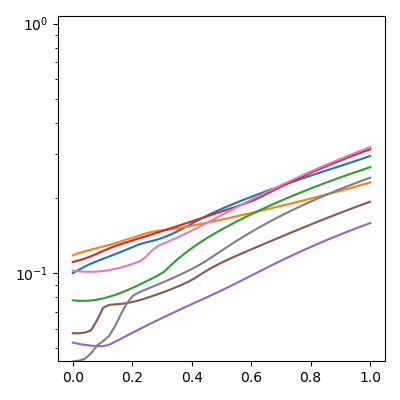}
\\[-1mm]
$\rho$ && $\rho$
\\[3mm]
(a) && (b) 
\\[5mm]
$\kappa(M_o+\PertVar{\rm Id})=s_1(\PertVar)/s_*(\PertVar)$ &&
$s_1(\PertVar)/s_i(\PertVar)$ 
\\
\includegraphics[width=0.40\linewidth]{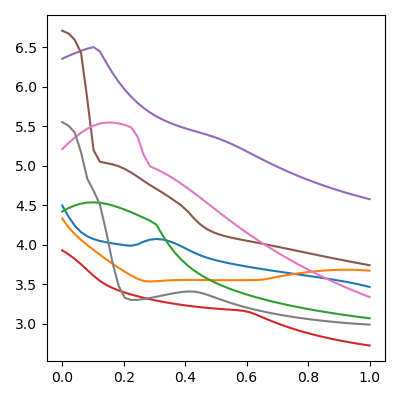} &&
\includegraphics[width=0.40\linewidth]{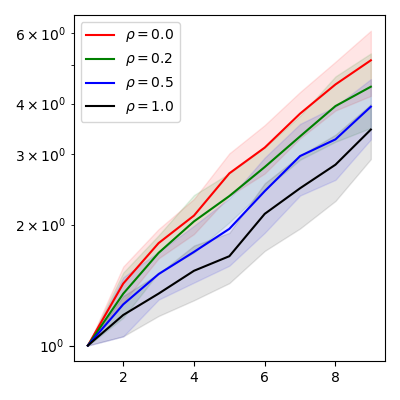}
\\[-1mm]
$\rho$ && $i$
\\[3mm]
(c) && (d) 
\end{tabular}
\end{center}
\caption{
Singular values of the matrix $Q$ from \equ{flattened_M}) in Lemma~\ref{lem:standard_quadratic_form} corresponding to the weight matrix $W_0$ from the $0$-th layer of the four-layer toy deep network (\ref{eq:toy}) using leaky-ReLU activation functions.
The singular values are a function of the coefficient $\rho$ of the subsequent adjustable ResNet layer $1$; $\rho=0$ corresponds to a ConvNet layer, while $\rho=1$ corresponds to a pure ResNet layer.
We plot (a) the largest singular value $s_1$, (b) the smallest nonzero singular value, and (c) the condition number $\kappa$ as a function of $\rho$ for each of the 8 runs of the experiment in a different color.
We also plot (d) the average and standard deviation of $s_1/s_i$ as a function of $i$ for four values of $\rho$.
The vertical axes of (a), (b), and (d) are on a log scale.
We observe that the singular values of the matrix $Q$ increase steadily with $\rho$ and the condition number decreases steadily with $\rho$, meaning that the eccentricity of the loss landscape also decreases steadily with $\rho$.
    }
    \label{fig:evolution}
\end{figure}

\noindent
{\bf Numerical validation.}~
To conclude the paper, we report on the results of a numerical experiment that validates our theoretical results on the singular values of deep networks and the conclusion that the loss landscape of a ResNet is more stable and less eccentric than that of a ConvNet.
We constructed a toy four-layer deep network
\begin{equation}
f = f_2 \circ  f_1 \circ f_{0} \circ f_{-1}
\label{eq:toy}
\end{equation}
with $f_k$ of the form (\ref{eq:layer}) with the nonlinear activation function $\phi_k$ the leaky-ReLU with $\eta=0.1$ in  (\ref{EQU:non_linearities_eta}).
For $f_{-1}, f_0, f_2$, we set $\rho=0$ in (\ref{eq:layer}), while for $f_1$ we let $\rho$ roam as a free parameter in the range $\rho\in[0,1]$ that includes as endpoints a ConvNet layer ($\rho=0$) and a ResNet layer ($\rho=1$).
The input $\TrainData$ was two-dimensional, and all other inputs, outputs, and bias vectors were 6-dimensional, making the weight matrix $W_2$ of size $6\times 2$ and all other weight matrices of size $6\times 6$.
We generated the entries of the input $\TrainData$, random weights $W_k$, and random biases $b_k$ as iid zero-mean unit Gaussian random variables and tasked the network with predicting the following six statistics of $\TrainData$ by minimizing the $\ell_2$ norm of the prediction error (squared error):
$$
y = 
\big[
\TrainData[1], \TrainData[2], {\rm mean}(\TrainData), \max(\TrainData), \min(\TrainData),  \|\TrainData\|_2
\big]^T.
$$
For each random $\TrainData,W_k,b_k$ and each value of $\rho\in[0,1]$, we computed the singular values of the matrix $Q$ from \equ{flattened_M} in Lemma~\ref{lem:standard_quadratic_form} that determine the local shape of the piecewise-quadratic loss landscape for the $0$-th layer. 
We conducted 8 realizations of the above experiment and summarize our results in Figure~\ref{fig:evolution}.

Figure~\ref{fig:evolution} not only validates the theory we have developed above, but it also enables us to draw the following concrete conclusions for gradient-based optimization of the family of DNs we have considered.
Recall that the loss surface of these networks consists of pieces of (hyper) ellipsoids according to (35) in Section \ref{subsec:quadr_loss}.
Independent of the location ($W$) on the ellipsoid, the  more eccentric the ellipsoid, the more the gradient points away from the local optimum.
%in a sub-optimal or even poor direction 
%
Figures~\ref{fig:evolution}(a) and (b) indicate for increasing $\rho$ that, since the singular values increase, the image of the unit ball under $Q$ becomes wider, and thus the level sets of the local loss surface become more narrow \cite{Edelman}.
%https://math.mit.edu/~edelman/publications/eigenvalues_and_condition.pdf
Moreover, Figure~\ref{fig:evolution}(c) and (d) show that the local loss surface becomes less eccentric as $\rho$ increases (a sphere would have a ratio equal to 1).
Like our theory, these numerical results strongly recommend ResNets over ConvNets from an optimization perspective.

\section{Conclusions}
\label{sec:conclusions}

In this paper, we have developed new theoretical results on the perturbation of the singular values of a class of matrices relevant to the optimization of an important family of DNs. 
Applied to deep learning, our results show that the transition from a ConvNet to a network with skip connections (e.g., ResNet, DenseNet) 
leads to an increase in the size of the singular values of a local linear representation and a decrease in the condition number.
For regression and classification tasks employing the classical squared-error loss function, the implications are that the piecewise-quadratic loss surface of a ResNet or DenseNet are less erratic, less eccentric and features local minima that are more accommodating to gradient-based optimization than a ConvNet.
Our results are the first to provide an exact quantification of the singular values/conditioning evolution under smooth transition from a ConvNet to a network with skip connections.
Our results confirm the empirical observations from \cite{li2017visualizing} depicted in Figure~\ref{fig:loss_surface}.

Our results also imply that extreme values of the  training data affect the optimization, an effect that is tempered when using mini-batches due to the inherent averaging that takes place.
Finally, our results shed new light on the impact of different nonlinear activation functions on a DN's singular values, regardless of its architecture.
In particular, absolute value activations (such as those employed in the Scattering Network \cite{mallat2012group}) do not impact the singular values governing the local loss landscape, as opposed to ReLU, which can decrease the singular values. 

Our derivations have assumed both deterministic settings and random settings. In the latter, some of our results are asymptotic in the sense that the bounds hold with probability as close to 1 as desired provided that the parameters are set accordingly. 
We have also provided results in the pre-asymptotic regime that enable one to assess the probability that the bounds will hold.

Finally, while our application focus here has been on analyzing the loss landscape of an important family of DNs, our journey has enabled us to develop novel theoretical results that characterize the evolution of matrix singular values under a range of perturbation regimes that we hope will be of independent interest and broader use.

\bigskip
\noindent
{\bf Acknowledgements.}~
Richard Baraniuk was supported by NSF grants CCF-1911094, IIS-1838177, and IIS-1730574; ONR grants N00014-18-12571, N00014-20-1-2534, and N00014-20-1-2787; AFOSR grant FA9550-22-1-0060; and a Vannevar Bush Faculty Fellowship, ONR grant N00014-18-1-2047.
Rudolf Riedi was supported by 
ONR grant N00014-20-1-2787.
Thanks to Micha Wasem for stimulating discussions leading up to Lemma~\ref{lem:wasem} and to Stephen Wright for pointing out the classification connection in~\cite{hui21}.

\bibliographystyle{plain}
\bibliography{Perturb}

\end{document}